\documentclass{article}

\usepackage{microtype}
\usepackage{graphicx}
\usepackage{subcaption}
\usepackage{booktabs} 

\usepackage{amsmath}
\usepackage{amssymb}
\usepackage{mathtools}
\usepackage{amsthm}
\usepackage{amsfonts}      
\usepackage{cases}
\usepackage{bbm}

\usepackage{thmtools,thm-restate}
\usepackage{wrapfig}

\usepackage{url}
\usepackage{xurl}
\usepackage{comment}
\usepackage[table]{xcolor}
\definecolor{row}{RGB}{233, 233, 235} 
\usepackage{nicefrac}      
\usepackage{microtype}
\usepackage{mathtools}
\usepackage{algorithm}
\usepackage{cancel}
\usepackage{makecell}

\usepackage{comment}
\usepackage{enumitem}
\usepackage{graphicx}
\usepackage{subcaption}
\usepackage{float}  
\usepackage{rotating} 
\usepackage{wrapfig}

\usepackage{hyperref}

\usepackage[accepted]{icml2025}

\usepackage[capitalize,noabbrev]{cleveref}
\crefformat{equation}{(#2#1#3)} 
\crefname{section}{Sec.}{Sec.}
\Crefname{section}{Section}{Sections}
\crefname{appendix}{App.}{App.}
\Crefname{appendix}{Appendix}{Appendices}
\crefname{table}{Tab.}{Tab.}
\Crefname{table}{Table}{Tables}
\crefname{figure}{Fig.}{Fig.}
\Crefname{figure}{Figure}{Figures}
\crefname{algorithm}{Alg.}{Alg.}
\Crefname{algorithm}{Algorithm}{Algorithms}
\crefname{theorem}{Thm.}{Thm.}
\Crefname{theorem}{Theorem}{Theorems}
\crefname{corollary}{Cor.}{Cor.}
\Crefname{corollary}{Corollary}{Corollaries}
\crefname{lemma}{Lem.}{Lem.}
\Crefname{lemma}{Lemma}{Lemmas}
\crefname{remark}{Rmk.}{Rmk.}
\Crefname{remark}{Remark}{Remarks}
\crefname{proposition}{Prop.}{Prop.}
\Crefname{proposition}{Proposition}{Propositions}

\usepackage{tikz,tikz-3dplot}
\usetikzlibrary{math, matrix, fadings, shapes, calc, fit, backgrounds, arrows,arrows.meta, decorations.pathreplacing, decorations.markings, positioning}

\usepackage{amsmath,amsfonts,bm}

\def\eqref#1{equation~\ref{#1}}

\def\1{\bm{1}}

\def\vb{{\bm{b}}}
\def\vc{{\bm{c}}}

\def\ve{{\bm{e}}}

\def\vh{{\bm{h}}}

\def\vk{{\bm{k}}}

\def\vs{{\bm{s}}}

\def\vu{{\bm{u}}}
\def\vv{{\bm{v}}}

\def\vx{{\bm{x}}}
\def\vy{{\bm{y}}}

\def\mA{{\bm{A}}}
\def\mB{{\bm{B}}}
\def\mC{{\bm{C}}}

\def\mI{{\bm{I}}}

\def\mW{{\bm{W}}}

\def\mLambda{{\bm{\Lambda}}}

\DeclareMathAlphabet{\mathsfit}{\encodingdefault}{\sfdefault}{m}{sl}
\SetMathAlphabet{\mathsfit}{bold}{\encodingdefault}{\sfdefault}{bx}{n}

\newcommand{\R}{\mathbb{R}}

\newcommand{\softplus}{\operatorname{SoftPlus}}

\newcommand{\Linear}{\operatorname{Linear}}

\DeclareMathOperator*{\argmin}{arg\,min}

\input{my_symbol.sty}

\usepackage{caption}
\usepackage{booktabs}
\usepackage{tabularx}
\usepackage{siunitx}
\usepackage{array,collcell}
\newcommand\AddLabel[1]{%
  \refstepcounter{equation}
  (\theequation)
  \label{#1}
}
\newcolumntype{M}{>{\hfil$\displaystyle}X<{$\hfil}} 
\newcolumntype{L}{>{\collectcell\AddLabel}r<{\endcollectcell}}
\newcolumntype{n}{>{\hsize=.25\hsize\hfil\hskip -.7in$\displaystyle\color{gray}}X<{$\hfil}}
\usepackage{longtable}
\usepackage{multirow}

\usepackage{xcolor}
\usepackage{xspace}
\newcommand{\MM}{Mamba-2\xspace}
\newcommand{\Mdt}{Mamba-$\Delta^\top$\xspace}

\newcommand{\mqar}{\textsc{MQAR}\xspace}
\newcommand{\keepn}{\textsc{Keep $n$-th}\xspace}
\newcommand{\keepfifth}{\textsc{Keep Fifth}\xspace}
\newcommand{\keepfirst}{\textsc{Keep First}\xspace}
\newcommand{\indhead}{\textsc{Induction Heads}\xspace}

\newcommand{\khop}{\textsc{$k$-hop Induction Heads}\xspace}

\icmltitlerunning{Understanding Selectivity in Mamba}

\usepackage{dsfont}
\begin{document}

\twocolumn[
\icmltitle{Understanding Input Selectivity in Mamba: \\ Impact on Approximation Power, Memorization, and Associative Recall Capacity}



\icmlsetsymbol{equal}{*}

\begin{icmlauthorlist}
\icmlauthor{Ningyuan Huang}{appletemp,fi}
\icmlauthor{Miguel Sarabia}{apple}
\icmlauthor{Abhinav Moudgil}{appletemp,mila}
\icmlauthor{Pau Rodr\'iguez}{apple}
\icmlauthor{Luca Zappella}{apple}
\icmlauthor{Federico Danieli}{apple}
\end{icmlauthorlist}

\icmlaffiliation{apple}{Apple}
\icmlaffiliation{appletemp}{Work done while at Apple}
\icmlaffiliation{mila}{Mila Research Institute, Montreal, Canada}
\icmlaffiliation{fi}{Flatiron Institute, New York, USA}

\icmlcorrespondingauthor{Ningyuan Huang}{thuang@flatironinstitute.org}

\icmlkeywords{Mamba, State Space Model, Input Selectivity, Associative Recall, Machine Learning, ICML}

\vskip 0.3in
]



\printAffiliationsAndNotice{}  

\begin{abstract}
State-Space Models (SSMs), and particularly Mamba, have recently emerged as a promising alternative to Transformers.
Mamba introduces input selectivity to its SSM layer (S6) and 
incorporates convolution and gating into its block definition.
While these modifications do improve Mamba's performance over its SSM predecessors, it remains largely unclear how Mamba leverages the additional functionalities provided by input selectivity, and how these interact with the other operations in the Mamba architecture.
In this work, we demystify the role of input selectivity in Mamba, investigating its impact on function approximation power, long-term memorization, and associative recall capabilities.
In particular: (i) we prove that the S6 layer of Mamba can represent projections onto \emph{Haar wavelets}, providing an edge over its Diagonal SSM (S4D) predecessor in approximating discontinuous functions commonly arising in practice; (ii) we show how the S6 layer can dynamically counteract memory decay; (iii) we provide analytical solutions to the \mqar associative recall task using the Mamba architecture with different mixers --- Mamba, \MM, and S4D. We demonstrate the tightness of our theoretical constructions with empirical results on concrete tasks. Our findings offer a mechanistic understanding of Mamba and reveal opportunities for improvement.

\end{abstract}

\section{Introduction}

State Space Models (SSMs) have recently emerged as a promising approach for long-range sequence modeling, due to their computational efficiency compared to Transformers, and parallelizability compared to (nonlinear) Recurrent Neural Networks (RNNs). In particular, Mamba \citep{gu2023mamba, dao2024transformers} demonstrated state-of-the-art performance on various language modeling tasks, with smaller model size and faster inference than Transformers. 
The success of Mamba has largely been attributed to the fact that the parameters in its SSM layer (S6) are input-dependent, leading to improved expressivity compared to its SSM predecessors \citep{cirone2024theoretical}.
The goal of this work is to provide a more structural explanation for Mamba's superior performance. We do so by answering two questions: (i) how does the S6 layer's expressivity translate into its practical performance? (ii) how can the S6 layer interact with the rest of the Mamba block to solve concrete tasks?

We answer question (i) by providing a fine-grained analysis of the S6 layer via the lens of function approximation and long-term memory. 
We prove that the S6 layer can represent projections onto \emph{Haar wavelets} and thus efficiently model discontinuous signals, which is relevant for solving practical tasks.
Moreover, we show that the S6 layer still suffers from exponential memory decay, but highlight a mechanism which allows it to dynamically counteract such decay.

Building on the understanding of the S6 layer, we then investigate how S6 interacts with the other components of the Mamba block (which also encompasses a short-convolution and a gate branch, see \cref{tab:compare_mambas}) to tackle the Multiple-Query Associative Recall (\mqar) task \cite{arorazoology}. 
We prove that 1-layer Mamba and \MM models can both solve \mqar, even without gating; we describe how S6 and the convolution interact to achieve this, and how \MM leverages its independent convolutions to get more parameter-efficient solutions.
We also show that a 1-layer Mamba can solve \mqar exactly \emph{even without input-dependence} in its SSM: this (perhaps unexpected) result helps cementing the importance of convolution and gating in the Mamba architecture. This analysis further informs us on how a variation to the functional form of Mamba, and particularly to how the SSM state matrix is affected by the input, can improve its performance on the \indhead task \cite{bietti2024birth, sanford2024onelayertransformersfailsolve}.

We complement our theoretical findings with numerical experiments on synthetic sequence modeling tasks. For the S6 layer, we demonstrate its approximation power on discontinuous functions, and its counteraction of memory decay, via the \keepn task --- a generalization of \keepfirst from \cite{chiang2022overcoming} requiring to memorize the $n$-th token in a sequence. Finally, for the full Mamba model, we confirm empirically that the model sizes prescribed theoretically by our analytical solutions to the \mqar and \indhead tasks are tight in practice. Overall, our contributions can be summarized as follows:
\begin{itemize}
    \item We prove that the S6 layer of Mamba can represent projections onto \emph{Haar wavelets}, providing an edge over the S4D layer in approximating discontinuous functions commonly arising in practice (\cref{sec:theory_approx}).
    \item We use sensitivity analysis to show the S6 layer generally suffers from exponential decay of memory, and describe how it can dynamically counteract it (\cref{sec:sensitivity}).
    \item We show how the Mamba architecture can exactly solve the \mqar task using different SSM mixers, with an explicit characterization of the required model size, which helps explaining their performance difference (\cref{sec:mqar_theory}).
    \item Our findings reveal opportunities to further improve Mamba, such as by changing the way input dependence is incorporated into the SSM state matrix (\cref{sec:induction_head_theory}).
\end{itemize}

\section{Related Work}

\paragraph{State-Space Models (SSMs)} SSMs (i.e., linear RNNs) have recently emerged as a promising sequence-modeling approach, with faster training than (nonlinear) RNNs, and faster inference time than Transformers.
To enable long-term memorization capability, \citet{gu2020hippo} designed SSMs as polynomial approximations of signals, by prescribing non-normal HiPPO matrices as state matrices. To improve computational efficiency, \citet{gu2022efficiently} proposed the S4 model: first by reparameterizing HiPPO as a sum of normal and low-rank matrices, then by further considering a \emph{diagonal} simplification in the S4D model \citep{gu2022parameterization}. All these SSMs are Linear Time-Invariant and thus computationally efficient, but consequently lack the ability to process information in a time-varying, input-dependent manner. Mamba \citep{gu2023mamba} overcomes this by using input-dependent SSM parameters, without sacrificing computational efficiency, showing performance competitive with Transformers on long-range language modeling tasks. \MM \citep{dao2024transformers} simplifies the state matrix to a scalar multiple of identity and modifies the Mamba model architecture, showing further empirical improvements.

\paragraph{Expressivity of SSMs} The analyses on the expressivity of SSMs can be divided into two main approaches: formal language theory and approximation theory. Studies following the first approach investigate what formal languages can SSM recognize \cite{merrillillusion, sarrof2024expressivecapacitystatespace, grazzi2024unlocking}. Studies ascribing to the second approach --- including this work --- characterize what function classes can SSMs approximate: \citet{li2022approximation} showed that SSMs can approximate linear functionals with exponential memory decay; \citet{wang2024inverse} extended this analysis to nonlinear RNNs, showing however that adding nonlinearity (in the hidden state recurrence) does not fix the memory decay issues. \citet{orvieto2024universality} proved that SSMs augmented with MLPs are universal approximators of regular functionals, but this improvement over \citet{li2022approximation} and \citet{wang2024inverse} requires the hidden state size to grow linearly with sequence length. \citet{cirone2024theoretical} extended the results from \citet{li2022approximation} to \emph{input-dependent} SSMs such as Mamba, showing their universal approximation on the class of nonlinear functionals arising from controlled differential equations. While \citet{cirone2024theoretical} highlighted how Mamba can approximate a \emph{larger} class of functionals than S4D, we explore the consequences of this observation by characterizing specific function classes arising in practice.

\vspace{-0.3em} 
\paragraph{Associative Recall Capability} Associative Recall (\textsc{AR}) describes the ability of a model to retrieve information from its memory, based on the input context. Tasks for evaluating associative recall capabilities include \indhead \cite{olsson2022context, sanford2024onelayertransformersfailsolve}, \khop \cite{sanford2024transformersparallelcomputationlogarithmic}, \textsc{Multiple-Query Associative Recall} (\mqar) \cite{arorazoology}, and \textsc{Needle-in-the-Haystack} \citep{gkamradt-haystack}.Empirically, Mamba \cite{gu2023mamba} and Mamba-2 \cite{dao2024transformers} demonstrated performance competitive with Transformers on (a simple version of) \indhead and \mqar, respectively.
Theoretical understanding of how language models perform associative recall begins to emerge: \citet{bietti2024birth} constructed a 2-layer Transformer with positional encoding that solves the \indhead task; \citet{sanford2024transformersparallelcomputationlogarithmic} extended such construction to a log-depth Transformer that solves the \khop task; \citet{arorazoology} showed that a gated convolution model can solve \mqar. However, these constructions all require the model size to scale with the input sequence length.
In this work, we show that $1$-layer Mamba models can solve \indhead and \mqar with model size \emph{independent of} the sequence length, highlighting the role of the convolution operation and the gate branch, that has been previously overlooked in the literature.

\section{Preliminaries: Linear RNNs as SSMs}\label{sec:prelim}
The application of a Linear RNN can be interpreted as the discrete solution of a \emph{linear} dynamical system, or \emph{SSMs}, 
\begin{equation}
\dot{\vh}(t) = \mA(t)  \vh(t) + \mB(t) \, x(t), \qquad   \vy(t) = \mC(t) \vh(t).
    \label{eqn:hidden_dyn}
\end{equation}

Here, the input $x(t)\in\mathbb{R}$ acts as forcing term for the \emph{hidden state} $\vh\in\mathbb{R}^{N}$ through the application of the \emph{input matrix} $\mB(t)\in\mathbb{R}^{N\times 1}$. The natural evolution of the hidden state is dictated by the \emph{state matrix} $\mA(t)\in\mathbb{R}^{N\times N}$. Finally, the output $\vy\in\mathbb{R}^{d_y}$ is obtained by linearly transforming the hidden state via the \emph{output matrix} $\mC(t)\in\mathbb{R}^{d_y\times N}$. Collectively, we call $\mA(t), \mB(t), \mC(t)$ the \emph{SSM parameters}.
For computational efficiency, modern Linear RNNs consider \emph{diagonal} state matrices, with negative eigenvalues $\mA(t)=\mLambda(t)=-\operatorname{diag}([\lambda_1(t), \ldots, \lambda_N(t)])$. This simplifies the solution of \cref{eqn:hidden_dyn}, and ensures stability in the evolution of the hidden state. Under this assumption, and considering an initial state $\vh(0)\equiv\boldsymbol{0}$, we can explicitly write the hidden state solution as an integral function of the input \cite{dahleh2004lectures}:
\begin{equation}
    \vh(t) = \int_0^t e^{\int_s^t\mLambda(r)\,dr} \mB(s) \, x(s)\,ds.
    \label{eqn:cont_sol_general}
\end{equation}

Generally, inputs to Linear RNNs are provided as (discrete) sequences of values $[\vx_t]_{t=1}^T$, and thus we consider a \emph{discretization} of system \cref{eqn:hidden_dyn}. This amounts to substituting the \emph{differential} equation with an (approximating) \emph{recurrent} one:
\begin{equation}
\cref{eqn:hidden_dyn}\, \approx \,
    \vh_{t} = \overline{\mLambda}_t \vh_{t-1} + \overline{\mB}_t \, x_t, \qquad
    \vy_t = \overline{\mC}_t \vh(t),
    \label{eqn:discrete_dyn}
\end{equation}
where $\overline{\mLambda}_t,\overline{\mB}_t,\overline{\mC}_t$ are the discrete counterparts of $\mLambda(t)$, $\mB(t)$, $\mC(t)$, respectively.
Notice we can recover an explicit solution to \cref{eqn:discrete_dyn} by unrolling the recurrence relation starting from $\vh_0\equiv\boldsymbol{0}$, to obtain
\begin{equation}
    \vh_t = \sum_{s=1}^t \left(\prod_{r=s+1}^{t} \overline{\mLambda}_r\right) \overline{\mB}_s \, x_s.
    \label{eqn:disc_sol_general}
\end{equation}

The discretized SSM parameters are usually obtained following a Zeroth-Order Hold (ZOH) scheme \cite{toth2008crucial}. Given a discrete time-step $\Delta(t) \in \R^+$, ZOH prescribes
\begin{equation}
\label{eqn:zoh_disc}%
\begin{split}
    \overline{\mLambda}_t &= e^{\mLambda(t) \Delta(t)}, \qquad\qquad\overline{\mC}_t = \mC(t),\\
    \overline{\mB}_t &= (\mLambda(t) \Delta(t))^{-1} (e^{\mLambda(t) \Delta(t)} - I) (\mB(t) \Delta(t)).\\
\end{split}
\end{equation}
Note that in the original Mamba formulation the authors use Forward Euler for $\mB(t)$ for simplicity, $\overline{\mB}_t=\mB(t) \Delta(t)$.

In general, there is some flexibility in the choice of the functional form that the system parameters can take. The main requirements are that: (i) $\overline{\mLambda}_t$ is diagonal(-izable), so that the product $\prod_{r=s+1}^{t} \overline{\mLambda}_r$ in \cref{eqn:disc_sol_general} can be computed efficiently; (ii) the eigenvalues of $\overline{\mLambda}_t$ are bounded in $[-1,1]$, to ensure stability; and that (iii) the SSM parameters do \emph{not} depend on the state $\vh(t)$, so to keep the system linear in $\vh(t)$, and allow to compute its solution in parallel along $t$. The most relevant choices analyzed in this paper are: S4D \cite{gu2022parameterization}, which models linear time-invariant dynamics,
\begin{equation}
\Delta(t) \coloneqq 1, \;\;  \mLambda(t) \coloneqq \mLambda, \;\; \mB(t) \coloneqq \mB, \;\; \mC(t) \coloneqq \mC; \label{eqn:s4d_params}
\end{equation}
and Mamba \cite{gu2023mamba}, which models time-varying (input-dependent) dynamics,
\begin{equation}   \label{eqn:mamba_discrete}%
\begin{split}
  \Delta(t) &\coloneqq \softplus( \Linear(x(t))), \quad  \mLambda(t)  \coloneqq \mLambda, \\
    \mB(t) &\equiv \mB(x(t)) \coloneqq \Linear(x(t)) ,      \\
    \mC(t) &\equiv \mC(x(t)) \coloneqq \Linear(x(t)).   
\end{split}
\end{equation}
Note that both the \MM mixer and \emph{Linear Attention} \cite{katharopoulos2020transformers} can be interpreted as specializations of Mamba \cite{dao2024transformers}, where $\overline{\mLambda}_t \equiv \lambda_t\mI$ and $\overline{\mLambda}_t \equiv\mI$, respectively.

\section{Mamba SSM Mixer Layer Analysis}\label{sec:layer}

In this section, we focus on the core of the Mamba architecture: its SSM mixer layer, also referred to as S6. The goal is to understand how input-selectivity impacts its expressivity and long-range memorization capacity. We show that an S6 layer can represent projections onto wavelets, thus efficiently modeling discontinuous signals. This is useful in practice, e.g., for isolating specific tokens in a sequence. For long-range memorization, we use sensitivity analysis to show that, while S6 suffers from exponential decay of memory (akin to S4D), input-selectivity allows to decrease the rate of such memory decay by ``freezing time''. We validate our theoretical insights by experimenting on the \keepn task.

\subsection{Function Approximation Power: Expressing Wavelets via the S6 Layer} \label{sec:theory_approx}

We begin by analyzing the expressivity of the S6 layer and the power of its input-dependent discretization in terms of function approximation capabilities. We prove that an S6 layer can approximate projections onto wavelets arbitrarily well (\cref{thm:mamba_wavelet}), while an S4D layer can at best project onto Fourier bases. Consequently, in approximating target functions with discontinuities, S6 achieves a faster approximation rate than S4D (\cref{cor:mamba_wavelet}). The advantage of S6 over S4D in approximating discontinuous functions translates into their performance differences in memorization tasks.

\paragraph{Linear RNNs as Time-Projection Onto Basis Functions} The idea that Linear RNNs could perform projections onto specific sets of basis functions is not new, and indeed served as theoretical grounding for the HiPPO work \cite{gu2020hippo}. However, to our knowledge, this interpretation has not yet been leveraged to explain the capabilities of modern Linear RNNs. In what follows, we take such interpretation to analyze their expressivity.
To streamline the analysis and slim notation, we focus on 1D inputs and view them as continuous signals, $\vx_s\equiv x(s)\in\mathbb{R}$, $s \in [0,T]$, without loss of generality. For direct comparison with S4D, we consider a simplified version of S6 where the input-dependence only affects the state matrix via $\Delta(x)$, and does \emph{not} affect $\mB(x_t)$ (i.e., $\mB(x_t) = [B_1, \ldots, B_N]^{\top}$ independent of the input $x_t$). Substituting this into \cref{eqn:cont_sol_general}, and recalling that $\mLambda = -\operatorname{diag}([\lambda_1, \ldots, \lambda_N])$, each component $n=1, \ldots, N$ of the hidden state can be evaluated separately as an inner product between time-dependent functions:
\begin{equation}
    h_n^{\texttt{M}}(t) = \int_{0}^t \underbrace{e^{-\lambda_n \int_{s}^t \Delta(x_r)\,dr} B_n  }_{\eqqcolon g_n^{\texttt{M}}(s; t,x)} \, x(s)\,ds
    = \left\langle g_n^{\texttt{M}}, x\right\rangle.
    \label{eqn:mamba_basis}
\end{equation}
We refer to $g_n^{\texttt{M}}(s; t, x)$ as the \emph{Mamba basis function}, with the notation emphasizing its general dependency on the input signal $x$ up to time $t$. 
For ease of comparison, we can recover an analogous formula to \cref{eqn:mamba_basis} also for S4D, by letting $\Delta(x_t)\equiv1$ $\forall t$ (see also \cref{eqn:s4d_params}). This gives
\begin{equation}
   h_n^{\texttt{S4D}}(t) = \int_{0}^{t} \underbrace{e^{-\lambda_n (t-s)} B_n}_{\eqqcolon g_n^{\texttt{S4D}}(s;t)} \, x(s)\,ds
   = \left\langle g_n^{\texttt{S4D}}, x\right\rangle.
   \label{eqn:s4d_basis}
\end{equation}
As we can see, S4D can provide only exponentials as basis functions. The approximation properties of these functions are limited: this is established in the literature, and ties back to the theory of Vandermonde matrices \citep{gautschi1987vandermonde}, as recently pointed out by \citet{orvieto2024universality}. Their poor performance are mainly due to: (i) the stability constraint, $\operatorname{Re}(-\lambda_n) \le 0$, which causes an exponentially fast decay to $0$, \emph{de-facto} limiting the effective support of said bases; (ii) the large degree of overlap between different bases (obtained by varying the only free parameter $\lambda_n$). The only way to curb these negative effects is by pushing the eigenvalues to be equispaced onto the complex unit disk, namely $e^{-\lambda_n}\to e^{i\frac{2\pi n}{N}}$. This is precisely the strategy recommended by \citet{orvieto2024universality}, however it reduces the application of the S4D layer to simply performing a Fourier transform. In contrast, the additional flexibility provided by the input-selectivity in Mamba allows for a much richer variety of basis functions \cref{eqn:mamba_basis} to be employed in the projection, an example of which is shown next.

\paragraph{Mamba Bases Can Represent Haar Wavelets} Here we provide the main theoretical result in this section, namely that the Mamba S6 layer can perform projections onto Haar wavelets. Due to their ability to capture local aspects of a function such as spikes and discontinuities, wavelets are generally better suited than Fourier bases in solving certain signal processing tasks, (e.g., needles-in-the-haystack \citep{gkamradt-haystack}, transient signals \citep{Mallat2012}).
Recall the Haar wavelets are defined by dilation and translation,
\begin{equation}
\begin{split}
   \psi_{0,0}(s) &= \psi(s) \coloneqq \mathbbm{1}_{[0, \frac{1}{2})}(s) -  \mathbbm{1}_{[\frac{1}{2},1]}(s)\\
    \psi_{j,k}(s) &\coloneqq 2^{j/2}\psi(2^j s - k),
\end{split}
\end{equation}
with $j \in \mathbb{N}$ denoting the dilation scale, and $ k = 0, \ldots, 2^j -1$ the translation. Higher-order wavelets correspond to localized and ``spiky'' bases; see \cref{fig:proof_wavelet} (left) for an illustration. 
\begin{restatable}{theorem}{ThmOne}~\label{thm:mamba_wavelet}
Consider a Haar wavelet $\psi_{j,k}: [0, 1] \to \R$, and the Mamba basis function \cref{eqn:mamba_basis} at $t=1$, 
$g_{j,k}^{\texttt{M}}(s; 1, x) = e^{-\lambda_{j,k} \int_{s}^1 \Delta_{j,k}(x_r) dr} B_{j,k}$. Let $\tilde{x}_s \coloneqq \operatorname{concat}[x_s;s]$ be the input signal augmented with time positional encoding. For any $\epsilon > 0$, there exist $3$ Mamba basis functions $g_{j,k}^{\texttt{M}_1}, g_{j,k}^{\texttt{M}_2}, g_{j,k}^{{\texttt{M}_3}}$ such that the approximation error
\begin{equation*}
    \left|\psi_{j,k}(s) - \left(g_{j,k}^{\texttt{M}_1}(s; 1, \tilde{x}) + g_{j,k}^{\texttt{M}_3}(s; 1, \tilde{x}) \\
    - 2 g_{j,k}^{\texttt{M}_2}(s; 1, \tilde{x}) \right) \right|
\end{equation*}
is smaller than $\epsilon$, $\forall s \in [0, 1]$.
\end{restatable}
The proof relies on tweaking the input-dependent discretization $\Delta(s)$ to output $\Delta(s)\to\infty$ or $\Delta(s)\to0$ (note that $\Delta$ can directly depend on the time variable $s$ instead of the input signal $x_s$, due to time Positional Encoding (PE)). This effectively allows Mamba to represent Heaviside functions as bases: by linearly combining shifted Heaviside bases, one can immediately recover the required Haar wavelets (see \cref{fig:proof_wavelet}, middle-right subplots). The details of the proof are reported in \cref{app:wavelet_proof}, where we also proceed to relax the inclusion of PE as an assumption for \cref{thm:mamba_wavelet}.

\Cref{thm:mamba_wavelet} translates into practical advantages of Mamba over S4D, as Haar wavelets are much better than Fourier bases for approximating \emph{discontinuous} functions common in practice. This is formalized in the following corollary.
\begin{restatable}{corollary}{CorOne}~\label{cor:mamba_wavelet}
    For a piecewise-constant function $\rho(t)$ with $m\geq1$ discontinuities, there exist $N$ Mamba basis functions \cref{eqn:mamba_basis} such that the $L^2$ approximation error
    $\|\rho-\sum_{n=1}^N g_n^{\texttt{M}}\|_{L^2}$
    is of order $\mathcal{O}(2^{-\frac{N}{3m}})$. On the other hand, S4D basis functions can achieve an approximation error of $\mathcal{O}(N^{-1})$.
\end{restatable}
\Cref{cor:mamba_wavelet} stems from \cref{thm:mamba_wavelet}, and from approximation results using Haar wavelets and Fourier bases available in the literature \cite{vetterli2001wavelets,eckhoff1993accurate}; see proof in \cref{app:wavelet_proof}. 
In the following, we illustrate how approximating wavelets translates into advantages on concrete tasks.

\paragraph{Task \keepn}

The goal of the \keepn task is to recover the $n$-th element in a randomly-generated sequence, $y_t=x_n$. This generalizes the \keepfirst task in \cite{chiang2022overcoming} where $n=1$.
The solution of \keepn can be directly represented by combining the projections of a piecewise-constant signal $x(s)=\sum_{i=1}^t x_i\1_{[i-1,i)}(s)$ onto two Heaviside functions $H(s-n), H(s-(n-1))$.
As we have shown in \cref{thm:mamba_wavelet}, one S6 layer in Mamba can reproduce precisely this type of projections, provided the input is augmented with time-positional information. Thus, we arrive at the following Corollary: 
\begin{restatable}{corollary}{CorKeepN}~\label{cor:keepN}
There exists an S6 layer that solves \keepn on input augmented with time Positional Encoding.
\end{restatable}

\begin{table}[tb!]
\caption{\keepfifth experimental results. Average accuracy across 3 seeds with $T=50$ and $|V|=128$. Standard error across 3 seeds is 0.00 for all models. Mamba and S4D models consist of \emph{embedding}, \emph{SSM} ($\vh\in\mathbb{R}^{8\times32}$), and \emph{linear} layers (without convolution and gating, see \cref{tab:compare_mambas}). Positional Encoding (PE) encodes the position in the last element of the embedding, resulting in $|V|$ fewer parameters.}
\label{tab:keep-fifth}
\tiny
\centering
\vspace*{1mm}
\begin{tabular}{cccccc}
\toprule
& \textsc{Mamba+PE} & \textsc{Mamba} & \textsc{S4D} & \textsc{S4D+PE} & \textsc{Transformer} \\
\midrule
Accuracy$\uparrow$ & \textbf{1.00} & 0.08 & 0.09 & 0.08 & \textbf{1.0} \\
Parameters & 9.2k & 9.3k & 8.8k & 8.7k & 6.4k \\
\bottomrule
\end{tabular}
\vspace*{-2mm}
\end{table}
The proof of \cref{cor:keepN} is reported in \cref{app:keepn_proof}; the results in \cref{tab:keep-fifth} verify empirically that Mamba with PE can perfectly solve the task (same as Transformers), whereas Mamba without PE and S4D both fail, highlighting the advantage of Mamba over S4D in approximating discontinuous functions in practice. Additional ablations on model size, sequence length, and the role of PE are reported in \cref{app:experiments-keepnth}.

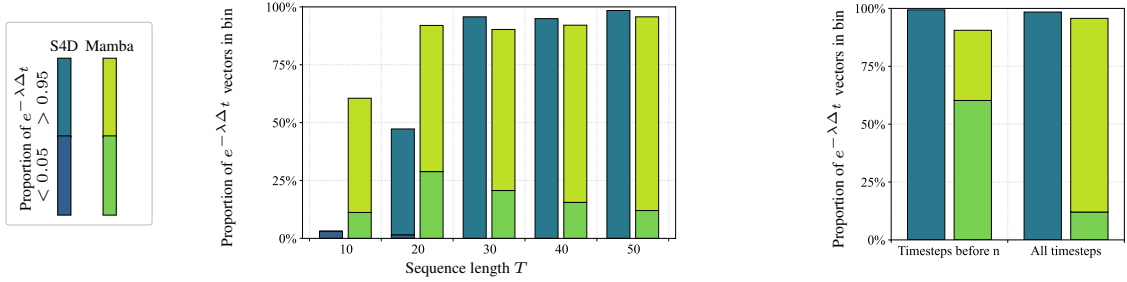
\begin{figure*}[h!]
\centering
  \begin{subfigure}[c]{15mm}
    \centering
    \begin{tikzpicture}
    \definecolor{S4D0}{RGB}{52, 94, 141};
    \definecolor{S4D1}{RGB}{41, 120, 142};
    \definecolor{Mamba0}{RGB}{121, 209, 81};
    \definecolor{Mamba1}{RGB}{189, 222, 38};
    \tikzset{legendEntry/.style = {rectangle, draw, minimum width=5pt, minimum height=30pt, inner sep=0pt}}
    \tikzset{tight matrix/.style={every outer matrix/.append style={inner sep=+0pt}}}
        \matrix (legend)[
            matrix of math nodes,
            tight matrix,
            row sep=-1pt,
            column sep=-4pt,
            nodes in empty cells,
            nodes={font=\tiny, anchor=center}
        ] at (0,0) {
            & \text{S4D} & \text{Mamba}\\
            |[rotate=90,inner ysep=1pt]| >0.95 & |[legendEntry, fill=S4D1]| & |[legendEntry, fill=Mamba1]|\\
            |[rotate=90,inner ysep=1pt]| <0.05 & |[legendEntry, fill=S4D0]| & |[legendEntry, fill=Mamba0]|\\
        };
    \node[anchor=south, font=\tiny, rotate=90, inner sep=0pt, outer sep=0pt] (multirow) at ($(legend-2-1.north west)!0.5!(legend-3-1.north east)$) {Proportion of $e^{-\lambda\Delta_t}$};
    \node[draw=lightgray,fit=(multirow)(legend),rounded corners=1pt]{};
    \end{tikzpicture}
  \end{subfigure}
  \hspace{10mm}
  \begin{subfigure}[c]{70mm}
    \centering
    \begin{sideways}
      \makebox[35mm][r]{\tiny Proportion of $e^{-\lambda\Delta_t}$ vectors in bin}
    \end{sideways}
    \includegraphics[height=40mm]{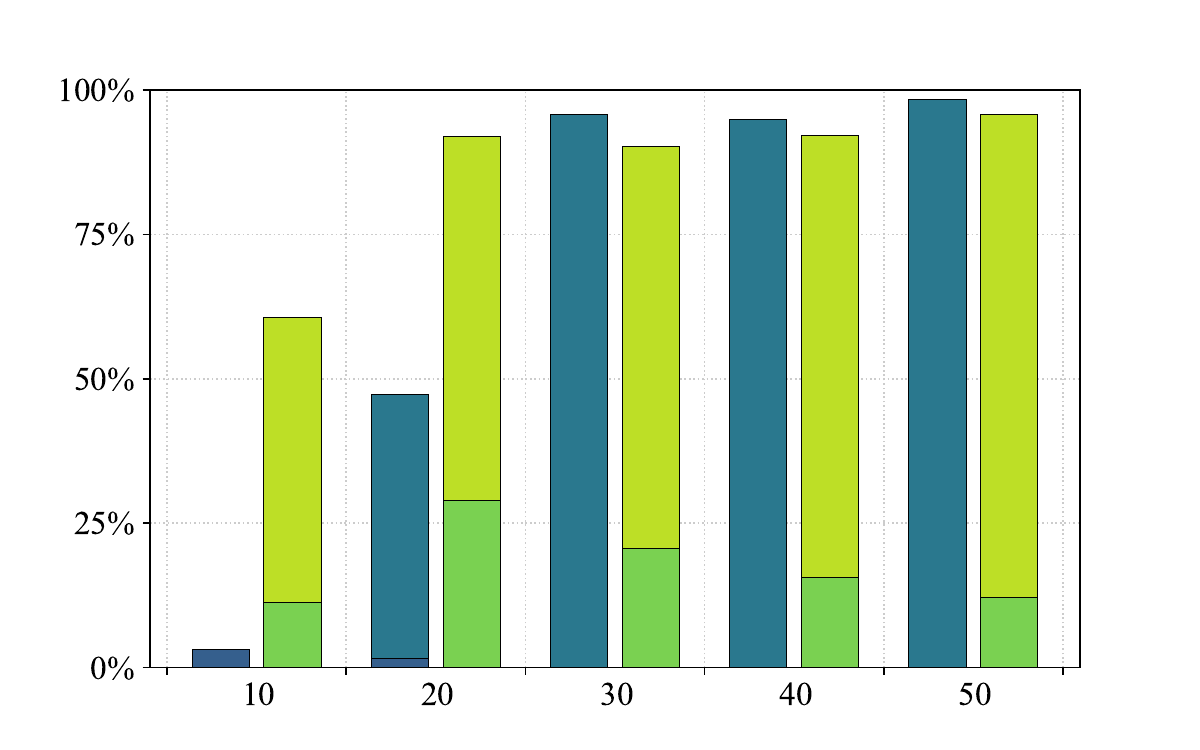}\\[-4mm]
    {\tiny Sequence length $T$}
    \label{fig:delta_t_histogram}
  \end{subfigure}
  \hspace{10mm}
  \begin{subfigure}[c]{45mm}
    \centering
    \begin{sideways}
      \makebox[35mm][r]{\tiny Proportion of $e^{-\lambda\Delta_t}$ vectors in bin}
    \end{sideways}
    \includegraphics[height=40mm]{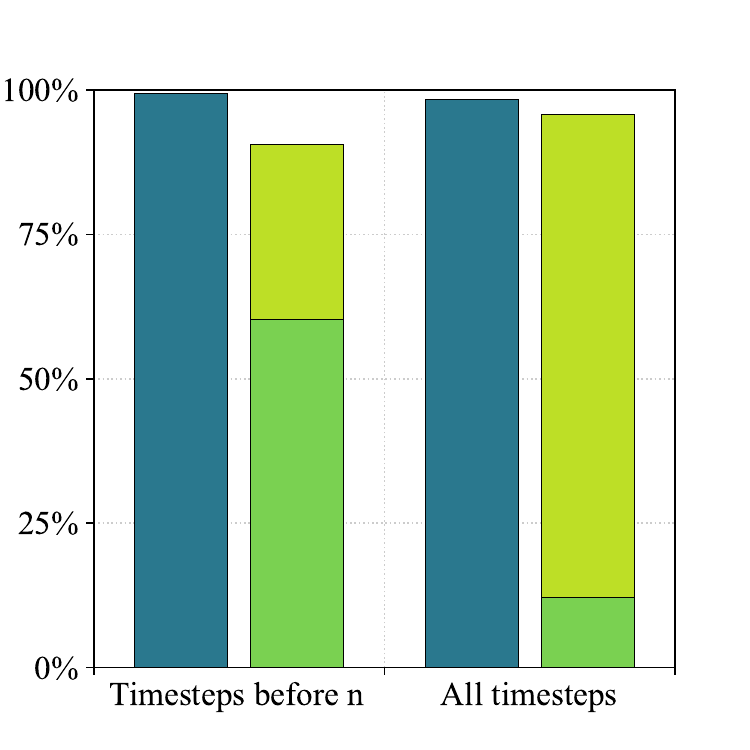}
    \\[-4mm]\phantom{Invisible x-label}
    \label{fig:delta_t_per_timestep}
  \end{subfigure}
  \caption{
  Distribution of $e^{-\lambda\Delta_t}$ computed on test inputs by models trained to successfully solve \keepn tasks for various sequence lengths $T$ (same setup as \cref{tab:keep-fifth}). The left histogram confirms that the Mamba model must push $e^{-\lambda\Delta_t}\to1$ as $T$ increases, as implied by \cref{lemma:constant}, to retain information throughout the sequence. Also the S4D must behave similarly. Meanwhile, the right histogram shows that, compared to S4D, Mamba has additional flexibility in forgetting irrelevant information ($e^{-\lambda\Delta_t}$ is mostly 0 before timestep $n$) and memorizing information selectively ($e^{-\lambda\Delta_t}$ is mostly 1 from timestep $n$ onwards), while S4D is forced to memorize indiscriminately.
    }
  \label{fig:keep-fifth}
\end{figure*}

\subsection{Long-Range Modelling: Sensitivity Analysis}
\label{sec:sensitivity}

In this section, we examine the long-range memorization capacity of SSM layers, by performing a sensitivity analysis of the layer output with respect to changes to the input, as the sequence length increases. To this end, we analyze the derivative of the SSM hidden state at time $t$ with respect to the past input at time $j$, $|\frac{\partial h_t}{\partial x_j}|$. We argue that preserving sensitivity (i.e., a non-zero derivative) is \emph{necessary} for memorization: if the past input has no impact on the current state, one cannot hope for any information about it to be retained.
With \cref{lemma:exp_decay}, we show how generally the sensitivity of an S6 layer decays exponentially fast, similarly to S4D. Thanks to input selectivity, however, the S6 layer can adjust the rate of this decay, thus dynamically tweaking the amount of information to retain. In \cref{lemma:constant} we illustrate this mechanism, which proves to be useful for solving the task in \cref{sec:induction_head_theory}.

For simplicity, we consider 1D inputs $x_t\in\mathbb{R}$. Given a generic, input-dependent recurrence relationship as in \cref{eqn:disc_sol_general}, we show in \cref{app:sensitivity_proof} that the sensitivity of the state with respect to its inputs at the $j$-th instant $x_j \in \R$ is given by
\begin{equation}
  \begin{split}
    \frac{\partial \vh_t}{\partial x_j} &=\frac{\partial}{\partial x_j} \textstyle\left(\sum_{s=1}^t \left(\prod_{r=s+1}^t \overline{\mLambda}_r\right) \overline{\mB}_s x_s\right)\\
    &=\textstyle\left(\prod_{r=j+1}^t \overline{\mLambda}_r \right) \left(\frac{\partial}{\partial x_j}(\overline{\mB}_jx_j) \right. \\
    &\textstyle\quad\left. + \frac{\partial \overline{\mLambda}_j}{\partial x_j}\sum_{s=1}^{j-1} \left(\prod_{r=s+1}^{j-1} \overline{\mLambda}_r\right) \overline{\mB}_s x_s \right).\\
  \end{split}
  \label{eqn:general_sensitivity_main}
\end{equation}

\begin{restatable}{lemma}{LemSensty}
~\label{lemma:exp_decay}
    Consider the hidden states arising from the S4D and S6 SSMs defined in \cref{eqn:s4d_params} and \cref{eqn:mamba_discrete}.
    The sensitivity of the $n$-th component of their states at time $t$ with respect to the input at time $j \ll t$ is given by, respectively,
    \begin{equation*}
       \begin{split}
        \left|\frac{\partial h_t^{\texttt{S4D}}}{\partial x_j}\right| &=  \tilde{c}(\lambda_n, B_n, x_{\le j})  \, e^{-\lambda_n(t-(j+1))},\\
         \left|\frac{\partial h_t^{\texttt{M}}}{\partial x_j}\right| &= \tilde{c}(\Delta, \lambda_n, B_n, x_{\le j}) \, e^{-\lambda_n\sum_{r=j+1}^t\Delta(x_r)}, \\  
       \end{split}
    \end{equation*}
    where $ \tilde{c}(\Delta, \lambda_n, B_n, x_{\le j})$ depends on the input subsequence $x_{\le j}$, independent of the sequence length $t$.
\end{restatable}

\Cref{lemma:exp_decay} shows that both S4D and S6 have exponential decay of sensitivity when the sequence length $t$ increases. For S4D, the only mitigation strategy is to set $\lambda_n \to 0$ (and thus $e^{\lambda_n} \to 1$). While S6 can implement the same strategy, it can also counteract the decay by adapting the input-dependent discretization, as formalized in \cref{lemma:constant}.

\begin{restatable}{lemma}{LemSenstyCst}
~\label{lemma:constant}
    Consider the discrete-time S6 in \cref{eqn:mamba_discrete} where $\overline{\mB}_t =  [B_1(x_t), \ldots, B_n(x_t)]^{\top}  \in \R^{N}$.
    Suppose there exists a constant $c \ge 0$ such that
    \begin{equation}
       \lim_{t \to \infty}  \lambda_n \sum_{r=1}^t \Delta(x_r) \le c. \label{eqn:ignore}
    \end{equation}
   Then the sensitivity of the $n$-th component of the state at time $t$ with respect to any input $x_j$ is lower bounded by
   \begin{equation}
   \lim_{t \to \infty} \left|\frac{\partial h_t^{\texttt{M}}}{\partial x_j}\right| \ge e^{-c} \left| \frac{\partial}{\partial x_j} B_n(x_s) \, \Delta(x_s) \, x_s \right| .\label{eqn:state_blowup}
    \end{equation} 
\end{restatable}

This implies that, to retain sensitivity for longer sequences, Mamba must necessarily push $\lambda\Delta(x_t)\to0$ (or equivalently, $e^{-\lambda\Delta(x_t)}\to1$). We illustrate this with the \keepn task shown in \cref{fig:keep-fifth}, where we report the distribution of the learned parameters $e^{-\lambda\Delta(\vx_t)}$ as we increase the sequence length. The shift towards $1$ appears clear, validating the condition discussed in \cref{lemma:constant}.

\begin{remark}
While in this section we mainly focus on the properties of the \emph{original} Mamba mixer layer \cite{gu2023mamba}, we note that the results proven in \cref{thm:mamba_wavelet}, \cref{lemma:exp_decay} and \cref{lemma:constant} hold analogously for the \MM mixer \cite{dao2024transformers}. 
We remind that the \MM mixer layer can be seen as a simplification of Mamba's, whereby the state matrix is parameterized by a single scalar, $\boldsymbol{\Lambda}=\lambda \boldsymbol{I}$, rather than its full diagonal. Nonetheless, we do not rely on Mamba's additional flexibility for our derivations; choosing a suitable scalar $\lambda$ suffices (see details in \cref{app:wavelet_proof}).
\end{remark}

\section{Full Mamba Architecture Analysis} \label{sec:architecture}

\begin{table*}[htb!]
\caption{Comparison of Mamba (including S4D as a special case) and \MM (single-head) mixers. We denote with $\odot$ the Hadamard (elementwise) product, $\otimes$ the Kronecker (outer) product, $\operatorname{conv}(\vx)_{[t]}$ the $t$-th output of the convolution, and $\sigma$ the pointwise nonlinearity.}
\label{tab:compare_mambas}
\vspace*{1mm}
\centering
\scriptsize
\begin{subequations}
\begin{tabularx}\textwidth{@{}MML@{}}
\toprule 
    \textsc{Mamba}&
    \textsc{\MM}   \\
\midrule 
    \mLambda\in \R^{d\times N},  \, \vx_t \in \R^d, \, \vh_t \in \R^{d \times N}&
    \mLambda = \lambda \in \R, \, \vx_t \in \R^{d}, \, \vh_t \in \R^{d\times N}&
    eqn:size_compare \\[2pt]
\midrule
    {\begin{tabularx}\linewidth{@{}Mn@{}}\hat{\vx}_t = \sigma(\operatorname{conv}(\vx)_{[t]}) \in \R^d & \end{tabularx}}&
    \hat{\vx}_t = \sigma(\operatorname{conv}_u(\Linear(\vx))_{[t]}) \in \R^d &
    eqn:mamba_conv
    \\[1pt]   
    {\begin{tabularx}\linewidth{@{}Mn@{}}\Delta_t = \operatorname{SoftPlus}(\Linear(\hat{\vx}_t)) \in \R^{d} &(\Delta_t^{\text{S4D}} = \mathbf{1})\end{tabularx}}&
    \Delta_t = \operatorname{SoftPlus}(\Linear(\vx_t)) \in \R &
    eqn:delta_compare
    \\[1pt]
    {\begin{tabularx}\linewidth{@{}Mn@{}}\mB_t = \Linear(\hat{\vx}_t) \in \R^N & (\mB_t^{\text{S4D}} = \mB)\end{tabularx}}&
    \mB_t = \sigma(\operatorname{conv}_B(\Linear(\vx))_{[t]}) \in \R^N &
    eqn:mamba_compute_B
    \\[1pt]
    {\begin{tabularx}\linewidth{@{}Mn@{}}\mC_t = \Linear(\hat{\vx}_t) \in \R^N & (\mC_t^{\text{S4D}} = \mC)\end{tabularx}}&
    \mC_t = \sigma(\operatorname{conv}_C(\Linear(\vx))_{[t]}) \in \R^N &
    eqn:mamba_compute_C
    \\[1pt]
\midrule
    \vh_t  = e^{\mLambda\odot(\Delta_t\otimes\bm{1}_N)} \odot \vh_{t-1} + (\Delta_t \odot \hat{\vx}_t) \otimes \mB_t&
    \vh_t =e^{\lambda \Delta_t}\vh_{t-1} +  \left( \Delta_t \hat{\vx}_t \right) \otimes \mB_t &
    eqn:recurrence_compare\\[1pt]
\midrule
    \multicolumn{2}{c}{$\vy_t = \vh_t \, \mC_t \in \R^d, \quad \tilde{\vy}_t = g(\vx_t) \odot \vy_t \coloneqq \sigma(\Linear(\vx_t)) \odot \vy_t \in \R^{d}$} &
    eqn:gating\\[1pt]
\bottomrule
\end{tabularx} 

\end{subequations}
\end{table*}

In this section, we describe the full Mamba architecture and study how its SSM mixer coordinates with the other components in the model to efficiently solve associative-recall tasks.
A full Mamba architecture includes an embedding layer, a number of Mamba mixer blocks, and an output layer. The mixer block is further composed of a short convolution, an SSM, and a gate --- we refer to \cref{tab:compare_mambas} for details in the differences between Mamba and \MM, but point out that \MM leverages \emph{three independent} short convolutions (rather than a \emph{single common} one) to compute its SSM parameters, as outlined in \cref{eqn:mamba_conv}.

While Mamba and \MM achieve performance competitive with Transformers and outperform their SSM predecessors in solving \mqar and \indhead,
the details of how this solution can be assembled by the architecture remain elusive, with only lower bounds on the SSM mixer size available in the literature \cite{arora2024simple, sanford2024transformersparallelcomputationlogarithmic}, lacking the consideration of other components of Mamba such as convolution and gating. Here we close this gap by providing analytical constructions for a 1-layer Mamba model that can exactly solve these tasks for any input. Perhaps counterintuitively, as we prove in \cref{lemma:s4d_mqar}, the components of a single Mamba mixer block are already powerful enough to solve \mqar exactly, even just using an S4D mixer layer (replacing the S6).
With \cref{lemma:mamba_nogate} and \cref{lemma:mqar_attn_sol} we further show how, thanks to the input selectivity of S6, Mamba and \MM can leverage leaner mechanisms to solve \mqar. Particularly, the S6 layer can use $\mB_t$ and $\mC_t$ to efficiently structure information within its hidden state, and retrieve it when required. Finally, we show how the ability to structure the hidden state (used for \mqar) can be combined with the capacity to dynamically adjust the rate of memory decay (investigated for \keepn) to exactly solve \indhead with a variant of S6. We name this variant \Mdt, and discuss it in \cref{sec:induction_head_theory}.

We remark that the constructions described in this section are just \emph{possible} solutions that the Mamba architectures can implement, and we do not exclude the existence of alternative ones. Nonetheless, in \cref{fig:mqar_sweeps} we verify empirically that our solutions are tight in terms of model size.

\begin{table*}[ht]
  \caption{Overview of exact solutions to the \mqar task that can be implemented by the Mamba model with S4D-mixer in \cref{lemma:s4d_mqar} (top), Mamba-mixer in \cref{lemma:mamba_nogate} (middle), and \MM-mixer in \cref{lemma:mqar_attn_sol} (bottom). While S4D-mixer lacks input selectivity in its SSM layer, it can solve MQAR via the gated-convolution mechanism on a larger embedding space (top). By contrast, both Mamba and \MM can solve MQAR with the same selective SSM layer construction \emph{without gating}, and differ on the choice of convolutions (middle, bottom).}
   \label{fig:proof_MQAR}
   \vspace*{1mm}
   \centering
   
   \includegraphics[trim={0 214mm 0 0},clip,width=\linewidth]{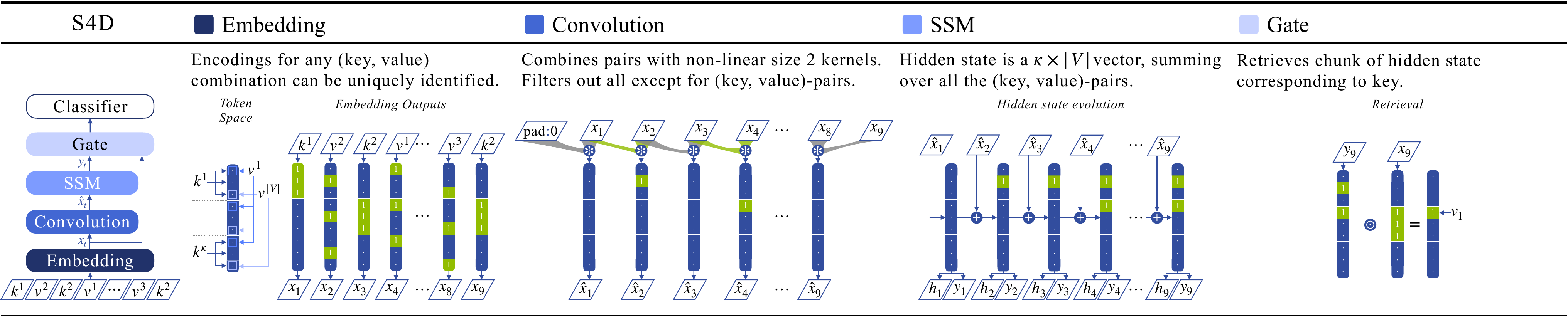}
   \includegraphics[trim={0 214mm 0 0},clip,width=\linewidth]{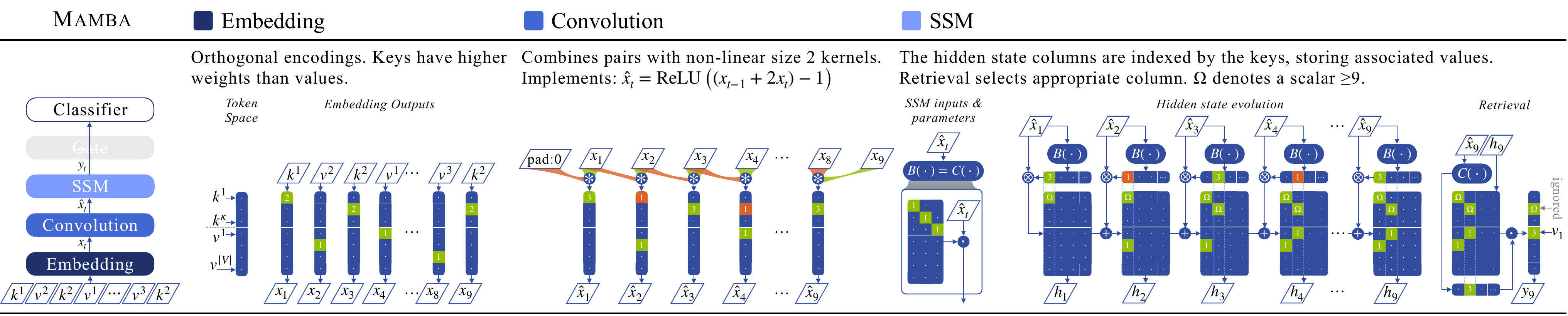}
   \includegraphics[trim={0 214mm 0 0},clip,width=\linewidth]{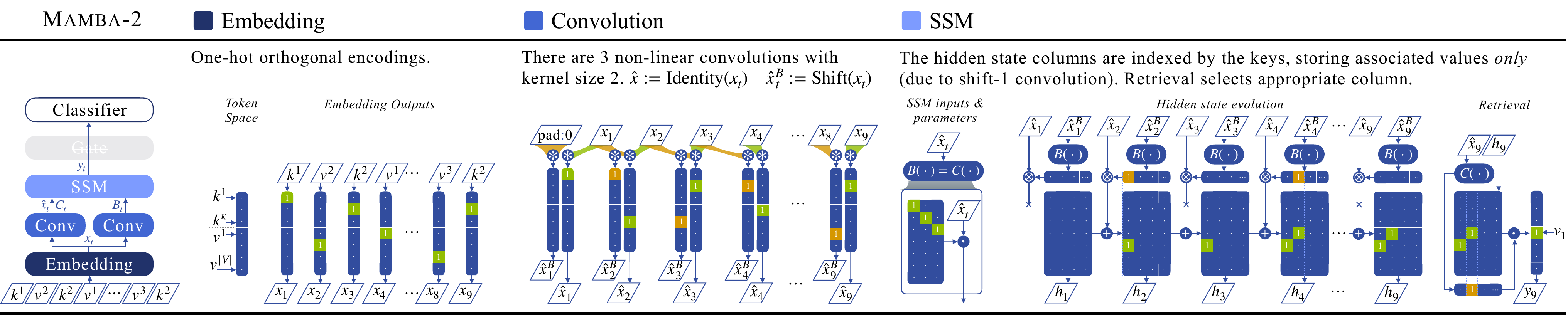}
  
\end{table*}

\subsection{$1$-Layer Mamba Can Solve \mqar} \label{sec:mqar_theory}
The \mqar task \cite{arorazoology} prescribes input and output sequences as follows
\begin{align*}
\vx &= [\underbrace{k_1,v_1,\ldots, k_{\kappa},v_{\kappa}}_{\kappa\text{ key-value pairs}},&&|\underbrace{\ldots, k_{i_1},\ldots,k_{i_{\kappa}}, \ldots}_{\text{shuffled keys, interwoven with noise}} ], \\
    \vy &= [\times, \times, \ldots,\times, \times, &&| \quad \ldots,v_{i_1},\ldots,v_{i_j},\ldots \quad ].
\end{align*}
The keys $k_i$ are randomly chosen from a key set of size $\kappa$, whereas the values $v_i$ and the noise are randomly taken from a vocabulary of size $|V|$. The goal is to correctly predict the value associated with the corresponding key at the query positions, while other non-query positions (denoted with $\times$) are ignored.

In this section we prove that the \mqar task can be exactly solved by three architectures: vanilla Mamba (\cref{lemma:mamba_nogate}), \MM (\cref{lemma:mqar_attn_sol}), and Mamba with an S4D mixer (\cref{lemma:s4d_mqar}), which we refer to as Mamba-S4D.
Next we provide a sketch of the proofs, which we further illustrate in \cref{fig:proof_MQAR}, and defer the full details to \cref{sec:task_proofs}. 
\begin{restatable}{theorem}{LemMambaNogate}~\label{lemma:mamba_nogate}
  There exists a $1$-layer Mamba model without gating that solves \mqar with $\kappa$ pairs using embedding size $d = O(\kappa+\log |V|)$, and state size $N = \kappa$.  
\end{restatable}
\begin{proof}[Proof sketch]
We first prove a construction using standard basis embeddings with $d = \kappa+|V|$, and then apply Johnson-Linderstrass (JL) Lemma (see \cref{JL:inner-prod}) to reduce dimensionality. We specify the Mamba model as follows.
A size-$2$ (nonlinear) convolution kernel combines token \emph{pairs}, extracting key-value pairs and removing non-informative ones (e.g., value-key, value-value). The SSM layer (S6) organizes the hidden state \emph{matrix} \cref{eqn:recurrence_compare} via $\mB_t$ \cref{eqn:mamba_compute_B} such that each column corresponds to a specific key, and holds the value embedding associated with said key (hence the state size $N= \kappa$). At query time, $\mC_t$ \cref{eqn:mamba_compute_C} uses the query(=key) embedding to retrieve the desired value from the correct column in the hidden state matrix. 
\end{proof}

\begin{restatable}{theorem}
{LemMambaTwo}~\label{lemma:mqar_attn_sol}
 There exists a $1$-layer \MM model without gating that solves \mqar with $\kappa$ pairs using embedding size $d = O(\log \kappa + \log |V|)$, and state size $N = \log \kappa$.
\end{restatable}
\begin{proof}[Proof sketch]
 The construction is similar to \cref{lemma:mamba_nogate} for Mamba, but with some simplifications due to the additional flexibility provided by \emph{three} independent convolutions in \MM. Specifically, we set $\operatorname{conv}_B=(1,0)$ as a shift-$1$ convolution kernel (i.e., shifting the input sequence to the right by one position), while $\operatorname{conv}_u \equiv \operatorname{conv}_C$ as the identity maps. Once again, JL Lemma allows us to reduce dimensionalities for both keys and values embeddings.
\end{proof}

\begin{remark} \label{rem:mamba2}
Our construction of \MM is similar to the construction in \cite{li2024power} that shows how convolution-augmented Transformers can solve \mqar. 
Indeed, in light of the convolution layer in \cref{{tab:compare_mambas}}, we can interpret \MM as a convolution-augmented subquadratic Transformer. 
\end{remark}

\begin{remark}
With the results from \cref{lemma:mqar_attn_sol,lemma:mamba_nogate}, we can infer that the extra convolutions included in the \MM layer over vanilla Mamba in general allow for a more parameter-efficient solution of MQAR, by noting that $ d = \log \kappa + \log |V| < \kappa + \log |V|$, and $N = \log \kappa < \kappa$.
\end{remark}

\begin{figure*}
    \centering
    \includegraphics[width=80mm]{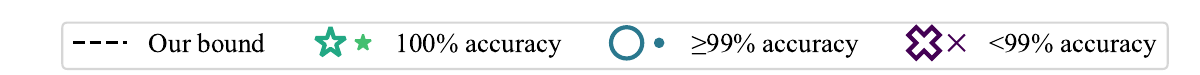}
\begin{minipage}[t][60mm][b]{0.65\textwidth}
\begin{figure}[H]
    \raggedright
    \hspace{3.5mm}%
    \makebox[15mm][c]{\scriptsize \textsc{S4D}}%
    \hspace{22mm}%
    \makebox[15mm][c]{\scriptsize \textsc{Mamba}}%
    \hspace{21mm}%
    \makebox[15mm][c]{\scriptsize \textsc{\MM}}%
    \\[-2mm]
    \centering
    \begin{sideways}
      \makebox[35mm][c]{\tiny Model dimension ($d$)}
    \end{sideways}%
    \includegraphics[trim={1mm 7mm 12mm 15mm},clip,width=35mm]{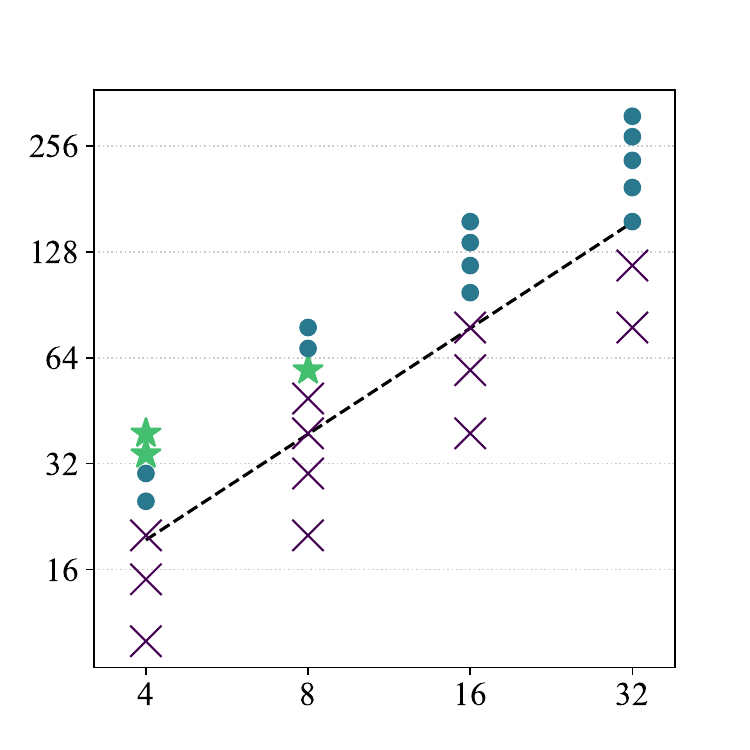}%
    \includegraphics[trim={1mm 7mm 12mm 15mm},clip,width=35mm]{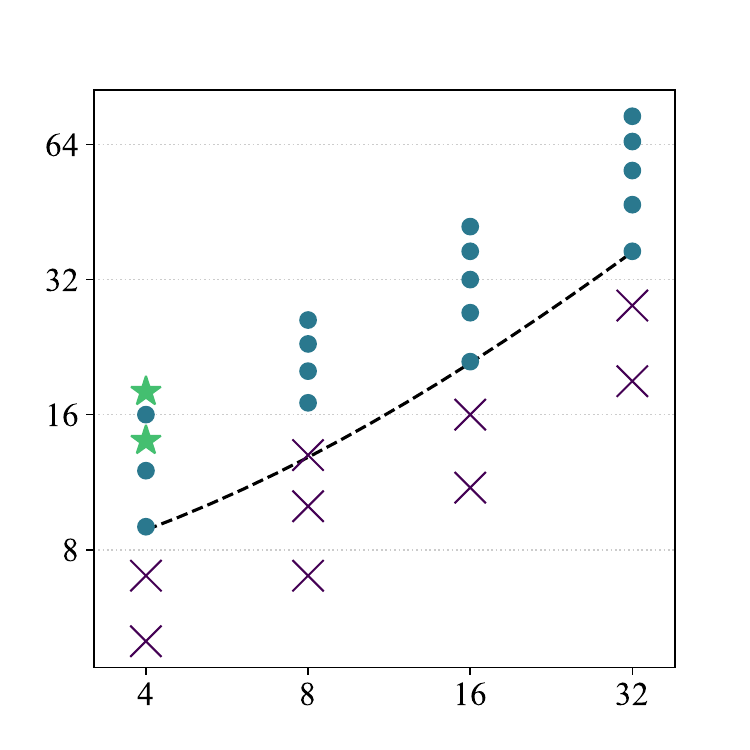}%
    \includegraphics[trim={1mm 7mm 12mm 15mm},clip,width=35mm]{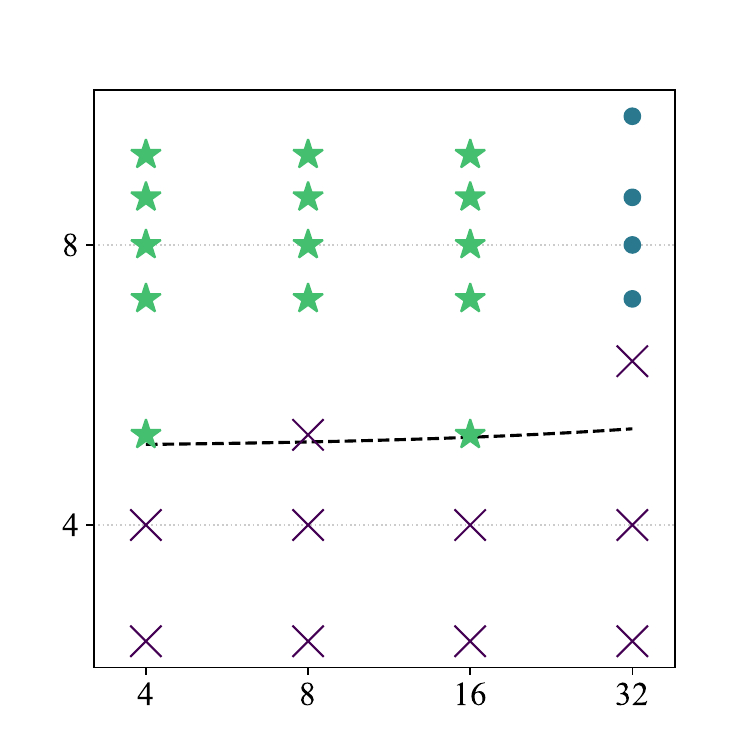}
    \\[-1mm]
    {\tiny Number of keys ($\kappa$)}%
    \vspace*{-2mm}%
    \caption{Trained models accuracy on \mqar task (best of 7 seeds), varying $\kappa$ and $d$. For S4D $N=4$, for Mamba $N=2\kappa$, and for \MM $N=8\ln{\kappa}$. We use $T=100$ and $|V|=128$ for all runs.
    The theoretical bounds on model size for assembling the solutions proposed in \cref{lemma:s4d_mqar,lemma:mqar_attn_sol,lemma:mamba_nogate} (black lines) separate reasonably well models that can achieve 100\% accuracy (above black lines) from those that do not (below). In terms of model size efficiency, \MM is better than Mamba, which in turn is better than S4D.}
    \label{fig:mqar_sweeps}
\end{figure}
\end{minipage}%
\hfill%
\begin{minipage}[t][60mm][b]{0.33\textwidth}
  \begin{figure}[H]
    \raggedright
    \hspace{23mm}%
    \makebox[15mm][c]{\scriptsize \textsc{ Mamba and \Mdt }}%
    \\[-2mm]
    \centering
    \begin{sideways}
      \makebox[35mm][c]{\tiny Model dimension ($d$)}
    \end{sideways}
    \includegraphics[trim={1mm 7mm 12mm 15mm},clip,width=35mm]{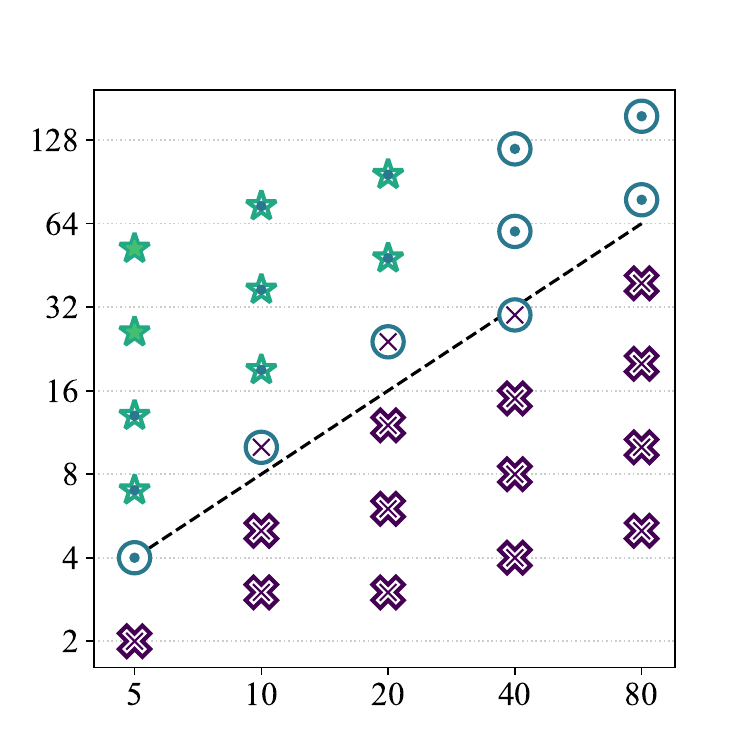}
    \\[-1mm]
    \hspace{10mm}{\tiny Vocabulary size ($|V|$)}%
    \vspace*{-2mm}%
    \caption{Trained models accuracy on \indhead task (best of 5 seeds), varying $|V|$ and $d$, $N=4|V|$. \Mdt's performance (outlined) is equal or better than Mamba's (filled) and only hits $100\%$ above the theoretical bound from \cref{lem:ind_head} (black). }
    \label{fig:one_hop}
  \end{figure}
\end{minipage}
\vspace*{-2mm}
\end{figure*}

\begin{restatable}{theorem}
{LemSfourD}~\label{lemma:s4d_mqar} 
There exists a $1$-layer Mamba model with an S4D mixer that solves \mqar with $\kappa$ pairs, using embedding size $d = O(\kappa \log |V| )$, and state size $N = 1$.  
\end{restatable}
\begin{proof}[Proof sketch]
We first present the construction using $d = O(\kappa |V|) $, and again apply JL Lemma to reduce to $d= O (\kappa \log |V|)$. 
To overcome the limitation given by the lack of input selectivity in the SSM layer, we organize the hidden state along the embedding dimension only, and partition it so that each chunk holds the value associated with a specific key. At query time, the retrieval of the desired key-value chunk is done by leveraging the gating layer.
\end{proof}

\begin{remark} \label{rem:s4d_gatedconv}
Our construction of the S4D-mixer relies both on gated convolution and the SSM recurrence.
\citet{arorazoology} provided a construction to solve \mqar based on gated convolution only.
In contrast, our construction relies on size-$2$ kernels only, thanks to the SSM recurrence.
\end{remark}

To validate our theoretical constructions, we train Mamba, \MM, and Mamba-S4D models with varying embedding size $d$, on \mqar tasks with different number of key-value pairs $\kappa$. The goal is to check how tight in practice are the theoretical bounds on model dimension derived above, and whether indeed trained models must respect them to solve the tasks. Results are reported in \cref{fig:mqar_sweeps}, where dashed curves denote our theoretical bounds and markers indicate empirical results. Notice our bounds in \cref{lemma:s4d_mqar,lemma:mqar_attn_sol,lemma:mamba_nogate} are close to the empirical threshold between models sizes that can recover exact solutions or not, illustrating the tightness of our theoretical results. See additional details in \cref{app:mqar}.

\subsection{$1$-Layer \Mdt Can Solve \indhead}  \label{sec:induction_head_theory}
The \indhead task was first introduced by \citet{olsson2022context} to study Associative Recall capabilities in Transformers. 
Here we use the formulation from \citet{sanford2024onelayertransformersfailsolve}: given an input sequence of tokens $[x_1, \ldots, x_t]$ from a finite vocabulary $x_t\in V$, the goal is to report, for each $x_t$, the token coming immediately after the latest previous occurrence in the input of token $x_i$. That is, the output $y_t$ must be $y_t=x_{j(t)+1}$ where $j(t) = \max\{j: j < t, x_j = x_t \}$ (or a ``blank'' token, $y_i = \times$, if $x_t$ appears for the first time).

Note the \indhead task is similar to \mqar, but with two significant differences. (i) There is no logical distinction between keys and values, so the model needs to identify the role of each token, but this can be handled by the short convolution, as we will show. More importantly, (ii) we need to retain information about only the \emph{latest} previous occurrence of a token, so the model should dynamically forget and remember information pertaining different tokens. Since the latter memorization ability was investigated in \cref{sec:sensitivity}, it is natural to leverage those insights in our solution. To do so efficiently, we introduce a slight variation to the S6 layer: the \Mdt SSM mixer, prescribing the following hidden state evolution
\begin{equation}
    \vh_t = e^{\mLambda \odot (\boldsymbol{1}_d \otimes\Delta(\hat{\vx}_t))} \odot \vh_{t-1} + \hat{\vx}_t \otimes \mB_t. \label{eqn:mamba-dt-trans}
\end{equation}
Comparing this to \cref{eqn:recurrence_compare}, the only difference lies in the action of $\Delta(\hat{\vx}_t)$, which now varies along the \emph{state} dimension $N$, rather than the \emph{embedding} dimension $d$. While \citet{gu2023mamba} hypothesized a similar performance for both versions, our findings reveal that the dependence along state dimension is better suited to solving the \indhead task:

\begin{restatable}{lemma}{LemIndHead}
~\label{lem:ind_head}
There exists a $1$-layer Mamba model with the \Mdt SSM mixer \cref{eqn:mamba-dt-trans} that solves \indhead with vocabulary $V$ using embedding size $d = 2|V|$ and state size $N = |V|$.
\end{restatable}

\begin{proof}[Proof sketch]
The proof follows closely the \mqar construction, in that we leverage the matrix structure in the hidden state such that its columns are indexed by the keys and store the associated values. In the \indhead task, though, each token $x_i$ acts as key in the $(x_i, x_{i+1})$ pair, and as value in the $(x_{i-1}, x_i)$ pair. To handle this distinction, we simply duplicate the embedding, and let the convolution layer correctly combine tokens information pairwise, so that after convolution each token encapsulates information both regarding a value and its preceding key.
Moreover, we use $\Delta(x_t)$ to selectively erase outdated information, or to retain currently valid information, depending on the input observed. Pushing $\Delta(x_t)\to\infty$ flushes a previously memorized value, while $\Delta(x_t)\to0$ preserves it. We refer to \cref{app:indhead_proof} for the detailed proof.
\end{proof}

\begin{remark} \label{rem:induction_head}
To solve \indhead, \citet{bietti2024birth, sanford2024transformersparallelcomputationlogarithmic} constructed $2$-layer Transformers relying on PE and with size scaling as sequence length. On the other hand, we propose a $1$-layer Mamba composing a convolution and a variant SSM layer. Notably, this allows us to drop the PE and thus have the model size depend only on $|V|$, and \emph{not} the sequence length, improving upon the constructions for Transformers. 
\end{remark}

To demonstrate the efficiency of our \Mdt variant on the \indhead task, we compare it against the Mamba baseline, and report results in \cref{fig:one_hop}. For all model sizes considered, \Mdt performs equally or better, demonstrating that selectivity along the state dimension (in the state matrix) improves Mamba's ability to solve the \indhead task.

\section{Conclusion and Future Work}

In this work, we demystify the role of input selectivity in Mamba, showing its impact on approximation power, long-term memory, and associative recall capabilities. We prove that the S6 layer can efficiently represent discontinuous signals and adaptively mitigate sensitivity decay. We also uncover the role of other architectural components in Mamba, particularly convolution and gating. We present a mechanistic explanation of how Mamba solves memorization and associative recall tasks, with tight theoretical model size bounds matching empirical results. Our findings reveal opportunities to further improve Mamba, such as an alternative way to inject input dependence within the SSM state matrix.

Our current theory does not consider the aspects of optimization and generalization, both of which are interesting future directions to explore. Moreover, our analysis focuses on simple associative recall tasks; extending it to more complicated tasks such as \khop \cite{sanford2024transformersparallelcomputationlogarithmic}, \textsc{Sequential Function Composition} \cite{chen2024theoreticallimitationsmultilayertransformer}, and \textsc{Pointer Value Retrieval} \cite{zhang2021pointer} would be a natural next step. %
Overall, our work and proposed improvements add to the growing understanding of SSMs and could accelerate their development.

\section*{Impact Statement}
The goal of this paper is to improve the understanding of the Mamba architecture, specifically through analyzing the role of input selectivity. Our findings contribute to the advancement of State Space Models, which may in turn further democratize access to Large Language Models, sharpening both the existing positive and negative aspects of LLMs. No additional societal impact is expected from this work.

\section*{Acknowledgements}
We thank Pierre Ablin, Samy Bengio, Adam Goliński, Aryo Lotfi, Jason Ramapuram, Jonathan Sheaffer (in alphabetical order) for their helpful feedback and critical discussions throughout the process of writing this paper. We would also like to thank Alberto Bietti for pointers to the associative recall tasks.

\bibliography{ref}
\bibliographystyle{icml2025}

\newpage
\appendix

\onecolumn

\counterwithin*{figure}{part}
\stepcounter{part}
\renewcommand{\thefigure}{A.\arabic{figure}}

\counterwithin*{table}{part}
\stepcounter{part}
\renewcommand{\thetable}{A.\arabic{table}}

\section{Approximation Power of Mamba}

\subsection{Notation}\label{app.notation}
We typically use bold upper case $\mA, \mB, \mC$ to denote matrices and bold lower case $\vx, \vy$ to denote vectors or sequence. In \cref{sec:prelim} and \cref{sec:layer}, the hidden state at time $t$ is denoted as $\vh(t)$ (in the continuous setting) or $\vh_t$ (in the discrete setting). In \cref{sec:architecture}, with a slight abuse of notation, the hidden state $\vh_t \in \R^{d \times N}$ denotes a matrix. We use $\overline{\mA}, \overline{\lambda}$ to denote the discretized versions of $\mA, \lambda$. We use $\R, \mathbb N$ to denote the reals and the natural number. The identity matrix is denoted as $\mI$, where the all-ones vector is denoted as $\boldsymbol{1}$. We let $\operatorname{diag}(\vv)$ be the diagonal matrix with diagonal filled with the vector $\vv$. We denote $\operatorname{SoftPlus}, \operatorname{ReLU}, \operatorname{SiLU}$ as the corresponding pointwise nonlinearity $\sigma$, and $\Linear$ as the linear layer. We let $\mathbbm{1}$ be the indicator function, $H(s)$ be the heaviside function. We denote with $\odot$ the Hadamard (elementwise) product and $\otimes$ the Kronecker (outer) product. We use $d$ for embedding size, $N$ for state size, and $t$ or $T$ for sequence length.

\subsection{Mamba Approximates Haar Wavelets}
\label{app:wavelet_proof}

In the following, we recall and outline the complete proof for \cref{thm:mamba_wavelet}.

\begin{figure*}[b!]
\centering
\includegraphics[width=.9\linewidth]{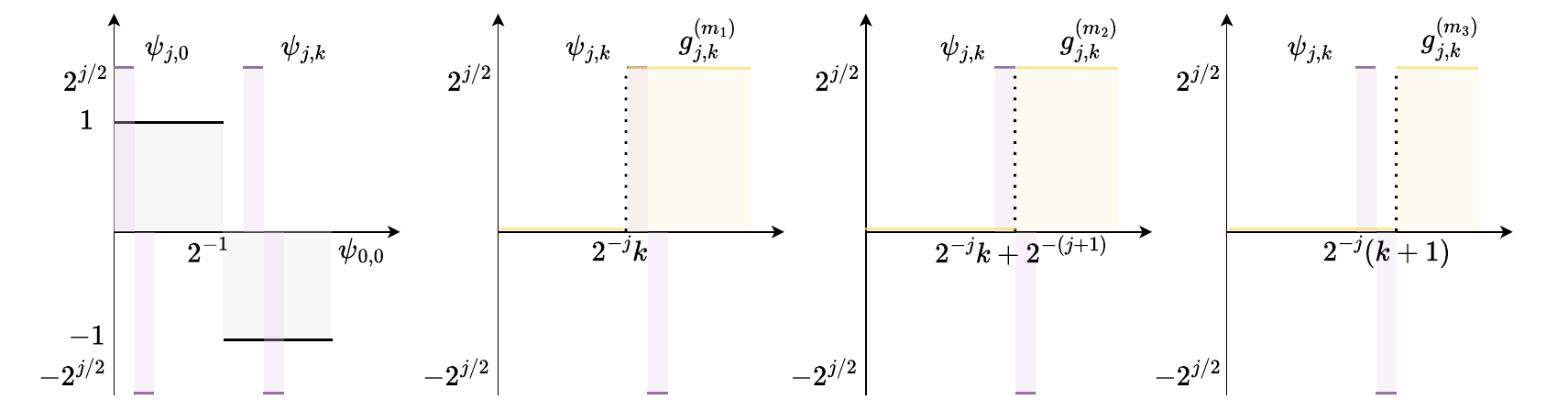}%
\caption{(Left) Example of Haar Wavelets; (Middle-Right) Shape of the $3$ Mamba basis $g_{j,k}^{\texttt{M}_1},g_{j,k}^{\texttt{M}_2},g_{j,k}^{\texttt{M}_3}$, whose linear combination can arbitrarily approximate the Haar wavelet $\psi_{j,k}$.}
\label{fig:proof_wavelet}%
\end{figure*}

\ThmOne*
\begin{proof}
The Haar wavelet at scale $j$ with translation $k$ is defined on the interval $s\in\left[0,1\right]$ as
\begin{align}
    \psi_{j,0}(s) &= \begin{cases}
        2^{j/2} & s \in [0, 2^{-(j+1)}) \\
        -2^{j/2} & s \in [ 2^{-(j+1)},  2^{-j}) \\
        0 & \text{otherwise}
    \end{cases}, \\
    \psi_{j,k}(s) &= \psi_{j,0}(s-2^{-j}k) = \begin{cases}
         2^{j/2} & s \in [2^{-j}k, 2^{-j}k+ 2^{-(j+1)}) \\
        -2^{j/2} & s \in [2^{-j}k + 2^{-(j+1)},  2^{-j}(k+1)) \\
        0 & \text{otherwise}
    \end{cases}.
\end{align}
See also \cref{fig:proof_wavelet} for an illustration.
Notice that each wavelet can be represented as a linear combination of three shifted Heaviside functions, namely:
\begin{equation}
    \psi_{j,k}(s) = 2^{\frac{j}{2}}\left(H\left(s-2^{-j}k\right) - 2H\left(s-\left(2^{-j}k+ 2^{-(j+1)}\right)\right) + H\left(s-2^{-j}(k+1)\right)\right),
    \label{eqn:wavelet_2_heavisides}
\end{equation}
where $H(s)=\mathbbm{1}_{s>0}$ denotes the Heaviside function. The goal is then to show that Mamba can indeed reproduce a shifted Heaviside as one of its basis functions $g_{j,k}^{\texttt{M}}$, by opportunely choosing its free parameters $\lambda_{j,k}, \Delta_{j,k} $ and $ B_{j,k}$ determining $g_{j,k}^{\texttt{M}}$.

To this end, we consider a 1D input. We set $\lambda_{j,k}=1$ and $B_{j,k}=2^{\frac{j}{2}}$, which reduces the Mamba basis function to
\begin{equation}
  g_{j,k}^{\texttt{M}}(s; 1, x) = 2^{\frac{j}{2}} e^{-\int_{s}^1 \Delta(x(r)) \,dr}.
  \label{eqn:mamba_basis_noBdt} 
\end{equation}
We then leave it to the last free parameter $\Delta(x(r))$ to perform most of the heavy-lifting. By extending the input signal $x$ to include the absolute time $t$ as an input, we can directly write $\Delta(x(t))=\Delta(t)$ (this requirement will be relaxed in \cref{prop:recover_time}). Notice that we can use the Mamba basis function to recover a (scaled) Heaviside centered at any given instant $\hat{t}_k$, provided $\Delta(t)$ behaves as follows:
\begin{equation}
    \text{}\Delta(s)\to\hat{\Delta}_k(s)=\begin{cases}
        \infty & \text{if } s<\hat{t}_k\\
        0 & \text{else}
    \end{cases}\quad\Longrightarrow\quad
  g_{j,k}^{\texttt{M}}(s; 1, x)=\begin{cases}
       0 & \text{if } s<\hat{t}_k\\
     2^{\frac{j}{2}} & \text{else}
    \end{cases}
  \to 2^{\frac{j}{2}}H(s-\hat{t}_k).
\end{equation}
Remember from \cref{eqn:mamba_discrete} that $\Delta(s)\equiv\Delta(s;b_\Delta,w_\Delta) \coloneqq \operatorname{SoftPlus}\left(w_{\Delta} s + b_{\Delta}\right)$. Note that 
\begin{equation}
 \lim_{x \to -\infty} \operatorname{Softplus}(x)=0, \qquad  \lim_{x \to \infty} \operatorname{Softplus}(x)=\infty.    
\end{equation}
Then, by choosing $b_{\Delta} = -w_{\Delta}\hat{t}_k$, and pushing $w_{\Delta}\to -\infty$, we can get arbitrarily close to approximating $\hat{\Delta}_k(s)$. Concretely, we have
\begin{equation}
  g_{j,k}^{\texttt{M}}(s; 1, x) \coloneqq 2^{\frac{j}{2}} e^{-\int_{s}^1 \Delta\left(r;b_\Delta=-w_\Delta \hat{t}_k,w_\Delta\right) \,dr}
  \;\xrightarrow{w_\Delta\to -\infty}\; 2^{\frac{j}{2}}H\left(s-\hat{t}_k\right).
  \label{eqn:mamba_basis_noBdt_limit} 
\end{equation}
By substituting $\hat{t}_k \in \{2^{-j}k, 2^{-j}k + 2^{-(j+1)}, 2^{-j}(k+1) \}$, the resulting mamba basis functions $g_{j,k}^{\texttt{M}_1}, g_{j,k}^{\texttt{M}_2}, g_{j,k}^{\texttt{M}_3}$ can approximate arbitrarily well $H(s - 2^{-j} k), H(s - (2^{-j}k + 2^{-(j+1)})), H( s - 2^{-j}(k+1) )$, respectively, which are precisely the shifted Heaviside functions appearing in \cref{eqn:wavelet_2_heavisides}. This allows us to finally write that $\forall \epsilon$, $\exists w_\Delta$ such that
\begin{equation}
    \left|\psi_{j,k}(t) - \left(g_{j,k}^{\texttt{M}_1} + g_{j,k}^{\texttt{M}_3} - 2 g_{j,k}^{\texttt{M}_2} \right) \right| < \epsilon.
\end{equation}
\end{proof}

Additionally, to remove the requirement of explicitly augmenting the Mamba input with time positional encoding, we show that Mamba can autonomously recover said time position by considering a constant input $x_t \equiv 1$ $\forall t$. This further relaxes the assumption in \cref{thm:mamba_wavelet} to using an additional Mamba layer receiving all-ones as input.
\begin{proposition}\label{prop:recover_time}
A $1$-layer Mamba can recover absolute time, given an all-ones input.
\end{proposition}
\begin{proof}
The proof follows by considering a Mamba layer \cref{eqn:mamba_basis} with $B\equiv 1, \lambda \equiv 0$. Substituting this into \cref{eqn:mamba_basis}, and providing an all-one input $x_t\equiv1$ $\forall t$ gives
\begin{equation}
    h^{\texttt{M}}(t) = \int_{0}^t x(s) ds = t.
\end{equation}
This applies analogously to the discrete view, $h^{\texttt{M}}_t = \sum_{s=1}^t x(s) = t.$
\end{proof}

Before proving \cref{cor:mamba_wavelet}, we recall the function approximation problem. Given a function $\rho \in L^2([0,1])$ (as a function of time $t \in [0, 1]$), and a set of bases parameterized by the SSM $\{g_n\}$ (e.g. Mamba or S4D), the goal is to project $\rho$ into the best size-$N$ basis $\mathcal{G}_N = \{g_1, \ldots, g_N\}$ that minimizes the approximation error,
\begin{equation}
    \argmin_{\mathcal{G}_N } \| \rho - \operatorname{proj}_{\mathcal{G}_N }(\rho) \|_{L^2} =    \argmin_{\mathcal{G}_N } \int_{0}^{1} \left( \rho(t) -   \operatorname{proj}_{\mathcal{G}_N }(\rho)(t) \right)^2 dt,\label{eqn:func_approx}
\end{equation}
where $N$ denotes the hidden state size and
\begin{equation}
   \operatorname{proj}_{\mathcal{G}_N }(\rho) \coloneqq \sum_{g_n \in \mathcal{G}_N} \langle \rho, g_n \rangle \, g_n.
\end{equation}
We note that the $N$ basis $\mathcal{G}_N$ is optimally chosen for the target function $\rho$ (instead of using a fixed set of $N$ bases independent of $\rho$). The inner products $\langle \rho, g_n \rangle = \int_{0}^t \rho(t) \overline{g_n(t)} dt$ for the various $n$ determine the coefficients of the projection of the target function $\rho$ onto the basis $\mathcal{G}_N$.

We are now ready to prove \cref{cor:mamba_wavelet}.

\CorOne*
\begin{proof}
 As discussed in \cref{sec:theory_approx}, a S4D basis function \cref{eqn:s4d_basis}  can represent any Fourier basis function $f_n = e^{i 2\pi n}$. When approximating a target function with discontinuities, the Fourier coefficients decay very slowly \citep{eckhoff1993accurate},
\begin{equation}
   \lim_{n \to \infty}  |\langle \rho, f_n \rangle| = O(n^{-1}).
\end{equation} 
This implies that using $N$ Fourier bases (and thus $N$ S4D basis functions) for approximating discontinuous functions yields an approximation error $O(N^{-1})$, proving the second part of \cref{cor:mamba_wavelet}. 

By \cref{thm:mamba_wavelet}, $N$ Mamba basis functions can approximate arbitrarily close any $N/3$ number of Haar wavelets. 
A set of $N/3$ Haar wavelets can achieve an approximation error of $O(2^{-N/3m})$ \citep{vetterli2001wavelets} when targeting a piecewise-constant function with $m \ge 1$ discontinuities. We provide a concrete argument for completeness. Suppose the function $\rho$ is piecewise constant with $m=1$ discontinuity. Then using an \emph{optimal} set of $N$ Haar wavelets $\mathcal{H}_N \subset \{\psi_{j,k} \}_{j \in \mathbb N, 0 \le k \le 2^j-1}$, we can achieve an approximation error of 
\begin{align}
  \min_{\mathcal{H}_N}   \| \rho - \operatorname{proj}_{\mathcal{H}_N}(\rho) \|_{L^2} &= \sum_{j =0}^{\infty} \sum_{k=0}^{2^j -1
} \langle \rho, \psi_{j,k} \rangle^2 - \sum_{\psi_{j,k} \in \mathcal{H}_N} \langle \rho, \psi_{j,k} \rangle^2 \\
    &=\sum_{j=0}^{\infty}  \langle \rho, \psi_{j,k(\rho)} \rangle^2 - \sum_{j=0}^{N-1} \langle \rho, \psi_{j,k(\rho)} \rangle^2  \\
    &= \sum_{j=N}^{\infty}  \langle \rho, \psi_{j,k(\rho)} \rangle^2 \\
    &= O(2^{-N}). \label{eqn:wavelet_adaptive_approx}
\end{align}
The second equality holds by noting that $\langle \rho, \psi_{j,k} \rangle = \int_{0}^{1}\rho(t) \psi_{j,k}(t) dt = 0$ if $\rho$ is constant in the interval $[2^{-j}(k-1), 2^{-j}k]$. Thus, for a piecewise-constant $\rho$ with one discontinuity, only one wavelet at each scale $j$ has $\langle \rho, \psi_{j,k(x)} \rangle \neq 0$. Given that we can choose adaptively the best $N$ wavelets, we pick the ones with nonzero coefficients across all $J=N$ scales. Then the approximation error is bounded the coefficient the highest scale, which has magnitude $O(2^{-N/2}$) and squared error $O(2^{-N})$. Similar analysis shows that for piecewise-constant functions with $m$ discontinuities, the optimal adaptive wavelet basis with $N$ wavelet functions achieves an approximation error of $O(2^{-N/m})$.    
\end{proof}

\begin{remark}
While the assumption of a piecewise-constant target function seems restrictive in \cref{cor:mamba_wavelet}, we note that this can be relaxed to a continuous target function $\rho$ by additionally assuming the input signals $x$ are piecewise-constant. The equivalence is due to the intermediate value theorem. Note that piecewise-constant input signals are ubiquitous in language modelling tasks, where $x_t$ takes values in a finite-size vocabulary.
\end{remark}

\subsection{Mamba can solve \keepn}
\label{app:keepn_proof}
Here we provide a direct application of the result from \cref{app:wavelet_proof}, to explain the ability of Mamba to exactly solve the \keepn task when equipped with Positional Encoding.

\CorKeepN*

\begin{proof}
To memorize the $n$-th position of a piecewise-constant signal $x(s)=\sum_{i=1}^t x_i\mathbbm{1}_{[i-1,i)}(s)$, it suffices to represent two Heaviside functions $h_1 = H(s-n), h_2 = H(s-(n-1))$ (recall $H(s-n) \coloneq \mathbbm{1}_{s > n}$), by noting that $h_1 - h_2$ is one at the interval $[n-1,n)$ and zero elsewhere, we have $\langle x, h_1 - h_2 \rangle = x_n$, as desired.

As shown in \cref{eqn:mamba_basis_noBdt_limit}, the Mamba basis function can represent any Heaviside function. Specifically, we can let $w_{\Delta} \to -\infty$, and
\begin{subequations}
    \begin{align}
        \Delta_1([x_t;t]) &= \softplus(w_{\Delta}t - w_{\Delta}n) = \begin{cases}
            \infty & t < n \\
            0 &t \ge n
        \end{cases}, \\
         \Delta_2([x_t;t]) &= \softplus(w_{\Delta}t - w_{\Delta}(n-1)) = \begin{cases}
            \infty & t < n-1 \\
            0 &t \ge n-1
        \end{cases}. 
    \end{align}
\end{subequations}
Suppose $\lambda \equiv 1, B = 1$ in the Mamba basis function \cref{eqn:mamba_basis}. Using the above $\Delta_1, \Delta_2$, we can obtain the Mamba basis functions representing the desired Heavisides, 
\begin{subequations}
    \begin{align}
     g_{1}^{\texttt{M}}(s; t, x) &= e^{-\int_{s}^t \Delta_1(x(r)) \,dr} \xrightarrow{w_\Delta\to -\infty} H(s-n).
  \label{eqn:mamba_basis_noBdt_n1} \\    
   g_{2}^{\texttt{M}}(s; t, x) &= e^{-\int_{s}^t \Delta_2(x(r)) \,dr} \xrightarrow{w_\Delta\to-\infty} H(s-(n-1)).
  \label{eqn:mamba_basis_noBdt_n2} 
    \end{align}
\end{subequations}

\end{proof}

\clearpage
\section{Sensitivity of Mamba}
\label{app:sensitivity_proof}
In this section, we prove \cref{lemma:exp_decay,lemma:constant}. These follow directly from the general sensitivity formula \cref{eqn:general_sensitivity_main}, which we derive in more details below:
Given a generic, input-dependent recurrence relationship as in \cref{eqn:disc_sol_general}, we have that the sensitivity of the state at time $t$ with respect to its input sequence at the $j$-th instant $x_j \in \R$ is given by
\begin{equation}
\begin{split}
    \frac{\partial \vh_t}{\partial x_j} &= \frac{\partial}{\partial x_j} \left(\sum_{s=1}^t \left(\prod_{r=s+1}^t \overline{\mLambda}_r\right) \overline{\mB}_s x_s\right)\\
     &=\sum_{s=1}^t \left[\frac{\partial}{\partial x_j}\left(\prod_{r=s+1}^t \overline{\mLambda}_r\right) \overline{\mB}_s x_s + \left(\prod_{r=s+1}^t \overline{\mLambda}_r\right) \frac{\partial}{\partial x_j}\left(\overline{\mB}_s x_s\right) \right]\\
    &=\sum_{s=1}^t \left[\frac{\partial \overline{\mLambda}_j}{\partial x_j}\left(\prod_{\substack{r=s+1,\\r\neq j}}^t \overline{\mLambda}_r\right) \overline{\mB}_s x_s \delta_{s<j} + \left(\prod_{r=s+1}^t \overline{\mLambda}_r\right) \frac{\partial}{\partial x_j}\left(\overline{\mB}_s x_s\right) \right]\\
    &=\left(\prod_{r=j+1}^t \overline{\mLambda}_r\right) \frac{\partial}{\partial x_j}\left(\overline{\mB}_jx_j\right) + \sum_{s=1}^{j-1} \frac{\partial \overline{\mLambda}_j}{\partial x_j}\left(\prod_{\substack{r=s+1,\\r\neq j}}^t \overline{\mLambda}_r\right) \overline{\mB}_s x_s\\
    &=\left(\prod_{r=j+1}^t \overline{\mLambda}_r\right) \left(\frac{\partial}{\partial x_j}\left(\overline{\mB}_jx_j\right) + \frac{\partial \overline{\mLambda}_j}{\partial x_j}\sum_{s=1}^{j-1} \left(\prod_{r=s+1}^{j-1} \overline{\mLambda}_r\right) \overline{\mB}_s x_s\right).\\
\end{split}
\label{eqn:general_sensitivity}
\end{equation}
Since $\overline{\mLambda}_r= -\operatorname{diag}([\overline{\lambda_1}(r), \ldots, \overline{\lambda}_n(r)])$ is diagonal and $\overline{\mB}_j = [\overline{B_1}(j), \ldots, \overline{B}_n(j)]$, we obtain from \cref{eqn:general_sensitivity} the sensitivity of the $n$-th component of the hidden state $\vh_{t,n} \equiv h_t$ with respect to the input $x_j$ as
\begin{equation}
  \frac{\partial h_t}{\partial x_j} = \left(\prod_{r=j+1}^t \overline{\lambda}_n(r) \right)  \left( \frac{\partial}{\partial x_j} (\overline{B}_n(j) \,  x_j) + \frac{\partial \overline{\lambda}_n(j)}{\partial x_j} \sum_{s=1}^{j-1} \left(\prod_{r=s+1}^{j-1} \overline{\lambda}_n(r) \right)\overline{B}_n(s) \, x_s \right).\label{eqn:sensitivity_component}
\end{equation}
With these ingredients, we are now ready to prove the main results in this section.

\LemSensty*
\begin{proof}
Recall the Mamba discretization \cref{eqn:zoh_disc} chooses input-dependent SSM parameters as
\begin{equation}
  \overline{\lambda}_n(r) = e^{-\lambda_n \Delta(x_r)}, \qquad \overline{B}_n(j) = B_n(x_j) \Delta(x_j) . \label{eqn:mamba_component}
\end{equation}

Substituting \cref{eqn:mamba_component}
to \cref{eqn:sensitivity_component}, we see that the first factor becomes
\begin{equation}
 \prod_{r=j+1}^t \overline{\lambda}_n(r)= e^{-\lambda_n \sum_{r=j+1}^t \Delta(x_r)},   \label{eqn:mamba_sensitivity_1}
\end{equation}
while the second factor becomes
\begin{align}
   & \frac{\partial}{\partial x_j} (\overline{B}_n(j) \,  x_j) + \frac{\partial \overline{\lambda}_n(r)}{\partial x_j} \sum_{s=1}^{j-1} \left(\prod_{r=s+1}^{j-1} \overline{\lambda}_n(r) \right)\overline{B}_n(s) \, x_s   \nonumber \\
  = & \, \frac{\partial}{\partial x_j} (B_n(x_j) \, \Delta(x_j) \, x_j) - \lambda  e^{-\lambda \Delta(x_j)}\frac{\partial\Delta(x_j)}{\partial x_j} \sum_{s=1}^{j-1} e^{-\lambda_n\sum_{r=s+1}^{j-1}\Delta(x_r)} B_n(x_s) \, \Delta(x_s)\,  x_s .\label{eqn:mamba_sense_second}
\end{align}

We are interested in the behavior of the sensitivity for $j$ fixed and $t\to\infty$ (i.e., the sensitivity of the current state with respect to early input in long-range sequences). Notice that the second factor does not scale with $t$ (since $j$ is fixed), and can be bound in terms of the parameters defining the transformations $\mB(x)$ and $\Delta(x)$, as well as $\lambda_n$ and $x$, as such:
\begin{equation}
\left|\cref{eqn:mamba_sense_second} \right| \le \tilde{c}(B_n, \Delta, \lambda_n, x_{\le j}).
\end{equation}
On the other hand, the first factor in \cref{eqn:mamba_sensitivity_1} shows in general \emph{exponential} dependence on $t$. Putting both together, we have
\begin{equation}
\begin{split}
    \left|\frac{\partial h_t^{\texttt{M}}}{\partial x_j}\right|
    \leq  \tilde{c}(\Delta, \lambda_n, B_n, x_{\le j})\,  e^{-\lambda_n\sum_{r=j+1}^t\Delta(x_r)} .
    \label{eqn:mamba_sensitivity}
\end{split}
\end{equation}

The proof for S4D is immediate by taking $\Delta(x_t) = 1, B_n(x_t) = B_n$ for all $t$, resulting in
\begin{equation}
\begin{split}
    \left|\frac{\partial h_t^{\texttt{S4D}}}{\partial x_j}\right|
    \leq   \tilde{c}(\lambda_n, B_n, x_{\le j})\,  e^{-\lambda_n (t-(j+1))} .
    \label{eqn:s4d_sensitivity}
\end{split}
\end{equation}

\end{proof}

\LemSenstyCst*
\begin{proof}
Recall for the scalar input sequence, each component of the Mamba hidden state is given by
\begin{equation}
    h_t^{\texttt{M}}   = \sum_{s=1}^t e^{-\lambda_n \sum_{r=s+1}^t \Delta(x_r)} \, B_n(x_s) \, \Delta(x_s) \, x_s. 
\end{equation}
Then the condition in \cref{eqn:ignore} implies
\begin{equation}
\lim_{t \to \infty} h_t^{\texttt{M}}  \ge \sum_{s=1}^t e^{-c}  B_n(x_s) \, \Delta(x_s) \, x_s.
\end{equation}
Straightforward computation of $\lim_{t \to \infty}\left|\frac{\partial h_t^{\texttt{M}}}{\partial x_j}\right|$ completes the proof.
\end{proof}

For comparison, we provide a similar sensitivity analysis for the softmax Attention layer as well.
\begin{proof}[Proof of Sensitivity of Softmax Attention]
\begin{equation}
\begin{split}
    \frac{\partial y_t^{\texttt{Attn}}}{\partial x_j} &= \frac{\partial}{\partial x_j}\left(\sum_{s=1}^{t} \frac{e^{x_t \mW_Q^{\top} \mW_K x_s}}{\sum_{r=1}^{t}e^{x_t \mW_Q^{\top} \mW_K x_r}} \mW_V x_s\right)\\
    &= \frac{e^{x_t \mW_Q^{\top} \mW_K x_j} (x_t \mW_Q^{\top} \mW_K \mW_V x_j + \mW_V)}{\sum_{r=1}^{t}e^{x_t \mW_Q^{\top} \mW_K x_r}}
    - \frac{e^{x_t \mW_Q^{\top} \mW_K x_j}x_t \mW_Q^{\top} \mW_K}{\left(\sum_{r=1}^{t}e^{x_t \mW_Q^{\top} \mW_K x_r}\right)^2}\sum_{s=1}^{t}e^{x_t \mW_Q^{\top} \mW_K x_s}\mW_V x_s\\
    &= \frac{e^{x_t \mW_Q^{\top} \mW_K x_j}}{\sum_{r=1}^{t}e^{x_t \mW_Q^{\top} \mW_K x_r}}\left(x_t \mW_Q^{\top} \mW_K \mW_V x_j + \mW_V
    - x_t \mW_Q^{\top} \mW_K\frac{\sum_{s=1}^{t}e^{x_t \mW_Q^{\top} \mW_K x_s}\mW_V x_s}{\sum_{r=1}^{t}e^{x_t \mW_Q^{\top} \mW_K x_r}}\right)\\
    &= \frac{e^{x_t \mW_Q^{\top} \mW_K x_j}}{\sum_{r=1}^{t}e^{x_t \mW_Q^{\top} \mW_K x_r}}\left(\mW_V + x_t \mW_Q^{\top} \mW_K \left(\mW_V x_j-\frac{\sum_{s=1}^{t}e^{x_t \mW_Q^{\top} \mW_K x_s}\mW_V x_s}{\sum_{r=1}^{t}e^{x_t \mW_Q^{\top} \mW_K x_r}}\right)\right).
\end{split}
\end{equation}
Even in this case, the second term can be bound by a constant $C$ in terms of $\|\mW_Q\|,\|\mW_K\|,\|\mW_V\|$ and $\|\vx\|$ (notice that the rightmost term, where a fraction of two sums over $t$ appear, behaves as a weighted average of $\mW_V x_s$, and hence does \emph{not} scale with $t$). Only the denominator at the first factor remains as a function of $t$, providing
\begin{equation}
    \left|\frac{\partial h_t^{\texttt{Attn}}}{\partial x_j}\right| \leq C \frac{1}{t} \left(\min_r e^{x_t\mW_Q^\top\mW_K x_r}\right)^{-1}.
\end{equation}

\end{proof}

\clearpage

\section{Proofs of \Cref{sec:architecture}} \label{sec:task_proofs}

\subsection{Proofs of Mamba Solving the \mqar Task} \label{app.proofs_mqar}

\paragraph{The \mqar Task}
As a reminder, the \mqar task receives an input in the form
\begin{equation}
    \vx = [\underbrace{
    k_1\;\; v_1\;\;\ldots\;\;
    k_i\;\; v_i\;\;\ldots\;\;
    k_{\kappa}\;\; v_{\kappa}}_{\kappa\text{ key-value pairs}}\;|\; \underbrace{v_\times\;\;\ldots\;\; v_\times\;\;
    k_{i_1}\;\; v_\times\;\;\ldots\;\; v_\times\;\;
    k_{i_j}\;\; v_\times\;\;\ldots\;\; v_\times\;\; k_{i_{\kappa}}\;\; v_\times\;\;\ldots}_{\text{shuffled queries (=keys), interwoven with noise}} ] \in\mathbb{R}^T,
\end{equation}
where the keys $k_i$ are randomly taken from a key set of size $\kappa$, and the values $v_i$ and the various noise tokens $v_\times$ are randomly taken from a vocabulary of size $|V|$. The goal is to output a sequence that, at the location of each query, reports the value matching the corresponding key, that is
\begin{equation}
    \vy = [\times\;\;\ldots\;\;\times|
    \times\;\;\ldots\;\;\times\;\;
    v_{i_1}\;\;\times\;\;\ldots\;\;\times\;\;
    v_{i_j}\;\;\times\;\;\ldots\;\;\times\;\;
    v_{i_\kappa}\;\;\times\;\;\ldots\;\;\times],
\end{equation}
while the other components of the output (denoted with $\times$) are ignored. 

For the rest of the proofs, we often make use of the Johnson-Lindenstrauss (JL) Lemma to reduce the embedding dimensionality. For completeness, we state and prove JL lemma below.
\begin{lemma}[JL Lemma] \label{JL:inner-prod}
Given a set of standard bases $\{\ve_1, \ldots, \ve_d \}$ and $\epsilon \in (0, 0.5)$, there exists a random projection matrix $M: \R^d \to \R^p$ where $p=O(\epsilon^{-2} \log d), M_{i,j} \overset{i.i.d.}{\sim} \frac{1}{\sqrt{p}}\operatorname{Unif}\{\pm1\}$ such that for all pairs $(i,j)$,
\begin{equation}
   | \langle M \ve_i, M \ve_j \rangle| \le \epsilon. \label{eqn:JL_eps}
\end{equation}
\end{lemma}
\begin{proof}
Let $\sqrt{p} M_{i,j} \coloneqq Z_{i,j}$, where $Z_{i,j}$ is the symmetric Bernoulli variable. For any $i,j \in [d]$, we have
\begin{align*}
  \langle M \ve_i, M \ve_j \rangle =  \langle M_{[:,i]}, M_{[:,j]} \rangle = \frac{1}{p} \sum_{l=1}^p Z_{l,i} Z_{l,j} 
  \equiv  \frac{1}{p} \sum_{l=1}^p Z_{l},
\end{align*}
where the last equality follows from the fact that the product of two independent Bernoulli variables is Bernoulli. In other words, the dot product of the two projected vectors is a sum of $k$ i.i.d. symmetric Bernoullis. By Hoeffding's inequality (e.g. \citet[Theorem 2.2.2]{vershynin2018high}), 
\begin{equation}
    \Pr{\frac{1}{p}\sum_{l=1}^p Z_l \ge \epsilon} \le \exp{(-\frac{p\epsilon^2}{2})}.
\end{equation}
Thus
\begin{equation}
\Pr{ | \langle M \ve_i, M \ve_j \rangle| \ge \epsilon}   = 2  \Pr{\frac{1}{p}\sum_{l=1}^p Z_l \ge \epsilon} \le 2 \exp{(-\frac{p\epsilon^2}{2})}. \label{eqn:prob_ij}
\end{equation}
Therefore, except with probability less than $2 \exp{(-\frac{p\epsilon^2}{2})}$, it holds that $| \langle M \ve_i, M \ve_j \rangle| \le \epsilon$.
Let $p = \frac{4}{\epsilon^2} \ln \frac{d}{\delta}$ for some $\delta \in (0,1)$. By a union bound, this holds for all $\binom{d}{2}$ pairs of $(\ve_i, \ve_j)$ except with probability
\begin{equation}
    \binom{d}{2} \, 2 \exp{(-\frac{p\epsilon^2}{2})} <  d^2\exp{(-\frac{p\epsilon^2}{2})} =  \exp( - 2 \ln \frac{d}{\delta} ) = \delta^2.
\end{equation}
Since there is a probability grater than $1 - \delta^2$ that $| \langle M \ve_i, M \ve_j \rangle| \le \epsilon$ holds for all $i,j$, this guarantees the existence of such $M$ by the probabilistic method. 

\end{proof}

\LemMambaNogate*
\begin{proof}
The proof is divided into two steps: We first construct a $1$-layer Mamba model without gating that solves \mqar using standard basis vectors $d=O(\kappa+|V|)$; based on such construction, we then apply the JL Lemma (\cref{JL:inner-prod}) together with a shifting trick to complete the proof for $d=O(\kappa + \log |V|)$. 

\textbf{Step 1: Construction With $d=(\kappa+O(|V|)$. } 
We consider one-hot encoding of keys and values. The main idea is to (i) use the size-$2$ convolution to combine key-value pairs and filter out other uninformative pairs (particularly, preserve key-value pairs and discard value-value pairs); (ii) use appropriate input-dependent SSM matrices to store and retrieve the key-value information in the hidden state; (iii) use the output layer to project to the value embedding subspace. Crucially, we observe that the hidden state of Mamba at time $t$ is a $d \times N$ matrix \cref{eqn:recurrence_compare}. Thus, we leverage this matrix structure such that each column of the state corresponds to a given key ($N = \kappa$), and holds the value associated with it, that is:
\begin{equation}
    \boldsymbol{h} =
    [\quad\boldsymbol{v}_{j_1}\quad|
    \quad\boldsymbol{v}_{j_2}\quad
    |\quad\dots\quad|
    \quad\boldsymbol{v}_{j_\kappa}\quad]\in\mathbb{R}^{d\times N }.
    \label{eqn:rough_h_mamba_mqar}
\end{equation}
We first describe how our proposed solution operates, and then prove that indeed our solution solves the \mqar task exactly.
\begin{itemize}
    \item \textbf{Embedding} The task of the embedding layer is to clearly distinguish values and keys. We achieve this by letting it map to orthogonal directions: let the embedding dimension be $d = |V|+\kappa$, and $\ve_{i} \in \R^d$ denote the standard basis vector. We impose
    \begin{equation}
        k_i \mapsto \vk_i \coloneqq k \cdot \ve_{i}, \quad\text{and}\quad v_i \mapsto \vv_i \coloneqq v\cdot\ve_{\kappa+i}
    \label{eqn:embedding_mamba_mqar}
    \end{equation}
    for some parameters $k,v>0$.
    
    \item \textbf{Convolution} We use a size-$2$ convolution to combine information of a key-value pair, as it is essential to the task solution.
    We remind that for a sequence $(\ldots, \vx_{t-1}, \vx_{t}, \ldots)$, the action of the convolution layer with size-$2$ kernel with left padding $\vx_0$ is given by
    \begin{equation}
        \hat{\vx}_{t}=\operatorname{conv}(\vx_{t-1}, \vx_{t}) \coloneqq\sigma\left(\boldsymbol{c}_0\odot\boldsymbol{x}_{t-1} + \boldsymbol{c}_1\odot\boldsymbol{x}_{t} - \boldsymbol{b}\right), \qquad \vx_0 \coloneqq \mathbf{0} ,\label{eqn:size2-conv}
    \end{equation}
    for certain parameters $\boldsymbol{c}_0,\boldsymbol{c}_1,\boldsymbol{b}\in\mathbb{R}^d$ and a nonlinearity\footnote{In the original Mamba definition, we have $\sigma\equiv\text{SiLU}$; for ease of illustration we consider instead $\sigma\equiv\text{ReLU}$, but notice that similar considerations still hold in this case, in light of the fact that $\text{SiLU}(x)\to\text{ReLU}(x)$ for $x\to\pm\infty$: it suffices then to opportunely scale the inputs.} $\sigma = \text{ReLU}$. For our construction, it suffices to pick $\boldsymbol{c}_0=c_0\boldsymbol{1}, \boldsymbol{c}_1=c_1\boldsymbol{1}$, and $\boldsymbol{b}=b\boldsymbol{1}$, with $c_0,c_1,b>0$.
        We want such nonlinear convolution to preserve information from the $(\vk_i,\vv_j)$ pairs, so we impose $c_0k>b$ and $c_1v>b$. We also need to prevent the value in $(\vv_i, \vk_j)$ from getting associated to the wrong key, but we want to preserve key information to be able to extract it at retrieval time, so we ask $c_0v \le b$ and  $c_1k > b$. The $(\vk_i, \vk_j)$ pair represents an edge-case, and we will show later how to deal with this.
    Finally, we want to ignore contributions from $(\vv_i,\vv_j)$ pairs (as they refer to ``noise'' tokens), but we will delegate this task to input-selectivity in the SSM layer.
    To summarize, we need to satisfy
    \begin{equation}
        c_0 k > b, \qquad c_1 v > b, \qquad c_1 k > b, \qquad c_0 v \le b. \label{eqn:conv_constraints}
    \end{equation}
    A feasible parameter combination satisfying \cref{eqn:conv_constraints} is given by:
    \begin{equation}
        v=1,\qquad k=2,\qquad c_0=1,\qquad c_1=2,\quad\text{and}\quad b=1.
        \label{eqn:parameters_mamba_mqar}
    \end{equation}
    
    Let us show how such nonlinear convolution acts on the four different types of input pairs, under the assumptions \cref{eqn:conv_constraints}:
    \begin{subequations}
      \begin{align}
       \operatorname{conv}((\vk_i, \vv_j)) & = \text{ReLU}( c_0 \vk_i + c_1 \vv_j - \vb) = (c_0k-b) \ve_{i} + (c_1v-b) \ve_{\kappa+j} \label{eqn:mamba_conv_kv} \\
        \operatorname{conv}((\vv_i, \vk_j)) & = \text{ReLU}( c_0 \vv_{i} + c_1 \vk_j - \vb) = (c_1k-b) \ve_{j} \label{eqn:mamba_conv_vk} \\
         \operatorname{conv}((\vk_i, \vk_j)) & = \text{ReLU}(c_0 \vk_i + c_1 \vk_j - \vb) = (c_0k-b) \ve_{i} + (c_1k-b) \ve_{j} \label{eqn:mamba_conv_kk} \\
          \operatorname{conv}((\vv_i, \vv_j)) & = \text{ReLU}(c_0 \vv_i + c_1 \vv_j - \vb) = (c_1v-b) \ve_{\kappa+j}.
      \end{align}  
    \end{subequations}

    \item \textbf{Mamba SSM} The SSM layer organizes a hidden state \emph{matrix} where its columns are indexed by the keys and store information of the associated values. To this end, we let
    \begin{equation}
     \mLambda = \boldsymbol{0}, \quad \Delta_t = \mathbf{1}, \quad \mB(\vx) = \mC(\vx) = [I_{\kappa} |  \boldsymbol{0}_{|V|}] \vx \equiv W_k \vx,
     \label{eqn:mamba_ssm_params}
    \end{equation}
    where $W_k$ projects the input to the key embedding subspace.
    Consequently, the SSM layer covers three roles at once:
    \begin{enumerate}
        \item Ensure that the right key get associated to the right column in the hidden state. This role is covered by the input matrix $\mB(\vx) = W_B\vx$. By picking $W_B= W_k = [I_\kappa | \boldsymbol{0}_{|V|}]$, we can see that only keys in the convolved input pair are used for populating the hidden state. This choice of the input matrix also ensures that information from $(\vv_i,\vv_j)$ pairs does not affect the hidden state.
        \item Propagate information down the sequence without corrupting it (i.e., memorization). The state matrix $A(\vx)$ can take care of this, provided we fix it to the all-one matrix. This can be achieved by prescribing $\mLambda \equiv \boldsymbol{0}$ in \cref{eqn:size_compare}. Notice that with this choice, the SSM layer simply performs a cumulative sum $ \boldsymbol{h}_t = \sum_{s=1}^t \hat{\vx}_s \otimes \mB_s$.
        \item Retrieve the correct column when needed. This task is taken care of by the output matrix $C(\vx) = W_C\vx$. As with the input matrix, it suffices to have $W_C=W_k=[I_\kappa | \boldsymbol{0}_{|V|}]$: also in this case, when an input containing a key is encountered, the $C(\vx)$ matrix will retrieve information only from the column corresponding to that specific key.
    \end{enumerate}
    
    \item \textbf{Output} The final output layer simply needs to correctly classify $\vv_i$ as the most likely value from the retrieved vector $\boldsymbol{y}_t$. To this end, it suffices to consider
    \begin{equation}
        W_o= \left[\begin{array}{c}
             \boldsymbol{v}_1^{\top} \\
             \vdots\\
             \boldsymbol{v}_{|V|}^{\top}\\
        \end{array}\right],
    \end{equation}
    as $\vv_i$ will return the maximum scalar product with $\boldsymbol{y}_t$.

\end{itemize}
Let us illustrate with a step-by-step example how the above construction yields the required result. Consider a generic input sequence $\vx$ for the \mqar task. Under the embedding layer prescribed in \cref{eqn:embedding_mamba_mqar}, the (embeded) input will be in the form
\begin{equation}
\begin{split}
    \vx=\left[\begin{array}{ccccccc|cccccc}
         \vk_{i_1} & \vv_{j(i_1)} & \vk_{i_2} & \vv_{j(i_2)} & \dots & \vk_{i_\kappa} & \vv_{j(i_\kappa)} & \vv_{\times} & \dots & \vk_{k_1} & \dots & \vk_{k_\kappa} & \dots
    \end{array}
    \right]\in\mathbb{R}^{d\times T},
\end{split}
\end{equation}
so that the $i_m$-th key is associated to the $j(i_m)$-th value. The order in which the keys appear at query time is also random and denoted by the $k_m$ subscript. Moreover, keys at query time are randomly interwoven with value tokens, denoted as  $\vv_{\times}$.
After convolution, this input gets mapped to
\begin{equation}
\begin{split}
    \hat{\vx}=\left[\begin{array}{ccccccc|cccccc}
         c_{1,k}\ve_{i_1} & c_{0,k}\ve_{i_1} & c_{1,k}\ve_{i_2} & c_{0,k}\ve_{i_2} & \dots & c_{1,k}\ve_{i_\kappa} & c_{0,k}\ve_{i_\kappa} & \boldsymbol{0} & \dots & c_{1,k}\ve_{k_1} & \dots & c_{1,k}\ve_{k_\kappa} & \dots\\
         + & + & + & + &  & + & + & + &  & + &  & + & \\
         \boldsymbol{0} & c_{1,v}\ve_{j(i_1)} & \boldsymbol{0} & c_{1,v}\ve_{j(i_2)} & \dots & \boldsymbol{0} & c_{1,v}\ve_{j(i_\kappa)} & \boldsymbol{0} & \dots & \boldsymbol{0} & \dots & \boldsymbol{0} & \dots
    \end{array}
    \right],
\end{split}
\end{equation}
where we denote $c_{0,k}=c_0k-b$, $c_{1,k}=c_1k-b$, $c_{1,v}=c_1v-b$ to slim notation, and separate the components of $\hat{\vx}$ into key-related (top) and value-related (bottom) for illustrative purposes. Notice how, for the initial part of the input, distinct keys (the various $\ve_{i_m}$ at the top) are only associated with their respective values (the various $\ve_{j(i_m)}$ at the bottom), or with $\boldsymbol{0}$.
Before moving on to the SSM layer, let us remind its action \cref{eqn:recurrence_compare}:
\begin{equation}
\begin{split}
    & \vy_t = \left( e^{\mLambda\odot(\Delta_t\otimes\bm{1}_N)} \odot \vh_{t-1} + (\Delta_t \odot \hat{\vx}_t) \otimes \mB_t \right) \, \mC_t = \sum_{s=1}^t \left(\prod_{m=s}^{t-1} e^{\mLambda\odot(\Delta_m \otimes\bm{1}_N)} \odot (\hat{\vx}_{s} \cdot \mB_s^{\top})\right) \,  \mC_t\\
    \xRightarrow{\mLambda\equiv \mathbf{0}, \Delta_m \equiv \mathbf{1}}\quad &\vy_t = \left(\vh_{t-1} + \hat{\vx}_t \cdot \mB_s^{\top}\right) \, \mC_t = \sum_{s=1}^t \left(\hat{\vx}_{s} \cdot \mB_s^{\top}\right) \, \mC_t.
\end{split}
\end{equation}
Notice how the hidden state naturally admits a matrix structure $\vh_t\in\mathbb{R}^{d\times N}$ due to the outer product $ \vx_s \cdot \mB_s^{\top}$, whose columns are updated by $\vx_t$, where $\mB_s$ determines which columns are affected. If $\mB_t \propto \ve_i$ (i.e., only nonzero at component $i$), then $\vx_t$ will contribute to only the $i$-th column in $\vh_t$. Similarly, taking $\mC_t \propto \ve_i$ ensures that only the $i$-th column from the hidden state is retrieved. This is precisely what we achieve with the choice outlined above. In light of this, the hidden state after the initial part of the input (where the key-value pairs are listed) admits the following form at time $2\kappa$
\begin{equation}
    \vh_{2\kappa} =\left[\begin{array}{c|c|c|c}
        \hat{k}\ve_{1} & \hat{k}\ve_{2} & \dots & \hat{k}\ve_{\kappa}\\
        + & + &  & +\\
        \hat{v}\ve_{j(1)} & \hat{v}\ve_{j(2)} & \dots & \hat{v}\ve_{j(\kappa)}
    \end{array}\right]
    ,
    \qquad\text{with}\quad \hat{k}=c_{1,k}^2+c_{0,k}^2,\quad \hat{v}=c_{1,v}c_{0,k}, \label{eqn:thm2_proof_hidden_coeef}
\end{equation}
and remains unchanged until the first key $\vk_{k_1}$ is encountered at query time, as $\mB(\vx) = \mB((\vv_i, \vv_i)) =\boldsymbol{0}$ for all the tokens in between. For each key $\vk_{i}$ encountered at query time, $\mC(\vx)$ then takes care of extracting the corresponding $i$-th column of $\vh$, which holds a vector proportional to the embedding of its associated value, $\ve_{j(i)}$. Finally, the output matrix computes scalar products of all the values embedding with the extracted vector, which will then be maximum for $\ve_{j(i)}$, thus accurately solving the \mqar task.

There is an edge case to this construction, which occurs when, at retrieval time, two keys appear adjacent to each other in a key-key pair $(\vk_{i},\vk_{l})$. In this case, $\vk_l$ represents the actual query, while $\vk_i$ acts as noise: the convolution will in fact contain information from two keys, implying that $C(\vx)$ will have two nonzero components at that point. However, with the correct parameter choice, we can have the query information dominate the noise, and still recover the desired solution. We have by \cref{eqn:mamba_conv_kk}
\begin{equation}
    \mC(\operatorname{conv}(\vk_{i},\vk_{l}))  = (c_0 k - b) \ve_{i} + (c_1 k - b) \ve_{l}.
\end{equation}
When tested against the hidden state at that instant, then, the output matrix will return a linear combination of the values associated to the two keys:
\begin{equation}
 \mC(\operatorname{conv}(\vk_{i},\vk_{l})) \,  \vh_t  = 
(c_0 k - b)  \left(\hat{k} \ve_{i}  + \hat{v} \ve_{j(i)} \right) + 
      (c_1 k - b) \left(\hat{k} \ve_{l}  + \hat{v} \ve_{j(l)} \right).
\end{equation}
To ensure the key-value pair of $\vk_l$ dominates that of $\vk_i$, we need to impose
\begin{equation}
   c_{0} k - b \ll c_{1} k - b
   \quad\Longleftarrow\quad
   c_0 \ll c_1,
\end{equation}
which is already satisfied by the parameter choice \cref{eqn:parameters_mamba_mqar}.

\textbf{Step 2: Dimensionality Reduction to $d=O(\kappa+\log |V|)$. } Having proved the construction with $d=O(\kappa+|V|)$, we now apply JL Lemma to reduce the embedding dimension and suitably adjust the Mamba architecture weights to ensure the desired output. Concretely:

\begin{itemize}
    \item \textbf{Embedding} We use the same key embeddings as above $\{\ve_1, \ldots, \ve_{\kappa} \}$ while reducing the value embedding dimensionality to satisfy almost-orthogonality. By JL Lemma (c.f.~\cref{JL:inner-prod}), given the value embeddings $\{ \ve_j\}_{j=1+\kappa}^{|V|+ \kappa}$ and $\epsilon \in (0, 0.5)$, there exists $M_v: \R^{|V|} \to \R^{p}$ where $p = O(\log |V|)$ and each of its entry is in $\{-\frac{1}{\sqrt{p}}, \frac{1}{\sqrt{p}}\}$ such that $\langle M_v \ve_{j_1}, M_v \ve_{j_2}\rangle \le \epsilon $ for any $j_1 \neq j_2$. Let $M \coloneqq \mI_{\kappa} \oplus M_v \in \R^{(\kappa + p) \times (\kappa + |V|)}$ be the direct sum of the identity matrix preserving the one-hot key embeddings and the JL matrix projecting the value embeddings. Let the normalized value embedding be $\bar{\vv}_j \coloneqq M \ve_{j}$ Let the \emph{shifted} value embedding be
    \begin{equation}
         \vv_j \coloneqq M \ve_{j} + \beta [\mathbf{0}_{\kappa} | \mathbf{1}_p]^{\top} = \sum_{i=1}^p (M_{i+\kappa,j} + \beta) \ve_{i+\kappa}\in \R^{\kappa+p},
    \end{equation}
   where each component of $\vv_j$ is in the  range $[-\frac{1}{\sqrt{p}}+\beta, \frac{1}{\sqrt{p}}+\beta] \equiv [v_{\min}, v_{\max}]$. Note that by letting $\beta > - \frac{1}{\sqrt{p}}$, we can ensure all components of $\vv_j$ are nonnegative. 
    \item \textbf{Convolution} We use a size-$2$ convolution as \cref{eqn:size2-conv}, to retain the value from $(\vk_i, \vv_j)$ pairs and discard the value from $(\vv_i, \vk_j)$ pairs. Ideally we want 
    \begin{subequations}
        \begin{align}
            \operatorname{conv}(\vk_i, \vv_j) &\coloneqq \operatorname{ReLU}(c_0 \vk_i + c_1 \vv_j - \vb) 
           \propto (c_0 - b)\vk_i + (c_1-b) \vv_j \\
            \operatorname{conv}(\vv_i, \vk_j) &\coloneqq \operatorname{ReLU}(c_0 \vv_i + c_1 \vk_j - \vb) \perp \operatorname{span}(\{\vv_1, \ldots, \vv_{|V|}\}) . 
        \end{align} \label{eqn:conv_desire}
    \end{subequations}
Given the shifted value embeddings, we must impose the following constraints to achieve \cref{eqn:conv_desire}:
    \begin{equation}
        \qquad c_0 k > b, \qquad c_1 v_{\min} > b, \qquad c_0 v_{\max} \le b. \label{eqn:conv_constraint_log}
    \end{equation}
    A feasible parameter combination satisfying \cref{eqn:conv_constraint_log} is given by
   \begin{equation}
        \beta = 2 \, (\implies [v_{\min} , v_{\max}] \subseteq [1,3] \text{ since } p \ge 1), \qquad k=10,\qquad c_0=1,\qquad c_1=10,\quad\text{and}\quad b=3.
        \label{eqn:parameters_mamba_mqar_log}
    \end{equation}
    Consequently, we have the desired convolution outputs
    \begin{subequations}
        \begin{align}
            \operatorname{conv}(\vk_i, \vv_j)
          &= (c_0 k-b) \ve_i + \sum_{i=1}^p \left(c_1 (M_{i+\kappa,j} +\beta) - b \right) \ve_{i + \kappa} = (c_0 k - b) \ve_{i} + c_1 \bar{\vv}_j + \underbrace{(c_1\beta - b) \sum_{i=1}^p \ve_{i+\kappa}}_{\coloneqq \vs} \\
            \operatorname{conv}(\vv_i, \vk_j)
            &=  (c_1 k - b) \ve_j. 
        \end{align} \label{eqn:conv_plugin}
    \end{subequations}
    Note that the edge case for $\operatorname{conv}(\vk_i, \vk_j)$ is taken care of since $c_0 \ll c_1$, while the $(\vv_i, \vv_j)$ pairs will be ignored in the SSM layer, following the same argument as in step 1.
    \item \textbf{Mamba SSM} The SSM layer is the same as in step 1 \cref{eqn:mamba_ssm_params}, propagating information through the hidden state (c.f.~\cref{eqn:thm2_proof_hidden_coeef}). At the query time for key $\vk_i, i=1, \ldots, \kappa$, the output $\vy_t =\left( (c_0 k - b)^2 + (c_1 k - b)^2 \right)\ve_{i} + (c_0 k - b) \left(c_1 \bar{\vv}_{j(i)} + \vs \right)$ contains the (scaled) normalized value embedding shifted by a constant vector $\vs$. 
    \item \textbf{Output} The output layer undoes the shift and classifies based on the normalized value embeddings. Recall $M_v \in \R^{|V| \times p}$ is the JL matrix from the value embedding projection.
    We set the output linear layer with the weight matrix $W_o = [\mathbf{0} | M_v^{\top}] \in \R^{|V| \times (\kappa + p)}$ and the bias vector $\vb_o = - (c_0k - b) W_o  \vs$. The final output is given by
    \begin{equation}
   W_o \vy_t + \vb_o =   (c_0k - b) W_o \left( c_1 \bar{\vv}_{j(i)} + \vs \right) -  (c_0k - b) W_o\vs =  (c_0k - b) c_1 M^{\top}_v M_v \ve_{j(i)},
    \end{equation}
     which yields the maximum at component $j(i)$ since $\langle M_v \ve_{j(i)}, M_v \ve_{l} \rangle \le \epsilon$ for any $l \neq j(i)$ by JL Lemma. This completes the proof. 
\end{itemize}

\end{proof}

\LemMambaTwo*
\begin{proof}
The idea is to execute the Mamba solution outlined in the proof of  \cref{lemma:mamba_nogate}, but in a \emph{leaner} manner due to the additional degrees of freedom in \MM: particularly, we leverage the fact that the convolutions for the value \cref{eqn:mamba_conv}, key \cref{eqn:mamba_compute_B}, and query \cref{eqn:mamba_compute_C} can be chosen independently (instead of using the same convolution in Mamba). The proof is divided into two steps: We first present the construction using standard basis vectors with $d = \kappa + |V|$; We then reduce the embedding dimension by applying JL lemma.

\textbf{Step 1: Construction With $d=O(\kappa+|V|)$. }
\begin{itemize}
    \item \textbf{Embedding - Same as Mamba} The role of the embedding layer is to clearly distinguish values and keys. We achieve this by letting it map to independent directions: let the embedding dimension be $d = |V|+\kappa$, and $\ve_{i} \in \R^d$ denote the standard basis vector. We let
    \begin{equation}
        k_i \mapsto \vk_i \coloneqq \ve_{i}, \quad\text{and}\quad v_i \mapsto \vv_i \coloneqq \ve_{\kappa+i}.
    \label{eqn:embedding_mamba2_mqar}
    \end{equation}
    \item \textbf{Convolution} Differently from Mamba that uses the same convolution kernel to compute the SSM input $\vu$ and parameters $\mB, \mC$, \MM has the additional flexibility of using three independent convolutions for computing $\vu, \mB, \mC$ (see details in \cref{tab:compare_mambas}). We now exploit this flexibility by setting:
    \begin{equation}
        \begin{split}
            \hat{\vx}_t^B&=\operatorname{conv}_B(\vx_{t-1}, \vx_t) = \sigma( \vc_0 \odot \vx_{t-1} + \vc_1 \odot \vx_t - \vb) \coloneqq \vx_{t-1},\\
            \hat{\vx}_t&=\operatorname{conv}_u(\vx_{t-1},\vx_{t}) \coloneqq \vx_{t},\\
            \hat{\vx}_t^C&=\operatorname{conv}_C(\vx_{t-1},\vx_{t}) \coloneqq \vx_t. 
        \end{split}
        \label{eqn:mamba2_conv}
    \end{equation}
    Consequently, the output from $\operatorname{conv}_B$ \emph{shifts} the input sequence to the right by one position, whereas the outputs from $\operatorname{conv}_u, \operatorname{conv}_C$ are the same as the input. 
    \item \textbf{\MM SSM - Same as Mamba} The role of the SSM layer is to associate key-value pairs, propagate information through the state, and retrieve the correct value given a query(=key). Unlike \cref{lemma:mamba_nogate}, the convolved input \cref{eqn:mamba2_conv} contains only key or value information but never mixes them. A simple choice is to set $\mB, \mC$ as the identity matrices, but this requires the state size be the same as the embedding size $N = d$. To further reduce the state size to $N = \kappa$, we use the same choice as Mamba \cref{eqn:mamba_ssm_params} by letting 
    \begin{equation}
     \bm{\lambda} = \boldsymbol{0}, \mB(\vx_t) =  \mC(\vx_t) = [\mI_{\kappa} |  \boldsymbol{0}_{|V|}] \vx_t \equiv W_k \vx_t,   
    \end{equation}
    \item \textbf{Output - Same as Mamba} Even in this case, as a classifier it suffices to pick
    \begin{equation}
        W_o= \left[\begin{array}{c}
             \boldsymbol{v}_1^{\top} \\
             \vdots\\
             \boldsymbol{v}_{|V|}^{\top}\\
        \end{array}\right].
    \end{equation}
\end{itemize}
With the definitions above, we can simplify the outcome of the \MM layer application. This in fact reduces to
\begin{equation}
    \vy_t = \sum_{s=1}^t \left(\hat{\vx}_s\,\mB( \hat{\vx}_s^B)^{\top} \right) \, \mC(\hat{\vx}_s^C)
    = \sum_{s=1}^t\left( \vx_{s} \, (W_k\vx_{s-1})^{\top} \right) W_k \vx_{t} = \sum_{s=1}^t \vx_s (\vx_{s-1}^{\top} W_k^{\top}W_k \vx_t) 
    \label{eqn:mamba2_app}
\end{equation}
where $\vx_{0} = \boldsymbol{0}$ is the zero padding vector. Note that the output $\vy_t$ for $t \ge 2\kappa$ contains the desired key-value association, $ \vx_s (W_k \vx_{s-1})^{\top} =  \vv_{j(i)} \vk_i^{\top}$ for $s = 2, 4, \ldots, 2\kappa$.

At query time where $\vx_t = \vk_i$, thanks to our construction of the Embedding layer, $\left\langle\vk_{i},\vx_{s-1}\right\rangle = 0$ for all $s$, \emph{except} when $\vx_{s-1}=\vk_{i}$: in that case we have instead $\left\langle\vk_{i},\vx_{s-1}\right\rangle=1$. 
Whatever the index $s-1$ at which this occurs, the associated value $\vx_s$ is precisely the one we are seeking, i.e., the value embedding $\vv_{j(i)}$ immediately following the key matching the corresponding query $\vk_i$. Thus, $\vy_t = \vv_{j(i)}$. Now, applying the output matrix amounts to computing scalar products between $\vv_{j(i)}$ and all possible value vectors, which will return $1$ only at the desired value per the orthogonal embedding construction.

\textbf{Step 2: Dimensionality Reduction to $d=O(\log \kappa+\log |V|)$. } We now reduce the embedding dimension $d = |V| + \kappa$ to $d = O(\log |V| + \log \kappa)$. To this end, we apply JL Lemma (c.f. \cref{JL:inner-prod}) to construct nearly orthogonal embedding vectors, while keeping the convolution, SSM, and output layers the same as step 1. By JL lemma, fixed $\epsilon \in (0,0.5)$, there exists a matrix $M \in \R^{d \times p}$ for $p = O(\log d)$ such that $|\langle M \ve_i, M \ve_j \rangle | \le \epsilon$ for all $i, j \in [d]$. We apply JL lemma separately for the key embedding subspace (with a JL matrix $M_{k} \in \R^{\log \kappa \times \kappa}$) and the value embedding subspace (with another JL matrix $M_{v} \in \R^{\log |V| \times |V|}$). We collect the final JL matrix via a direct sum, $M = M_k \oplus M_v$. Let $\vk_i \coloneqq M \ve_i, \vv_j \coloneqq M \ve_{\kappa + j}$. Then for $i \neq j$, we have $\langle \vk_i, \vk_j \rangle \le \epsilon, \langle \vv_i, \vv_j \rangle \le \epsilon$. On the other hand, $\langle \vk_j, \vk_j \rangle = \sum_{i=1}^p M_{i,j}^2 \approx 1$.
Similarly, we see that $\langle \vv_j, \vv_j \rangle \approx 1$.
It remains to show such low-dimensional embeddings, followed by convolutions, SSM, and output layer, yields the desired solution. The action of the convolution layer on the low-dimensional embeddings is the same as in step 1, shifting the attention-keys by one position to the right while keeping the attention-queries and attention-values unchanged.
Also the action of the SSM layer is the same as step 1 \cref{eqn:mamba2_app}, where the hidden state is a sum of key-value association matrices $\vv_{j(i)} \vk_i^{\top}$, except with embedding dimension $O(\log |V| + \log \kappa)$. At query time, upon encountering $\vk_i$, the retrieved output is $\vy_t \propto \vv_{j(i)} + \epsilon (\sum_{l \neq j(i)} \vv_l)$ for $\epsilon \ll 1$, in light of the fact that the low-dimensional embeddings are nearly orthogonal by construction. Then, applying the output matrix $W_o \vy_t $ yields the maximum component at $j(i)$, as desired.
\end{proof}

\begin{remark}\label{rem:solve_generic}
The above construction with $N = \log \kappa$ works for generic inputs where the values are drawn randomly from the vocabulary with sufficient size. Such construction may fail in the adversarial case where most values are the same and the number of keys $\kappa$ is large. Concretely, suppose the input is $[k_1, v_1, k_2, v, \ldots, k_{\kappa}, v]$, such that all the values at time $t=4,6, \ldots, 2\kappa$ are the same token $v$, and $v_1 \neq v$. Then at the retrieval time for the query token $k_1$, the signal value is $\vv_1$, whereas the noisy values from other keys contribute to $\epsilon (\kappa - 1) \vv$. Then if $\epsilon (\kappa - 1) > 1$, the model might fail to retrieve the desired value. This can be counteracted by decreasing $\epsilon$; as per JL Lemma, though, this might come at the cost of scaling the embedding dimension --- without however impacting its logarithmic behavior (remember $p=O(\epsilon^{-2}\log d)$).
\end{remark}

\medskip
\LemSfourD*

\begin{proof}
The proof is divided into two steps: we first construct a $1$-layer Mamba model using S4D as SSM layer, that solves \mqar using $d=O(\kappa |V|)$; we then apply JL Lemma with a shifting trick to complete the proof for $d=O(\kappa \log |V|)$. 

\textbf{Step 1: Construction With $d=O(\kappa |V|)$. } On a high level, the idea is to organize the hidden state of the SSM in chunks, each collecting a vector representing the value associated to a specific key, similarly to the proof in \cref{lemma:mamba_nogate}. However, unlike the Mamba layer, the S4D layer does not have access to input-dependent matrices $B(\vx), C(\vx)$, implying that the $N$ columns of the S4D state are the same up to scaling, and hence cannot encapsulate additional information regarding the input. We then decide to work with a single-column as hidden state, and instead partition it along the embedding dimension $d$. Ideally, we want the hidden state before retrieval to be a long column vector,
\begin{equation}
    \boldsymbol{h}_t =
    [\quad\boldsymbol{v}_1^{\top}\quad|
    \quad\boldsymbol{v}_2^{\top}\quad
    |\quad\dots\quad|
    \quad\boldsymbol{v}_{\kappa}^{\top}\quad]^{\top} 
    \in \R^{\kappa |V|}.
    \label{eqn:rough_h_s4d_mqar}
\end{equation}
This can be achieved by specifying each component of the full Mamba architecture as follows.
\begin{itemize}
    \item \textbf{Embedding} The goal of the embedding layer is to organize keys and values in a form that is suitable to assemble a hidden state as in \cref{eqn:rough_h_s4d_mqar}. More in detail, we let a key $k_i$ and a value $v_i$ be mapped to, respectively
    \begin{align}
        k_i\mapsto \boldsymbol{k}_i&=k\cdot[0,\dots,0|
        \quad\dots\quad|
        \smallunderbrace{1,\dots,1}_{|V|i:|V|(i+1)}|
        \quad\dots\quad|0,\dots,0|\dots]^{\top}\\
        v_i\mapsto \boldsymbol{v}_i&=v\cdot[0,\dots,\smallunderbrace{1}_i,\dots,0|
        \quad\dots\quad|
        0,\dots,\smallunderbrace{1}_{\kappa|V|+i},\dots,0]^{\top}\\
        &=v\cdot[\quad\ve_i^{\top}\quad|
        \quad\dots\quad|
        \quad\ve_i^{\top}\quad]^{\top},
    \end{align}
    for some fixed parameters $k>0,v>0$.
    Notice that, in light of this, any combination $\vk_i+\vv_j$ can form a dictionary: it has a maximum value at a component uniquely defined by the pair $i,j$.
    \item \textbf{Convolution} The (short) convolution layer is responsible for filtering out irrelevant information from the sequence, and retaining only the one pertaining $(\boldsymbol{k}_i,\boldsymbol{v}_j)$ pairs. To this end, we limit ourselves to a convolution with kernel size 2: this acts on any pair $(\boldsymbol{x}_t,\boldsymbol{x}_{t+1})$ of the input sequence by mapping it to
    \begin{equation}
        (\boldsymbol{x}_t,\boldsymbol{x}_{t+1})\mapsto \sigma\left(\boldsymbol{c}_0\odot\boldsymbol{x}_t + \boldsymbol{c}_1\odot\boldsymbol{x}_{t+1} - \boldsymbol{b}\right).
    \end{equation}
    By carefully picking the parameters $k,v,\boldsymbol{c}_0,\boldsymbol{c}_1,\boldsymbol{b}$, we can ensure that only a $(\boldsymbol{k}_i,\boldsymbol{v}_j)$ pair ``survives'' the operation, and everything else gets mapped to the null vector. This for example can be achieved by setting $k=10,v=1,\boldsymbol{c}_0=10\cdot\boldsymbol{1},
    \boldsymbol{c}_1=\boldsymbol{1},\boldsymbol{b}=k \vc_0$. With this choice, we see that the convolution maps
    \begin{align}   
    (\vk_i,\vv_j)&\mapsto[0,\dots,0|
    \quad\dots\quad|
    0,\dots,\smallunderbrace{1}_{|V|i+j},\dots,0|
    \quad\dots\quad|
    0,\dots,0]^{\top}\\
    &=[\quad\boldsymbol{0}^{\top}\quad|
    \quad\dots\quad|
    \quad\ve_j^{\top}\quad|
    \quad\dots\quad|
    \quad\boldsymbol{0}^{\top}\quad]^{\top}\\
        (\vk_i,\vk_j)&\mapsto[0,\dots,0]^{\top}\\
        (\vv_i,\vk_j)&\mapsto[0,\dots,0]^{\top}\\
        (\vv_i,\vv_j)&\mapsto[0,\dots,0]^{\top},
    \end{align}
    \item \textbf{S4D SSM} The SSM layer simply needs to accumulate and propagate the combined values down the sequence. We remind that from \cref{eqn:zoh_disc} and \cref{eqn:s4d_params}, the output of S4D is given by
    \begin{equation}
        \vh_t^{\texttt{S4D}} = \sum_{s=1}^t \vx_s  \cdot \left(\overline{\mLambda}^{\, t-(s+1)} \mB_s \right)^{\top}.
    \end{equation}
    We make use of a ``trivial'' SSM where $\overline{\mLambda}=\mB=1$, resulting in a hidden state which, after collecting the initial $(\vk_i,\vv_{j(i)})$ pairs, is constant in the form:
    \begin{equation}
    \begin{split}
        \boldsymbol{h}_t&=[0,\dots,\smallunderbrace{1}_{j_1},\dots,0|
        0,\dots,\smallunderbrace{1}_{|V|+j_2},\dots,0|\quad\dots\quad|
        0,\dots,\smallunderbrace{1}_{\kappa|V|+j_\kappa},\dots, 0]^{\top}\\
        &=[\quad\ve_{j_1}^{\top}\quad|
    \quad\ve_{j_2}^{\top}\quad
    |\quad\dots\quad|
    \quad\ve_{j_\kappa}^{\top}\quad]^{\top}.
    \end{split}
    \end{equation}
    \item \textbf{Gate} The role of the gating mechanism \cref{eqn:gating} is to retrieve the part of the hidden state which refer to the requested key. The gate branch acts on a linear transformation of the original sequence: by picking this transformation as the identity. we ensure that, when a key is encountered, only the corresponding value is retrieved from the hidden state, in fact:
        \begin{equation}
            \tilde{\boldsymbol{y}}_t=\boldsymbol{k}_i\odot\boldsymbol{h}_t=[0,\dots,0|\dots|
            0,\dots,\smallunderbrace{1}_{|V|i+j(i)},\dots,0|
            0,\dots,0|\dots].
    \end{equation}
    \item \textbf{Output} The final output layer simply needs to test the retrieved vector $\boldsymbol{y}_t$ against all values $\boldsymbol{v}_i$: only the correct one will return a scalar product different from 0. It suffices to pick
    \begin{equation}
        W_o^{\text{S4D}}=\left[\begin{array}{c}
             \boldsymbol{v}_1^{\top} \\
             \vdots\\
             \boldsymbol{v}_{|V|}^{\top}\\
        \end{array}\right]=\left[\begin{array}{c|c|c}
             \boldsymbol{e}_1^{\top} & \ldots & \boldsymbol{e}_1^{\top}  \\
             \vdots&\vdots&\vdots\\
             \boldsymbol{e}_{|V|}^{\top} & \ldots & \boldsymbol{e}_{|V|}^{\top} \\
        \end{array}\right]
        . \label{eqn:output_s4d}
    \end{equation} 
\end{itemize}

\textbf{Step 2: Dimensionality Reduction to $d=O(\kappa \log (|V|+1))$. } Having proved the construction with $d=O(|V|)$, we now apply JL Lemma to reduce the embedding dimension and suitably adjust the Mamba architecture weights to ensure the desired output. Concretely:
\begin{itemize}
    \item \textbf{Embedding} We use JL lemma to identify $d=|V|+1$ almost-orthogonal vectors within a $p \sim O(\log (|V|+1))$-dimensional space. Namely, there exists a random matrix $M: \R^{d} \to \R^p$ such that any two (different) vectors in the set $\{M \ve_1, \ldots, M\ve_{|V|}, M\ve_{|V|+1} \}$ are almost-orthogonal, in the sense that  $|\langle M\ve_i, M \ve_j \rangle| < \epsilon$ for some small $\epsilon \in (0, 0.5)$. Without loss of generality, we ensure that the last of these vectors is parallel to the all-one vector $\boldsymbol{1}_p$: notice this is always possible via an opportune rotation, which does not affect the scalar product of the recovered vectors (and hence, their almost-orthogonality). We let the embedding layer perform the following map
 \begin{align}
        k_i\mapsto \boldsymbol{k}_i&=k\cdot[0,\dots,0|
        \quad\dots\quad|
        \smallunderbrace{1,\dots,1}_{p i:p(i+1)}|
        \quad\dots\quad|0,\dots,0|\dots]^{\top} \in \R^{\kappa p},\qquad i=1\dots\kappa\\
        v_i\mapsto \boldsymbol{v}_i
        &=[(M\ve_i +\beta \mathbf{1})^{\top}|
        \quad\dots\quad|
    (M\ve_i +\beta \mathbf{1})^{\top}]^{\top} \in \R^{\kappa p},\qquad i=1\dots|V|,
    \end{align}
    where $M: \R^{d} \to \R^{p}$, $M_{i,j} \in \{-\frac{1}{\sqrt{p}} , \frac{1}{\sqrt{p}}\}$ is the (rotated) projection matrix recovered with JL Lemma, and $\beta >  \frac{1}{\sqrt{p}}$, so that each component of the value embedding $\vv_i$ is nonnegative, and falls in the range of $[\beta - \frac{1}{\sqrt{p}},  \beta + \frac{1}{\sqrt{p}}]$. 
    \item \textbf{Convolution} We use a size-$2$ convolution as in step 1, requiring it to retain only the $(\vk_i, \vv_j)$ pair information while sending other pairs $(\vv_i, \vk_j), (\vv_i, \vv_j), (\vk_i, \vk_j)$ to zero. To this end, we let
    \begin{equation}
        \beta = 1, \qquad k=10, \qquad c_0 = 10, \qquad c_1 = 1, \qquad b = k c_0. 
    \end{equation}
    Note that by the choice of $\beta$ and $p \ge 1$, the range of components in $\vv_i$ is limited to $[0, 2]$.
   Consequently, we have
    \begin{align}   
   \operatorname{conv}(\vk_i,\vv_j) &= \operatorname{ReLU}(c_0 \vk_i + c_1 \vv_j - b)
    =[\quad\boldsymbol{0}^{\top}\quad|
    \quad\dots\quad|
    \quad (M\ve_j + \beta \mathbf{1})^{\top}\quad|
    \quad\dots\quad|
    \quad\boldsymbol{0}^{\top}\quad]^{\top}; \\
     \operatorname{conv}(\vv_i,\vk_j) &=  \operatorname{conv}(\vk_i,\vk_j) =  \operatorname{conv}(\vv_i,\vv_j) = 0.
    \end{align}   
    \item \textbf{S4D SSM} The SSM layer proceeds similarly as the construction in step 1, yielding the state at $t > 2\kappa$ as
     \begin{equation}
     \vh_t =  [ (M\ve_{j_1} + \beta \textbf{1})^{\top} |
  (M \ve_{j_2}+ \beta \textbf{1})^{\top}\quad
    |\quad\dots\quad|
    (M \ve_{j_\kappa}+ \beta \textbf{1})^{\top}]^{\top}.
     \end{equation}
    \item \textbf{Gate and Output} Also the gating layer and the output layer proceed similarly as the construction in step 1. After gating, when a key $k_i$ is encountered, we have
     \begin{equation}
     \hat{\vy}_t = \vh_t\odot\vk_i =\cdot[0,\dots,0|
        \quad\dots\quad|
        \smallunderbrace{(M \ve_{j_i}+ \beta \textbf{1})^{\top}}_{p i:p(i+1)}|
        \quad\dots\quad|0,\dots,0|\dots]^{\top}.
     \end{equation}
    And finally, after applying the output matrix, we obtain
    \begin{equation}
            W_o^{\text{S4D}}\hat{\vy}_t=\left[\begin{array}{c|c|c}
             (M\boldsymbol{e}_1)^{\top} & \ldots & (M\boldsymbol{e}_1)^{\top}  \\
             \vdots&\vdots&\vdots\\
             (M\boldsymbol{e}_{|V|})^{\top} & \ldots & (M\boldsymbol{e}_{|V|})^{\top} \\
        \end{array}\right]\left[\begin{array}{c}
        \boldsymbol{0}\\
        \vdots\\
        (M \ve_{j_i}+ \beta \textbf{1})\\
        \vdots\\
        \boldsymbol{0}\\
        \end{array}\right]\approx\left[\begin{array}{c}
        \epsilon+\epsilon\beta \\
        \vdots\\
        1+\epsilon\beta\\
        \vdots\\
        \epsilon+ \epsilon\beta\\
        \end{array}\right],
    \end{equation}
    in light of the fact that both $(M\ve_i)^{\top}\cdot M\ve_j \approx \epsilon$ if $i \neq j$ otherwise $(M\ve_i)^{\top}\cdot M\ve_i \approx 1$ per JL construction, and $(M\ve_i)^{\top}\cdot\boldsymbol{1}\le\epsilon$ per the assumption that $M \ve_{|V|+1}$ is parallel to $\mathbf{1}_p$.
    This allows us to recover the correct value, completing the proof.
\end{itemize}

\end{proof}

\subsection{Proofs of Mamba Solving the \indhead Task}

\paragraph{The \indhead Task}
\label{app:indhead_proof}
As a reminder, for the \indhead task, the input is a sequence of tokens $[x_1, \ldots, x_t]$ from a finite vocabulary $V$;
The output is a sequence of tokens $[y_1, \ldots, y_t]$ from the augmented vocabulary $V \cup \{ \times \}$, where $y_i$ equals the input token 
right after the latest previous occurrence of the input token $x_i$, i.e., $y_i=x_{j(i)+1}$ where $j(i) = \max\{j: j < i, x_j = x_i \}$; otherwise $y_i = \times$. An example input and output drawn from the vocabulary $V= \{1,2,3,4\}$ with sequence length $8$ is shown below:
\begin{equation*}
\begin{matrix}
  t &=[ &1, &2, &3, &4, &5, &6, &7, &8]\\
  x &=[ &2, &1, &3, &2, &4, &3, &2, &4]\\
  y &=[ &\times, &\times, &\times, &1, &\times, &2, &4, &3] 
\end{matrix}.
\end{equation*}
Note that the input token $2$ appears three times, at instants $t=1,4,7$. Thus, at $t=7$, the \emph{latest} previous occurrence of $j(7) = 4$, which yields the output $y_7 = x_{j(7)+1} = x_{4+1}= 4$. 

\LemIndHead*

\begin{proof}
We follow a procedure similar to the \mqar construction, in that we leverage the matrix structure in the hidden state such that its columns are indexed by the key token and store the associated value token. However, differently from the \mqar task, in the \indhead task there is no distinction between the key and value set, but rather all tokens are drawn from the same vocabulary $V$ -- i.e., each token $x_i$ acts as key in the $(x_i, x_{i+1})$ pair, and as value in the $(x_{i-1}, x_i)$ pair.
To resolve this, we use a doubling-embedding trick in Mamba, together with suitable choices of convolution.
Moreover, the \indhead task requires finding the \emph{latest} previous occurrence; we will achieve this by leveraging the input-dependent state matrix. 

The main idea is to \emph{double} the embedding size in Mamba, which enables the convolution layer to perform \emph{concatenations} of the adjacent embedding pairs. Concretely: we let the state size $N = |V|$, and the embedding size $d = 2 |V|$. We design the architecture as follows.
\begin{itemize}
    \item \textbf{Embedding} We use $2|V|$-dimensional standard basis vectors to embed the vocabulary $V= \{1, 2, \ldots, |V| \}$, i.e., 
    \begin{equation}
       v_i \mapsto \begin{bmatrix}
            \ve_{v_i} \\
            \ve_{v_i}
        \end{bmatrix} \in \R^{d} \equiv \R^{2 |V|}.
    \end{equation}
    \item \textbf{Convolution} We use size-$2$ convolution (with left-padding $\mathbf{0}$) combining the pair $(\vx_{i-1}, \vx_i)$ by summing the first $|V|$-dimensions of $\vx_{i-1}$ with the last $|V|$-dimensions of $\vx_i$, effectively \emph{concatenating} $(\ve_{x_{i-1}}, \ve_{x_i})$. We let
    \begin{equation}
        \hat{\vx}_i \equiv \operatorname{conv}(\vx_{i-1}, \vx_i) = \vc_0 \odot \vx_{i-1} + \vc_1 \odot \vx_i, \qquad \text{where } 
        \vc_0 = \left[\begin{array}{c}\boldsymbol{1} \\ \boldsymbol{0}\end{array}\right] , \vc_1 = \left[\begin{array}{c}\boldsymbol{0} \\ \boldsymbol{1}\end{array}\right] .
    \end{equation}
    Note that we describe the proof for \emph{linear} convolution here to simplify notation, but it holds also for nonlinear convolution \cref{eqn:mamba_conv}, noting that each embedding vector $\vx_i$ and the convolution weights $\vc_0, \vc_1$ are nonnegative, effectively reducing the nonlinearity $\sigma = \operatorname{ReLU}$ (or $\operatorname{SiLU}$) to be the identity map. The same reasoning applies to the design of $\mB, \mC$, as we discuss next.
    \item \textbf{\Mdt SSM}
    Since the convolved output $\hat{\vx}_i$ contains $x_{i-1}$ in its first $|V|$ dimensions and $x_i$ in its last $|V|$ dimensions, we choose the state matrix $\overline{\mLambda}$ and the input matrix $\mB$ to depend on the first $|V|$ dimensions of $\hat{\vx}$ (i.e., extracting the key), and the output matrix $\mC$ to depend on the last $|V|$ dimensions (i.e., extracting the query). To this end, we let $\mLambda =-\boldsymbol{1} \in \R^{d \times N}, w_{\Delta} \gg 0$, and
    \begin{subequations}
    \begin{align}
     \Delta(\hat{\vx}_i) &\coloneqq \operatorname{SoftPlus}(\Linear(\hat{\vx}_i)), \text{ where } \Linear(\hat{\vx}_i) \coloneqq w_{\Delta}  [\mI_{{|V|}} \mid \mathbf{0}_{{|V|}}] \hat{\vx}_i \in \R^N , \label{eqn:mamba-dt-transpose-delta}\\
      \overline{\mLambda}(\hat{\vx}_i) &\coloneqq e^{\mLambda \odot (\boldsymbol{1}_d \otimes\Delta(\hat{\vx}_i))}  = e^{-\mathbf{1} \otimes \Delta(\hat{\vx}_i)} \in \R^{d \times N} \equiv \R^{|V| \times |V|} ,\label{eqn:mamba-dt-transpose}\\
      \mB(\hat{\vx}_i) &\coloneqq \Linear(\hat{\vx}_i) =  [\mI_{{|V|}} \mid \mathbf{0}_{{|V|}}] \, \hat{\vx}_i \in \R^N, \label{eqn:mamba-dt-transpose-B}\\
      \mC(\hat{\vx}_i) &\coloneqq  \Linear(\hat{\vx}_i) =  [\mathbf{0}_{{|V|}} \mid \mI_{{|V|}} ] \, \hat{\vx}_i \in \R^N.   
    \end{align}
    \end{subequations}
    \item \textbf{Output} $W_o = [\mathbf{0}_{|V|} \mid I_{|V|} ] \in \R^{|V| \times d}$
\end{itemize}

We now show the correctness of such construction. Consider the generic input and output sequences:
\begin{equation}
\begin{matrix}
  t &=[ &1, &2, &3, &4, &5, &6, &7, &8, \, \,  \ldots]\\
  x &=[ &v_2, &v_1, &v_3, &v_2, &v_4, &v_3, &v_2, &v_4,  \, \, \ldots]\\
  y &=[ &\times, &\times, &\times, &v_1, &\times, &v_2, &v_4, &v_3, \, \, \ldots]. 
\end{matrix} \, \label{eqn:ind_head_data}
\end{equation}

After the embedding and convolution layers, the  SSM input is a sequence of $d$-dimensional vectors for $d=2|V|$,
\begin{equation}
    \hat{\vx} = \left[\begin{array}{c|c|c|c|c|c|c|c|c}
            \mathbf{0} & \ve_{v_2} & \ve_{v_1} & \ve_{v_3} & 
            \ve_{v_2} & \ve_{v_4} & \ve_{v_3} & 
            \ve_{v_2}  &\ldots \\
            \ve_{v_2} & \ve_{v_1} & \ve_{v_3} & 
            \ve_{v_2} & \ve_{v_4} & \ve_{v_3} & 
            \ve_{v_2} &
            \ve_{v_4} &\ldots\\
        \end{array}\right] \in \R^{2|V| \times t},
\end{equation}
where $\hat{\vx}_i$ stores the $(x_{i-1}, x_i)$ pair.

The action of the SSM layer organizes the hidden state matrix of size $d \times N \equiv 2|V| \times |V|$ by the input matrix $\mB(\hat{\vx}_i)$ taking outer-product with $\hat{\vx}_i$, followed by retrieving the desired column via the output matrix $\mC(\hat{\vx}_i)$. Now by the choice of SSM parameters, we have
\begin{equation}
\left[\begin{array}{c}
\mB(\hat{\vx})\\\hline
\mC(\hat{\vx})
\end{array}\right] = \left[\begin{array}{c|c|c|c|c|c|c|c|c}
            \mathbf{0}&
            \ve_{v_2} & \ve_{v_1} & \ve_{v_3} &
            \ve_{v_2} & \ve_{v_4} & \ve_{v_3} &
            \ve_{v_2} & \ldots\\\hline
            \ve_{v_2} & \ve_{v_1} & \ve_{v_3} &
            \ve_{v_2} & \ve_{v_4} & \ve_{v_3} &
            \ve_{v_2} &
            \ve_{v_4} & \ldots
\end{array}\right]\in \R^{2|V| \times t},
\label{eqn:mamba-dtperp-dts}
\end{equation}
where we stack them together to visualize that $\mB(\hat{\vx})$ amounts to shifting $\mC(\hat{\vx})$ to the right by one position, due to the design of $\operatorname{conv}_B$.

We now verify the desired behavior in the SSM layer. Suppose temporarily the state matrix is $\overline{\mLambda} = \boldsymbol{1} \in \R^{d \times N}$. Then the hidden state at time $s \le t$ would be a cumulative sum,
\begin{equation}
    \vh_s =\sum_{i=1}^s \hat{\vx}_i \mB(\hat{\vx}_i)^{\top} = \sum_{i=2}^s  \begin{bmatrix}
            \ve_{x_{i-1}} \\
            \ve_{x_i}
        \end{bmatrix} \ve_{x_{i-1}}^{\top}. \label{eqn:mamba_dt_trans_hidden}
\end{equation}
    Thus, the $j$-th column of the hidden state matrix would store the sum of all $\hat{\vx}_i$ where the key $\mB(\hat{\vx}_i) = \vx_{i-1} = \ve_j$. Yet the \indhead task requires storing the \emph{latest} associated value only (not \emph{all} associated values). To this end, we leverage the input-dependence of the state matrix, and particularly of $\Delta(x_t)$. Recall the \Mdt layer is given by   
 \begin{align}
      \vh_s 
      &= e^{\mLambda \odot (\boldsymbol{1} \otimes\Delta(\hat{\vx}_s))}  \odot \vh_{s-1} + \hat{\vx}_s \mB(\hat{\vx}_s)^{\top}, \\
      \vy_s &=\vh_s \,  \mC(\hat{\vx}_s) .
     \end{align}
We design $\Delta(\hat{\vx}_t) \in \R^N \equiv \R^{|V|}$ such that when the input contains the key information, the corresponding key column in the state is \emph{erased} (while the other columns remain the same). Without loss of generality, suppose $\mB(\hat{\vx}_s)=\ve_j$. By the definition of $\mB$ \cref{eqn:mamba-dt-transpose-B}, this implies that $[\mI_{|V|} \mid \boldsymbol{0}_{|V|}] \hat{\vx}_s = \ve_j$. By the choice of $w_{\Delta } \gg 0$ and the definition of $\Delta$ \cref{eqn:mamba-dt-transpose-delta}, we have 
\begin{equation}
 \Delta(\hat{\vx}_i) = \operatorname{SoftPlus}(w_{\Delta} [\mI_d \mid \mathbf{0}_d] \hat{\vx}_i ) = w_{\Delta} \ve_j.   
\end{equation}
Therefore \cref{eqn:mamba-dt-transpose} yields the state matrix as
\begin{equation}
  \overline{\mLambda}_s = e^{-\boldsymbol{1} \otimes \Delta(\hat{\vx}_s)} = e^{-\boldsymbol{1} \otimes (w_{\Delta}\ve_j)}= \boldsymbol{1} \otimes \exp{[0, \ldots, \underbrace{-w_{\Delta}}_{j}, \ldots, 0]^{\top}}  \overset{w_{\Delta} \to \infty}{=} \boldsymbol{1} \otimes (\mathbf{1} - \ve_{j}) \in \R^{d \times N}, \label{eqn:mamba-dt-trans-state-mat}
\end{equation}
which has an all-zeros $j$-th column and all-ones columns elsewhere. Consequently, the $j$-th column of the hidden state $\vh_s[:,j]$ is erased by the action $e^{-\boldsymbol{1} \otimes \Delta(\hat{\vx}_s)}  \odot \vh_{s-1}$, and then updated with the current input containing the latest value by the action $ \hat{\vx}_s \otimes \mB(\hat{\vx}_s)$, as desired. We remark that such erasure operation is akin to the construction of the S6 layer for solving the \textsc{Keep $n$-th} task in \cref{cor:keepN}, in which we have Mamba approximate a Heaviside by tweaking $\Delta(\vx_t)$, so to erase information from all tokens before a given one. We also see that such \emph{selective} erasure works consistently well for long sequences when $t \to \infty$, since it preserves all other columns (except the $j$-th one) by setting $\Delta(\vu_s)[l] = 0$ for $l \neq j$, and thereby satisfying the condition in \cref{lemma:constant} to avoid sensitivity decay.

Finally, the SSM output is given by $\vy_s = \vh_s \mC_s = \vh_s \ve_{x_s}$, which retrieves the $x_s$-th column of the state that stores the token immediately after the \emph{latest} previous occurrence of $x_s$, i.e. $\vy_s = \begin{bmatrix}
            \ve_{x_{j(s)}} \\
            \ve_{x_{j(s)+1}}
        \end{bmatrix} $ . We then apply the output matrix $W_o \vy_s =  \ve_{x_{j(s)+1}}$ to obtain the target value, which completes the proof.

\end{proof}

\clearpage

\section{Additional Experiment Details} \label{app:experiments}

\subsection{Training Details}
For all experiments, unless otherwise noted, we train with the Adam optimizer \cite{kingma2014adam}, for $600$ epochs using an initial learning rate $\eta=0.03$ and cosine annealing down to $\eta=1\times 10^{-6}$. The training set is composed of $10^5$ randomly generated samples with a fixed seed and the batch size is $16$, which results in up to $3.75\cdot10^6$ gradient updates. We perform early stopping if the validation loss reaches below $10^{-6}$ or if six hours have elapsed since the beginning of training. The validation set and test sets have $10^3$ and $10^5$ samples respectively and are generated with the same function but using different seeds. Reported accuracies are always obtained from the test set after the last epoch of training.

\subsection{Task \textsc{Keep $n$-th}} \label{app:experiments-keepnth}

In this section, we provide additional details and ablation of the \keepn task used for \cref{sec:layer}, with experimental set-up and partial results reported in \cref{tab:keep-fifth}. 

\textbf{Model} All the results in \cref{tab:keep-nth-sweep-PE} - \cref{tab:keep-nth-sweep-cond} use $1$-layer models. The \textsc{Mamba} and \textsc{S4D} models (with or without PE) are simplified architectures, which consists of embedding layer, SSM layer, and output linear layer, \emph{without convolution nor gating} from the Mamba mixer block. The simplification is intended to investigate the role of the SSM layer alone (i.e. S6 versus S4D), without confounding factors from other components in the mixer block.

\textbf{Experiment Set-up} For each input sequence $\vx=(x_1, \ldots, x_T)$, we draw $x_i \overset{i.i.d}{\sim} \operatorname{Unif}(\{1,\ldots, |V|\})$ randomly from a vocabulary with size $|V|=128$. The target output is the $n$-th token in the input sequence, i.e. $y_t = x_n$ for $ n < t \le T$. The predicted output at time $t = n, \ldots, T$ are taken to compute (cross-entropy) loss for training, and accuracy for evaluation.

\textbf{Discussion} When equipped with PE, Mamba manages to achieve 100\% accuracy on \keepn, regardless of sequence length. This is thanks to its ability to dynamically adjust (via $\Delta(x_t)$) for how long the hidden state retains memory of the target token (in position $n=5$). On the other hand, S4D is lacking such ability, and already fails at the task for $T=20$ (\cref{tab:keep-nth-sweep-PE,tab:keep-nth-sweep-PE-cond}). Then again Transformers do not need to retain memory, as they can look back to the whole sequence at each step, in light of their attention mechanism, and have no issue solving the task for any $T$. When removing PE from Mamba, however (\cref{tab:keep-nth-sweep,tab:keep-nth-sweep-cond}), the model loses its way to discriminate the specific token that must be retrieved, and performance drops to that of S4D, as expected.

\footnotesize
\begin{longtable}{llllllllll}
\caption{Ablation on \keepn: \textsc{Mamba+PE}, \textsc{S4D+PE}, \textsc{Transformers} with varying sequence length $T=10,20$, embedding dimension $d$, and state size $N$.}\label{tab:keep-nth-sweep-PE}\\ 
    \toprule
     &  & \multicolumn{3}{c}{T=10} & \multicolumn{3}{c}{T=20} \\
     &  & acc. & \# epch. & \# prm. & acc. & \# epch. & \# prm. \\
    \midrule
    \endfirsthead
    \toprule
     &  & \multicolumn{3}{c}{T=10} & \multicolumn{3}{c}{T=20} \\
     &  & acc. & \# epch. & \# prm. & acc. & \# epch. & \# prm. \\
    \midrule
    \endhead
    \midrule
    \midrule
    \endfoot
    \bottomrule
    \endlastfoot
    \multirow[c]{9}{*}{\shortstack[l]{\textsc{Mamba}\\\textsc{(+PE)}}}
     &  d=8, N=8  & 1.00 (0.00) & 107 & 2.2k & 1.00 (0.00) & 37 & 2.2k \\
     &  d=8, N=32  & 1.00 (0.00) & 241 & 2.8k & 1.00 (0.00) & 39 & 2.8k \\
     &  d=8, N=64  & 1.00 (0.00) & 161 & 3.5k & 1.00 (0.00) & 47 & 3.5k \\
     &  d=32, N=8  & 1.00 (0.00) & 56 & 9.2k & 1.00 (0.00) & 15 & 9.2k \\
     &  d=32, N=32  & 1.00 (0.00) & 19 & 11.5k & 1.00 (0.00) & 14 & 11.5k \\
     &  d=32, N=64  & 1.00 (0.00) & 88 & 14.5k & 1.00 (0.00) & 19 & 14.5k \\
     &  d=64, N=8  & 0.99 (0.00) & 553 & 18.7k & 0.99 (0.01) & 208 & 18.7k \\
     &  d=64, N=32  & 1.00 (0.00) & 429 & 23.3k & 1.00 (0.00) & 12 & 23.3k \\
     &  d=64, N=64  & 1.00 (0.00) & 345 & 29.4k & 1.00 (0.00) & 12 & 29.4k \\
    \cline{1-8}
    \multirow[c]{9}{*}{\shortstack[l]{\textsc{S4D}\\\textsc{(+PE)}}}
     &  d=8, N=8  & 0.99 (0.00) & 600 & 2.1k & 0.43 (0.04) & 600 & 2.1k \\
     &  d=8, N=32  & 0.99 (0.00) & 600 & 2.5k & 0.46 (0.05) & 600 & 2.5k \\
     &  d=8, N=64  & 0.99 (0.00) & 600 & 3.0k & 0.43 (0.03) & 600 & 3.0k \\
     &  d=32, N=8  & 0.94 (0.00) & 600 & 8.7k & 0.72 (0.01) & 600 & 8.7k \\
     &  d=32, N=32  & 0.93 (0.01) & 600 & 10.3k & 0.72 (0.01) & 600 & 10.3k \\
     &  d=32, N=64  & 0.93 (0.01) & 600 & 12.4k & 0.73 (0.01) & 600 & 12.4k \\
     &  d=64, N=8  & 0.79 (0.00) & 600 & 17.5k & 0.14 (0.01) & 600 & 17.5k \\
     &  d=64, N=32  & 0.79 (0.00) & 600 & 20.6k & 0.14 (0.00) & 600 & 20.6k \\
     &  d=64, N=64  & 0.79 (0.00) & 600 & 24.8k & 0.14 (0.00) & 600 & 24.8k \\
    \cline{1-8}
    \multirow[c]{3}{*}{\shortstack[l]{\textsc{Trans-}\\\textsc{former}}} & l=1, d=16 & 1.00 (0.00) & 16 & 5.8k & 1.00 (0.00) & 19 & 6.0k \\
     & l=1, d=32 & 1.00 (0.00) & 12 & 15.1k & 1.00 (0.00) & 12 & 15.4k \\
     & l=1, d=64 & 1.00 (0.00) & 10 & 42.3k & 1.00 (0.00) & 9 & 42.9k \\
    \cline{1-8}
\end{longtable}
\clearpage
\begin{longtable}{lllllllllllll}
\caption{Ablation on \keepn: \textsc{Mamba+PE}, \textsc{S4D+PE}, \textsc{Transformers} with varying sequence length $T=30,40,50$, embedding dimension $d$, and state size $N$.}\label{tab:keep-nth-sweep-PE-cond}\\ 
    \toprule
     &  & \multicolumn{3}{c}{T=30} & \multicolumn{3}{c}{T=40} & \multicolumn{3}{c}{T=50} \\
     &  & acc. & \# epch. & \# prm. & acc. & \# epch. & \# prm. & acc. & \# epch. & \# prm. \\
    \midrule
    \endfirsthead
    \toprule
     &  & \multicolumn{3}{c}{T=30} & \multicolumn{3}{c}{T=40} & \multicolumn{3}{c}{T=50} \\
     &  & acc. & \# epch. & \# prm. & acc. & \# epch. & \# prm. & acc. & \# epch. & \# prm. \\
    \midrule
    \endhead
    \midrule
    \midrule
    \endfoot
    \bottomrule
    \endlastfoot
    \multirow[c]{9}{*}{\shortstack[l]{\textsc{Mamba}\\\textsc{(+PE)}}}
     &  d=8, N=8  & 0.94 (0.04) & 456 & 2.2k & 0.85 (0.15) & 325 & 2.2k & 0.22 (0.15) & 600 & 2.2k \\
     &  d=8, N=32  & 1.00 (0.00) & 285 & 2.8k & 0.58 (0.37) & 416 & 2.8k & 0.28 (0.10) & 600 & 2.8k \\
     &  d=8, N=64  & 1.00 (0.00) & 180 & 3.5k & 0.59 (0.31) & 420 & 3.5k & 0.98 (0.02) & 346 & 3.5k \\
     &  d=32, N=8  & 1.00 (0.00) & 20 & 9.2k & 1.00 (0.00) & 50 & 9.2k & 1.00 (0.00) & 241 & 9.2k \\
     &  d=32, N=32  & 1.00 (0.00) & 20 & 11.5k & 1.00 (0.00) & 21 & 11.5k & 1.00 (0.00) & 225 & 11.5k \\
     &  d=32, N=64  & 1.00 (0.00) & 19 & 14.5k & 1.00 (0.00) & 23 & 14.5k & 1.00 (0.0) & 130 & 14.5k \\
     &  d=64, N=8  & 1.00 (0.00) & 17 & 18.7k & 1.00 (0.00) & 164 & 18.7k & 0.99 (0.00) & 600 & 18.7k \\
     &  d=64, N=32  & 1.00 (0.00) & 24 & 23.3k & 1.00 (0.00) & 66 & 23.3k & 0.99 (0.00) & 600 & 23.3k \\
     &  d=64, N=64  & 1.00 (0.00) & 29 & 29.4k & 1.00 (0.00) & 226 & 29.4k & 0.98 (0.01) & 600 & 29.4k \\
    \cline{1-11}
    \multirow[c]{9}{*}{\shortstack[l]{\textsc{S4D}\\\textsc{(+PE)}}}
     &  d=8, N=8  & 0.14 (0.04) & 600 & 2.1k & 0.03 (0.00) & 600 & 2.1k & 0.03 (0.00) & 600 & 2.1k \\
     &  d=8, N=32  & 0.46 (0.05) & 600 & 2.5k & 0.15 (0.01) & 600 & 2.5k & 0.04 (0.01) & 600 & 2.5k \\
     &  d=8, N=64  & 0.43 (0.03) & 600 & 3.0k & 0.13 (0.02) & 600 & 3.0k & 0.04 (0.00) & 600 & 3.0k \\
     &  d=32, N=8  & 0.72 (0.01) & 600 & 8.7k & 0.09 (0.00) & 600 & 8.7k & 0.08 (0.00) & 600 & 8.7k \\
     &  d=32, N=32  & 0.72 (0.01) & 600 & 10.3k & 0.09 (0.00) & 600 & 10.3k & 0.09 (0.00) & 600 & 10.3k \\
     &  d=32, N=64  & 0.73 (0.01) & 600 & 12.4k & 0.09 (0.00) & 600 & 12.4k & 0.09 (0.00) & 600 & 12.4k \\
     &  d=64, N=8  & 0.14 (0.01) & 600 & 17.5k & 0.09 (0.00) & 600 & 17.5k & 0.09 (0.00) & 600 & 17.5k \\
     &  d=64, N=32  & 0.14 (0.00) & 600 & 20.6k & 0.10 (0.00) & 600 & 20.6k & 0.10 (0.00) & 600 & 20.6k \\
     &  d=64, N=64  & 0.14 (0.00) & 600 & 24.8k & 0.09 (0.00) & 600 & 24.8k & 0.09 (0.00) & 600 & 24.8k \\
    \cline{1-11}
    \multirow[c]{3}{*}{\shortstack[l]{\textsc{Trans-}\\\textsc{former}}} & l=1, d=16 & 1.00 (0.00) & 16 & 5.8k & 1.00 (0.00) & 23 & 6.3k & 1.00 (0.00) & 17 & 6.4k \\
     & l=1, d=32 & 1.00 (0.00) & 12 & 15.1k & 1.00 (0.00) & 12 & 16.0k  & 1.00 (0.00) & 13 & 16.4k\\
     & l=1, d=64 & 1.00 (0.00) & 9 & 42.3k & 1.00 (0.00) & 9 & 44.2k & 1.00 (0.00) & 9 & 44.9k \\
    \cline{1-11}
\end{longtable}

\begin{longtable}{lllllllllllll}
\caption{Ablation on \keepn: \textsc{Mamba} and \textsc{S4D} with varying sequence length $T=10,20$, embedding dimension $d$, and state size $N$.}\label{tab:keep-nth-sweep}\\ 
    \toprule
     &  & \multicolumn{3}{c}{T=10} & \multicolumn{3}{c}{T=20} \\
     &  & acc. & \# epch. & \# prm. & acc. & \# epch. & \# prm. \\
    \midrule
    \endfirsthead
    \toprule
     &  & \multicolumn{3}{c}{T=10} & \multicolumn{3}{c}{T=20} \\
     &  & acc. & \# epch. & \# prm. & acc. & \# epch. & \# prm. \\
    \midrule
    \endhead
    \midrule
    \midrule
    \endfoot
    \bottomrule
    \endlastfoot
    \multirow[c]{9}{*}{\textsc{Mamba}} 
     &  d=8, N=8  & 0.20 (0.01) & 600 & 2.3k & 0.05 (0.00) & 600 & 2.3k \\
     &  d=8, N=32  & 0.18 (0.03) & 600 & 2.9k & 0.05 (0.00) & 600 & 2.9k \\
     &  d=8, N=64  & 0.17 (0.02) & 600 & 3.6k & 0.05 (0.00) & 600 & 3.6k \\
     &  d=32, N=8  & 0.21 (0.01) & 600 & 9.3k & 0.10 (0.00) & 600 & 9.3k \\
     &  d=32, N=32  & 0.25 (0.03) & 600 & 11.6k & 0.11 (0.00) & 600 & 11.6k \\
     &  d=32, N=64  & 0.28 (0.03) & 600 & 14.7k & 0.11 (0.00) & 600 & 14.7k \\
     &  d=64, N=8  & 0.19 (0.00) & 600 & 18.8k & 0.11 (0.00) & 600 & 18.8k \\
     &  d=64, N=32  & 0.20 (0.00) & 600 & 23.4k & 0.11 (0.00) & 600 & 23.4k \\
     &  d=64, N=64  & 0.21 (0.00) & 600 & 29.6k & 0.11 (0.00) & 600 & 29.6k \\
     \cline{1-8}
    \multirow[c]{9}{*}{\textsc{S4D}} 
     &  d=8, N=8  & 0.08 (0.00) & 600 & 2.2k & 0.04 (0.00) & 600 & 2.2k \\
     &  d=8, N=32  & 0.08 (0.00) & 600 & 2.6k & 0.04 (0.00) & 600 & 2.6k \\
     &  d=8, N=64  & 0.08 (0.00) & 600 & 3.2k & 0.04 (0.00) & 600 & 3.2k \\
     &  d=32, N=8  & 0.20 (0.00) & 600 & 8.8k & 0.10 (0.00) & 600 & 8.8k \\
     &  d=32, N=32  & 0.20 (0.00) & 600 & 10.4k & 0.10 (0.00) & 600 & 10.4k \\
     &  d=32, N=64  & 0.20 (0.00) & 600 & 12.5k & 0.10 (0.00) & 600 & 12.5k \\
     &  d=64, N=8  & 0.20 (0.00) & 600 & 17.7k & 0.11 (0.00) & 600 & 17.7k \\
     &  d=64, N=32  & 0.20 (0.00) & 600 & 20.8k & 0.11 (0.00) & 600 & 20.8k \\
     &  d=64, N=64  & 0.20 (0.00) & 600 & 24.9k & 0.11 (0.00) & 600 & 24.9k \\
     \cline{1-8}
\end{longtable}
\clearpage
\begin{longtable}{lllllllllllll}
\caption{Ablation on \keepn: \textsc{Mamba} and \textsc{S4D} with varying sequence length $T=30,40,50$, embedding dimension $d$, and state size $N$.}\label{tab:keep-nth-sweep-cond}\\ 
    \toprule
     &  & \multicolumn{3}{c}{T=30} & \multicolumn{3}{c}{T=40} & \multicolumn{3}{c}{T=50} \\
     &  & acc. & \# epch. & \# prm. & acc. & \# epch. & \# prm. & acc. & \# epch. & \# prm. \\
    \midrule
    \endfirsthead
    \toprule
     &  & \multicolumn{3}{c}{T=30} & \multicolumn{3}{c}{T=40} & \multicolumn{3}{c}{T=50} \\
     &  & acc. & \# epch. & \# prm. & acc. & \# epch. & \# prm. & acc. & \# epch. & \# prm. \\
    \midrule
    \endhead
    \midrule
    \midrule
    \endfoot
    \bottomrule
    \endlastfoot
    \multirow[c]{9}{*}{\textsc{Mamba}} 
     &  d=8, N=8  & 0.04 (0.00) & 600 & 2.3k & 0.04 (0.00) & 600 & 2.3k & 0.03 (0.00) & 600 & 2.3k \\
     &  d=8, N=32  & 0.04 (0.00) & 600 & 2.9k & 0.04 (0.00) & 600 & 2.9k & 0.03 (0.00) & 600 & 2.9k \\
     &  d=8, N=64  & 0.04 (0.00) & 600 & 3.6k & 0.04 (0.00) & 600 & 3.6k & 0.03 (0.00) & 600 & 3.6k \\
     &  d=32, N=8  & 0.09 (0.00) & 600 & 9.3k & 0.09 (0.00) & 600 & 9.3k & 0.08 (0.00) & 600 & 9.3k \\
     &  d=32, N=32  & 0.09 (0.00) & 600 & 11.6k & 0.09 (0.00) & 600 & 11.6k & 0.08 (0.00) & 600 & 11.6k \\
     &  d=32, N=64  & 0.09 (0.00) & 600 & 14.7k & 0.09 (0.00) & 600 & 14.7k & 0.08 (0.00) & 600 & 14.7k \\
     &  d=64, N=8  & 0.09 (0.00) & 600 & 18.8k & 0.09 (0.00) & 600 & 18.8k & 0.09 (0.00) & 600 & 18.8k \\
     &  d=64, N=32  & 0.09 (0.00) & 600 & 23.4k & 0.09 (0.00) & 600 & 23.4k & 0.09 (0.00) & 600 & 23.4k \\
     &  d=64, N=64  & 0.09 (0.00) & 600 & 29.6k & 0.09 (0.00) & 600 & 29.6k & 0.09 (0.00) & 600 & 29.6k \\
     \cline{1-11}
    \multirow[c]{9}{*}{\textsc{S4D}} 
     &  d=8, N=8  & 0.03 (0.00) & 600 & 2.2k & 0.03 (0.00) & 600 & 2.2k & 0.03 (0.00) & 600 & 2.2k \\
     &  d=8, N=32  & 0.03 (0.00) & 600 & 2.6k & 0.03 (0.00) & 600 & 2.6k & 0.03 (0.00) & 600 & 2.6k \\
     &  d=8, N=64  & 0.03 (0.00) & 600 & 3.2k & 0.03 (0.00) & 600 & 3.2k & 0.03 (0.00) & 600 & 3.2k \\
     &  d=32, N=8  & 0.08 (0.00) & 600 & 8.8k & 0.09 (0.00) & 600 & 8.8k & 0.09 (0.00) & 600 & 8.8k \\
     &  d=32, N=32  & 0.08 (0.00) & 600 & 10.4k & 0.09 (0.00) & 600 & 10.4k & 0.09 (0.00) & 600 & 10.4k \\
     &  d=32, N=64  & 0.08 (0.00) & 600 & 12.5k & 0.09 (0.00) & 600 & 12.5k & 0.09 (0.00) & 600 & 12.5k \\
     &  d=64, N=8  & 0.09 (0.00) & 600 & 17.7k & 0.09 (0.00) & 600 & 17.7k & 0.09 (0.00) & 600 & 17.7k \\
     &  d=64, N=32  & 0.09 (0.00) & 600 & 20.8k & 0.09 (0.00) & 600 & 20.8k & 0.09 (0.00) & 600 & 20.8k \\
     &  d=64, N=64  & 0.09 (0.00) & 600 & 24.9k & 0.09 (0.00) & 600 & 24.9k & 0.09 (0.00) & 600 & 24.9k \\
     \cline{1-11}
\end{longtable}

\clearpage
\subsection{Task \mqar} \label{app:mqar}

\begin{figure}[h!]
    \centering
    \includegraphics[width=100mm]{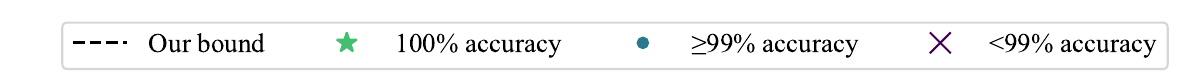}
    \\[5mm]
    $|V|=256$
    \\[2mm]
    \hfill
    \begin{subfigure}[b]{0.3\textwidth}
        {\tiny \textsc{S4D}}
        \\ \centering
        \begin{sideways}
            \makebox[30mm][c]{\tiny Model dimension ($d$)}
        \end{sideways}%
        \includegraphics[trim={1mm 7mm 12mm 15mm},clip,width=35mm]{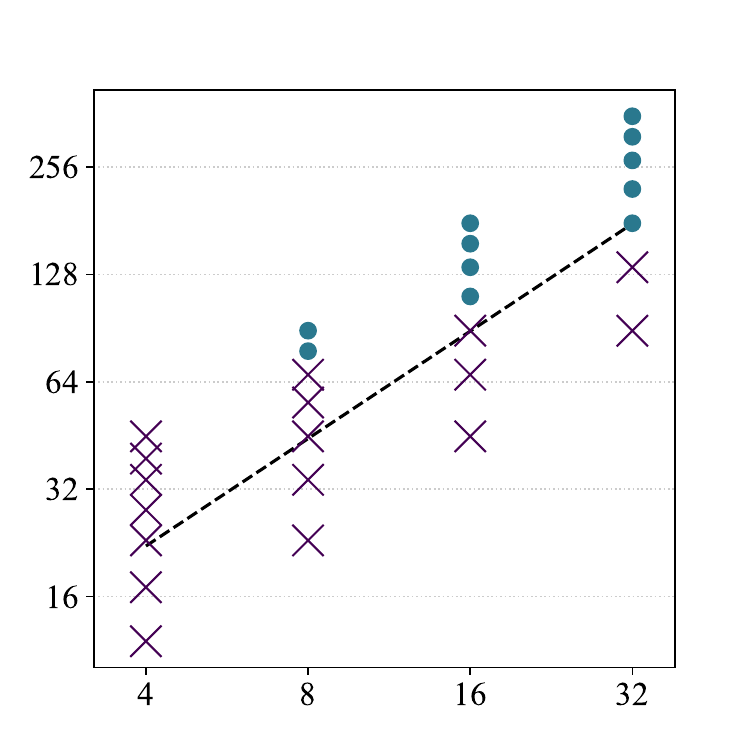}%
    \end{subfigure}
    \hfill
    \begin{subfigure}[b]{0.3\textwidth}
        {\tiny \textsc{Mamba}}
        \\ \centering
        \includegraphics[trim={1mm 7mm 12mm 15mm},clip,width=35mm]{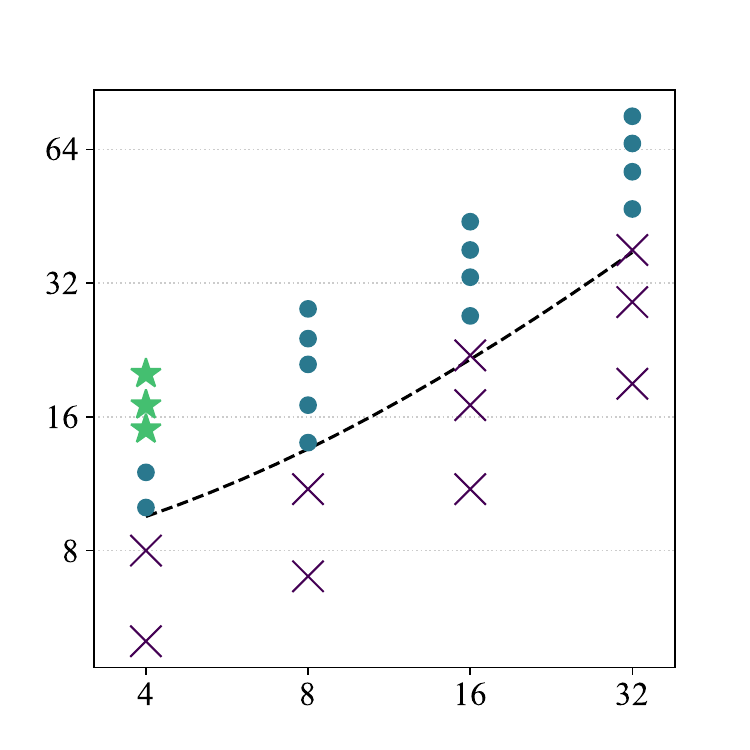}
    \end{subfigure}
    \hfill
    \begin{subfigure}[b]{0.3\textwidth}
        {\tiny \textsc{Mamba-2}}
        \\ \centering
    \includegraphics[trim={1mm 7mm 12mm 15mm},clip,width=35mm]{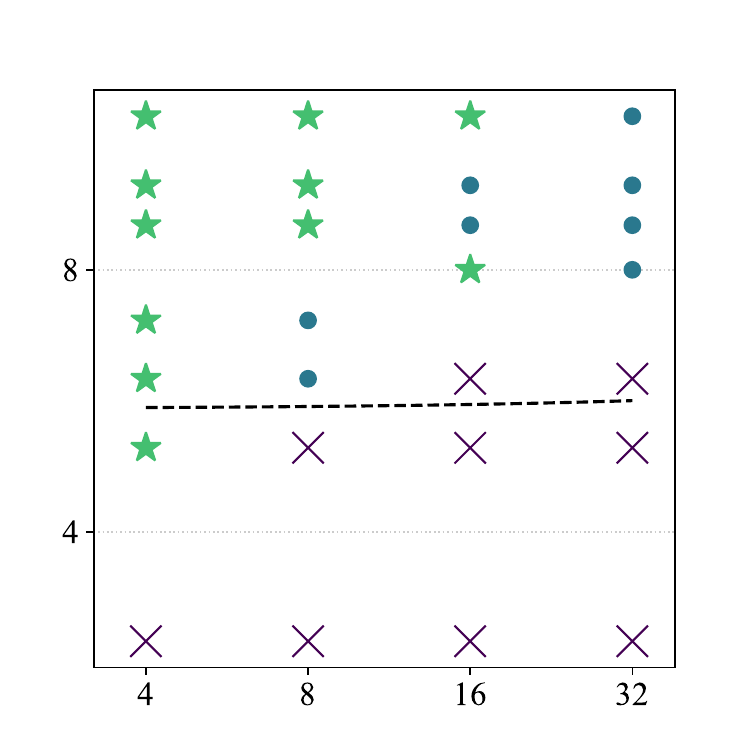}%
    \end{subfigure}
    \hfill
    \\[-1mm]
    {\tiny Number of keys ($\kappa$)}
    \\[5mm]
    $|V|=512$
    \\[2mm]
    \hfill
    \begin{subfigure}[b]{0.3\textwidth}
        {\tiny \textsc{S4D}}
        \\ \centering
        \begin{sideways}
            \makebox[30mm][c]{\tiny Model dimension ($d$)}
        \end{sideways}%
        \includegraphics[trim={1mm 7mm 12mm 15mm},clip,width=35mm]{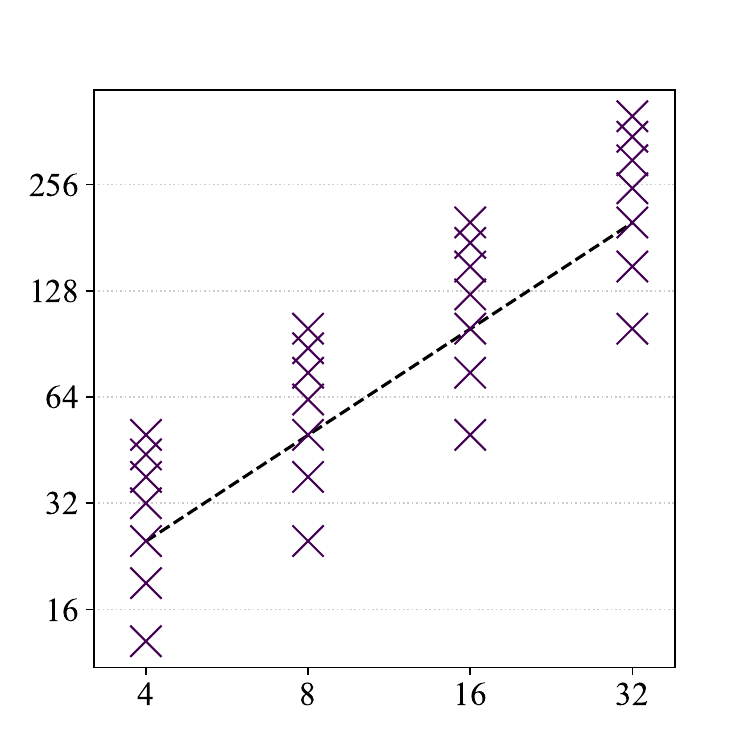}%
    \end{subfigure}
    \hfill
    \begin{subfigure}[b]{0.3\textwidth}
        {\tiny \textsc{Mamba}}
        \\ \centering
        \includegraphics[trim={1mm 7mm 12mm 15mm},clip,width=35mm]{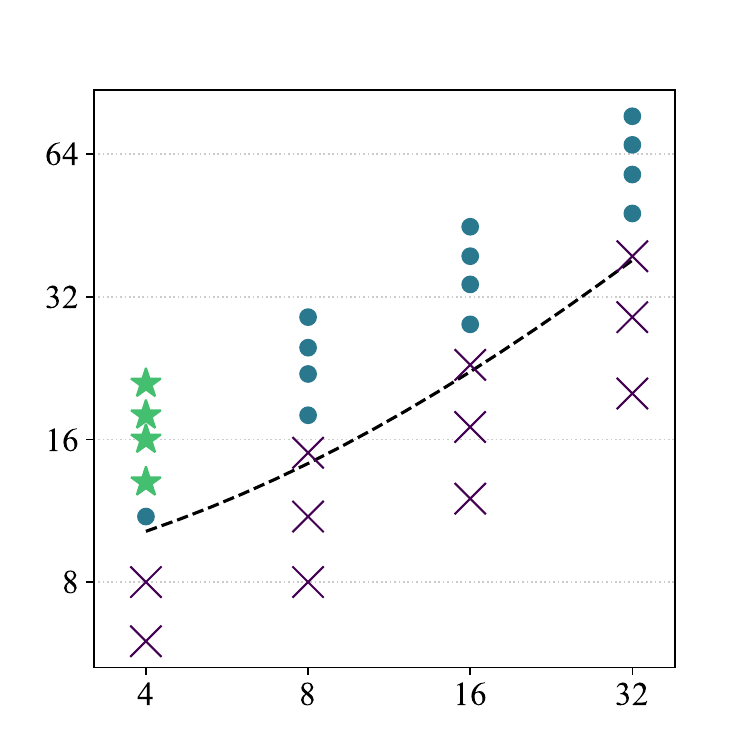}
    \end{subfigure}
    \hfill
    \begin{subfigure}[b]{0.3\textwidth}
        
        {\tiny \textsc{Mamba-2}}
        \\ \centering
        \includegraphics[trim={1mm 7mm 12mm 15mm},clip,width=35mm]{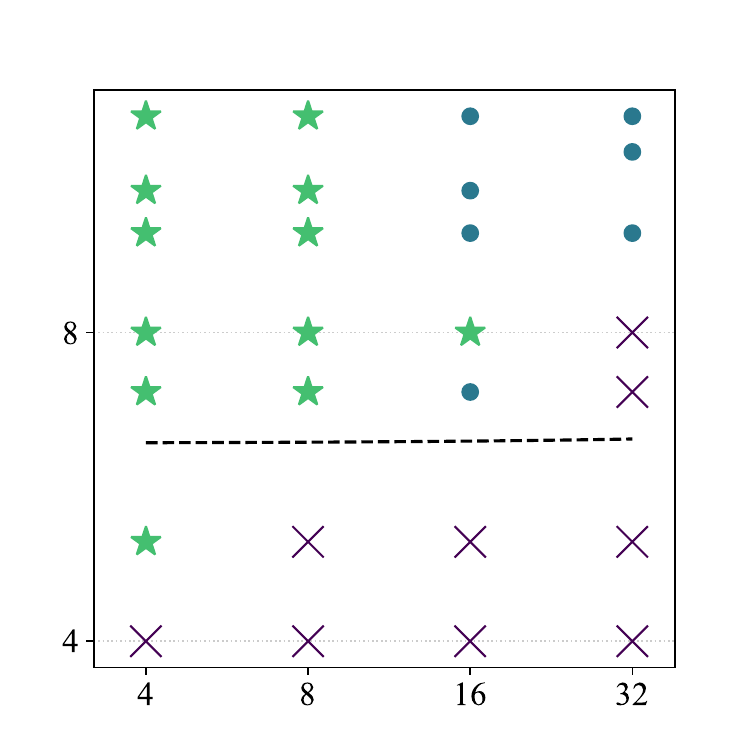}%
    \end{subfigure}
    \hfill
    \\[-1mm]
    {\tiny Number of keys ($\kappa$)}
    \caption{Trained models accuracy on \mqar (best of 7 seeds) across $\kappa$, and $d$. For S4D $N = 4$, for Mamba $N=2\times \kappa$, and for Mamba2 $N=8\times\ln{\kappa}$. $T=100$ and $|V| \in \{256, 512\}$ for all runs.}
    \label{fig:MQAR_three_graphs}
\end{figure}

\textbf{Models} All the results in \cref{fig:MQAR_three_graphs} use $1$-layer models. The Mamba and \MM models explicitly disable the gating branch. This simplification is intended to verify our constructions without gating in \cref{{lemma:mamba_nogate}} and \cref{lemma:mqar_attn_sol}. The Mamba-S4D model retains the full Mamba architecture, but only swapping the S6 layer with the S4D layer; not to be confused with the original S4D model proposed in \cite{gu2022parameterization}.

\textbf{Experiment Set-up} We generate the data described in \cref{app.proofs_mqar} as follows. For each input sequence of the form $$\vx = [k_1, v_1, \ldots, k_{\kappa}, v_{\kappa}, \ldots, \mid k_{i_1}, \ldots, k_{i_2}],$$ we draw the key token $k_i \overset{i.i.d}{\sim} \operatorname{Unif}(\{1,\ldots, \kappa\})$, and value token $v_j \overset{i.i.d}{\sim} \operatorname{Unif}(\{1,\ldots, |V|\})$. The target output sequence consists of masked tokens except at the the query chunk where the input query is a key (e.g., at $k_{i_1}, k_{i_2}$ in the example above). We compute loss during training (and accuracy for evaluation) only at the query positions, informed from the target output sequence. 

\textbf{Discussion} Here we expand on the results from \cref{fig:mqar_sweeps} in the main text, by reporting the accuracy of Mamba, \MM, and Mamba-S4D trained on \mqar for varying model sizes. In \cref{fig:MQAR_three_graphs}, we sweep over values of the value vocabulary size $|V|$, to show how our theoretical bounds hold while varying this parameter. The bounds are still reasonably tight, and the observations drawn from \cref{fig:mqar_sweeps} still hold in this case. The extra caveat (which does not however invalidate our claim) is that by increasing $|V|$ we are making the task more difficult to solve, and our training procedure for the simplest Mamba-S4D fails to achieve satisfactory performance for the model sizes considered. Notice also that, for \MM, the simplest task $\kappa=4$ can achieve $100\%$ accuracy even \emph{below} the theoretical curve proposed in our theorems. This can be attributed to the following factors. On the one hand, our bounds rely on JL Lemma, which provides only \emph{asymptotical} behaviors which might not be verified in practice for $\kappa$ so small. On the other hand, it is perfectly feasible that at this regime the architecture can recover a more efficient solution than the one theorized.

\begin{figure}[h!]
    \centering
    \includegraphics[width=100mm]{figures/mqar_sweep/legend.pdf}
    \\[5mm]
    \textsc{S4D}
    \\[2mm]
    \begin{subfigure}[b]{0.20\textwidth}
        {\tiny $N=1$}
        \\ \centering
        \begin{sideways}
            \makebox[30mm][c]{\tiny Model dimension ($d$)}
        \end{sideways}%
        \includegraphics[trim={1mm 7mm 12mm 15mm},clip,width=\textwidth]{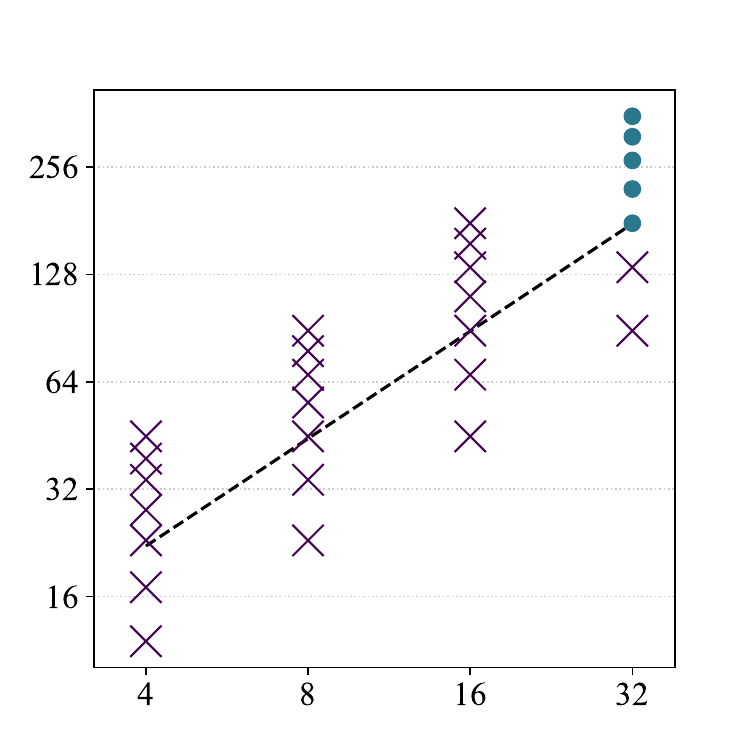}%
    \end{subfigure}
    \hfill
    \begin{subfigure}[b]{0.20\textwidth}
        {\tiny $N=2$}
        \\ \centering
        \includegraphics[trim={1mm 7mm 12mm 15mm},clip,width=\textwidth]{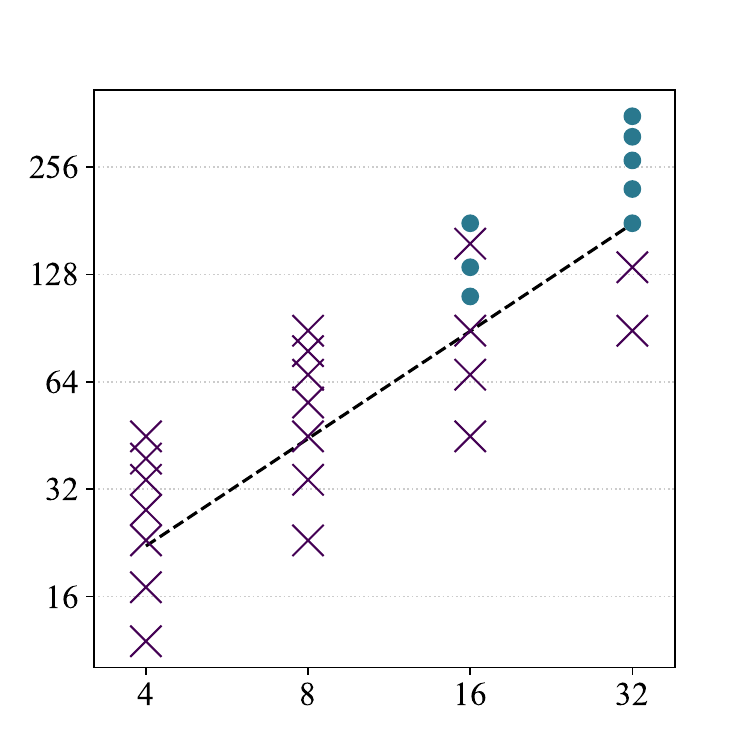}%
    \end{subfigure}
    \hfill
    \begin{subfigure}[b]{0.20\textwidth}
        {\tiny $N=4$}
        \\ \centering
        \includegraphics[trim={1mm 7mm 12mm 15mm},clip,width=\textwidth]{figures/mqar_sweep/s4d_n4_vocab256.pdf}%
    \end{subfigure}
    \hfill
    \begin{subfigure}[b]{0.20\textwidth}
        {\tiny $N=8$}
        \\ \centering
        \includegraphics[trim={1mm 7mm 12mm 15mm},clip,width=\textwidth]{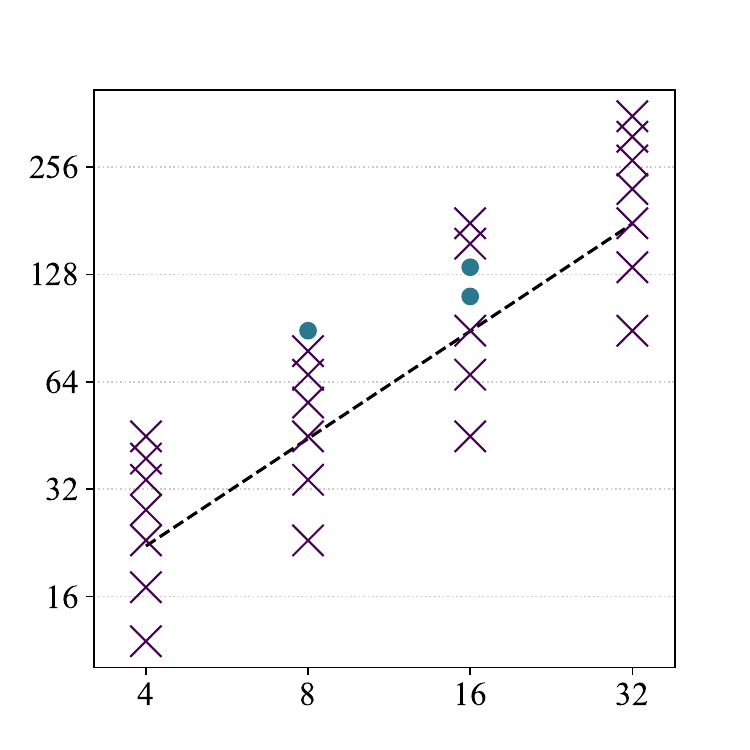}%
    \end{subfigure}
    \\[-1mm]
    {\tiny Number of keys ($\kappa$)}
    \\[5mm]
    \textsc{Mamba}
    \\[2mm]
    \begin{subfigure}[b]{0.20\textwidth}
        {\tiny $N=0.5\times\kappa$}
        \\ \centering
        \begin{sideways}
            \makebox[30mm][c]{\tiny Model dimension ($d$)}
        \end{sideways}%
        \includegraphics[trim={1mm 7mm 12mm 15mm},clip,width=\textwidth]{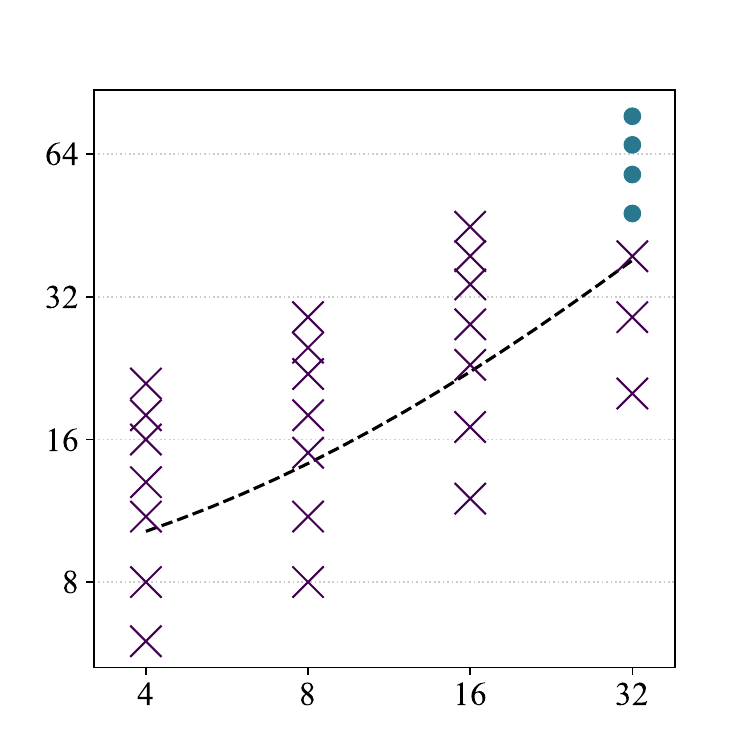}%
    \end{subfigure}
    \hfill
    \begin{subfigure}[b]{0.20\textwidth}
        {\tiny $N=\kappa$}
        \\ \centering
        \includegraphics[trim={1mm 7mm 12mm 15mm},clip,width=\textwidth]{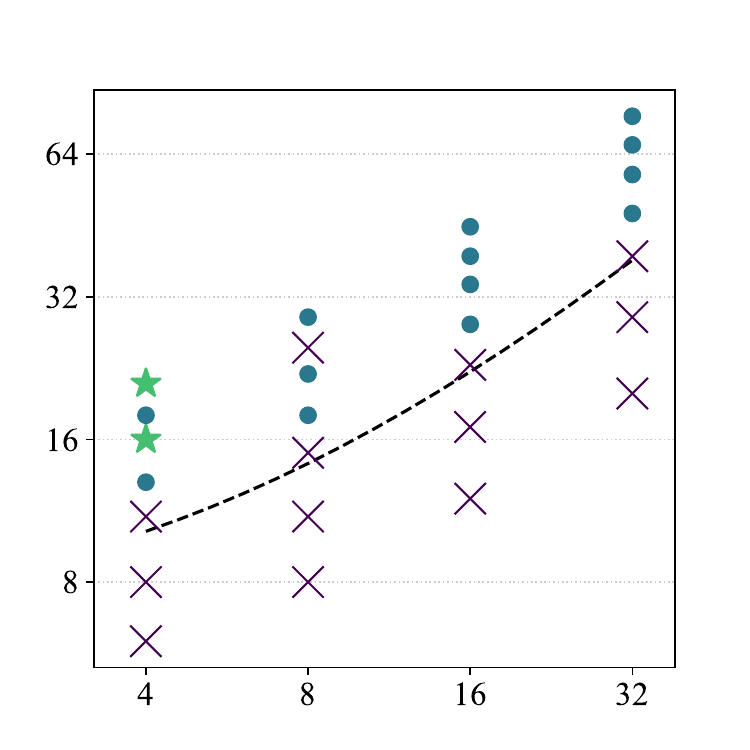}%
    \end{subfigure}
    \hfill
    \begin{subfigure}[b]{0.20\textwidth}
        {\tiny $N=2\times\kappa$}
        \\ \centering
        \includegraphics[trim={1mm 7mm 12mm 15mm},clip,width=\textwidth]{figures/mqar_sweep/mamba_n2_vocab512.pdf}%
    \end{subfigure}
    \hfill
    \begin{subfigure}[b]{0.20\textwidth}
        {\tiny $N=4\times\kappa$}
        \\ \centering
        \includegraphics[trim={1mm 7mm 12mm 15mm},clip,width=\textwidth]{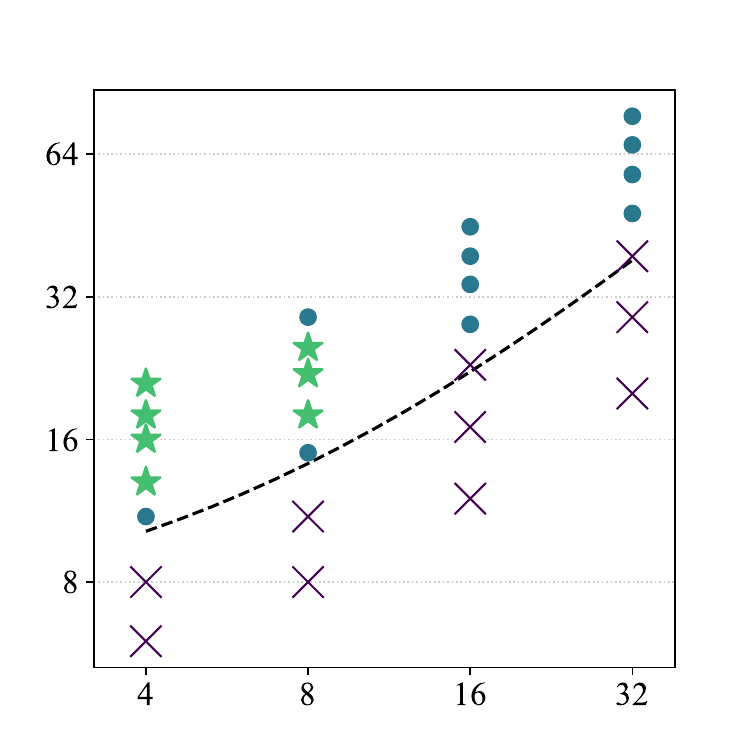}%
    \end{subfigure}
    \\[-1mm]
    {\tiny Number of keys ($\kappa$)}
    \\[5mm]
    \textsc{Mamba-2}
    \\[2mm]
    \begin{subfigure}[b]{0.20\textwidth}
        {\tiny $N=\log\kappa$}
        \\ \centering
        \begin{sideways}
            \makebox[30mm][c]{\tiny Model dimension ($d$)}
        \end{sideways}%
        \includegraphics[trim={1mm 7mm 12mm 15mm},clip,width=\textwidth]{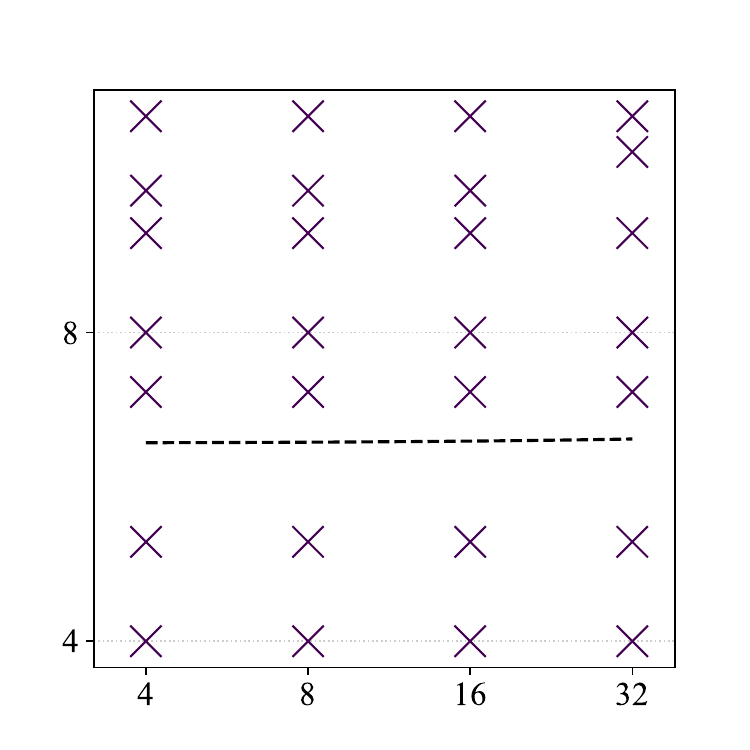}%
    \end{subfigure}
    \hfill
    \begin{subfigure}[b]{0.20\textwidth}
        {\tiny $N=2\times\log\kappa$}
        \\ \centering
        \includegraphics[trim={1mm 7mm 12mm 15mm},clip,width=\textwidth]{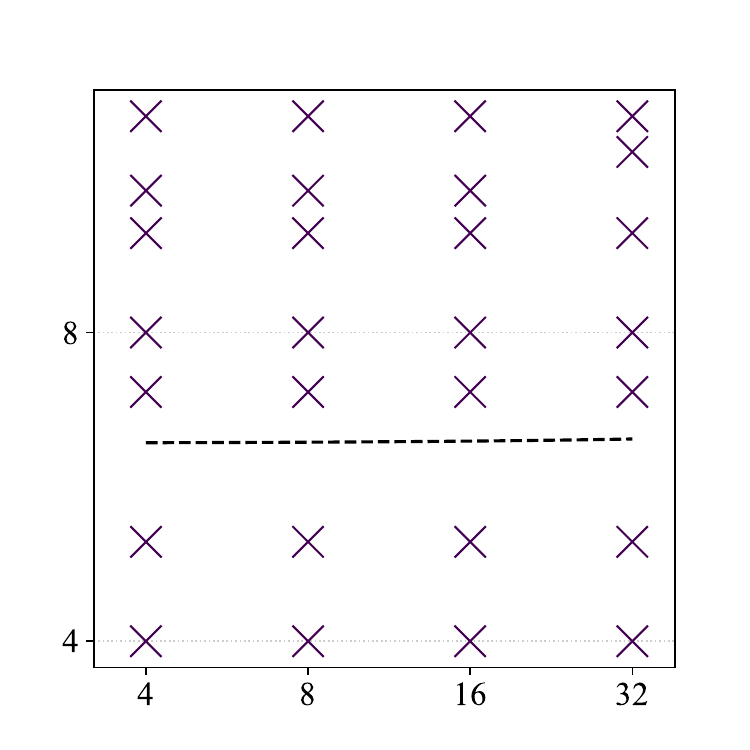}%
    \end{subfigure}
    \hfill
    \begin{subfigure}[b]{0.20\textwidth}
        {\tiny $N=4\times\log\kappa$}
        \\ \centering
        \includegraphics[trim={1mm 7mm 12mm 15mm},clip,width=\textwidth]{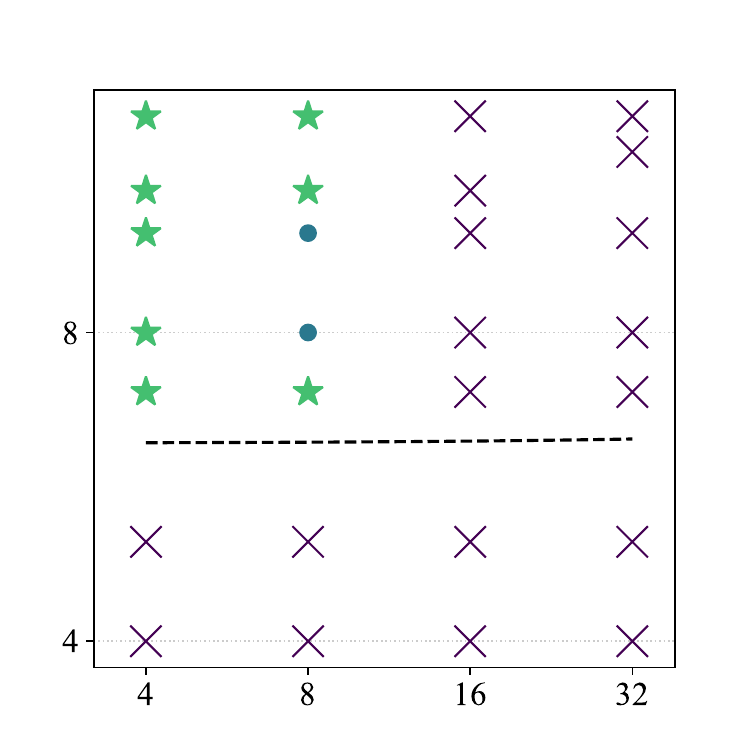}%
    \end{subfigure}
    \hfill
    \begin{subfigure}[b]{0.20\textwidth}
        {\tiny $N=8\times\log\kappa$}
        \\ \centering
        \includegraphics[trim={1mm 7mm 12mm 15mm},clip,width=\textwidth]{figures/mqar_sweep/mamba2_n8_vocab512.pdf}%
    \end{subfigure}
    \\[-1mm]
    {\tiny Number of keys ($\kappa$)}
    \caption{Trained models accuracy on \mqar (best of 7 seeds) across $N$, $\kappa$, and $d$. For Mamba and \MM $|V|=512$, while for S4D $|V|=256$ (as it failed to reach satisfactory accuracy for larger $|V|$). $T=100$ for all runs.}
    \label{fig:MQAR_three_graphs_Nsweep}
\end{figure}

With \cref{fig:MQAR_three_graphs_Nsweep}, we further complement our results by sweeping over values of the $N$ state size parameter. We remind that, according to \cref{lemma:s4d_mqar,lemma:mamba_nogate,lemma:mqar_attn_sol}, our theorized \mqar solutions require a value of at least $N=1$, $N=\kappa$ and $N\sim\log\kappa$ for S4D, Mamba and \MM, respectively. Indeed, in \cref{fig:MQAR_three_graphs_Nsweep} we observe that varying $N$ does not have a particular impact on the final accuracy of S4D. For Mamba, on the other hand, we see that for $N<\kappa$ the training procedure fails to recover an exact solution to \mqar. Similarly, for \MM, no solution is recovered for $N<4\log\kappa$. These results further validate the tightness of our theoretical solutions.

\clearpage
\subsection{Task \indhead}

\textbf{Model} All the results in \cref{fig:MQAR_three_graphs} use $1$-layer Mamba that explicitly disables the gating branch. This simplification is intended to verify our constructions without gating in \cref{lem:ind_head}. 

\begin{figure}[h!]
    \centering
    \includegraphics[width=100mm]{figures/one_hop/legend.pdf}
    \\[5mm]
    \hfill
    \begin{subfigure}[b]{0.3\textwidth}
        {\tiny $N=|V|$}
        \\ \centering
        \begin{sideways}
            \makebox[30mm][c]{\tiny Model dimension ($d$)}
        \end{sideways}%
        \includegraphics[trim={1mm 7mm 12mm 15mm},clip,width=35mm]{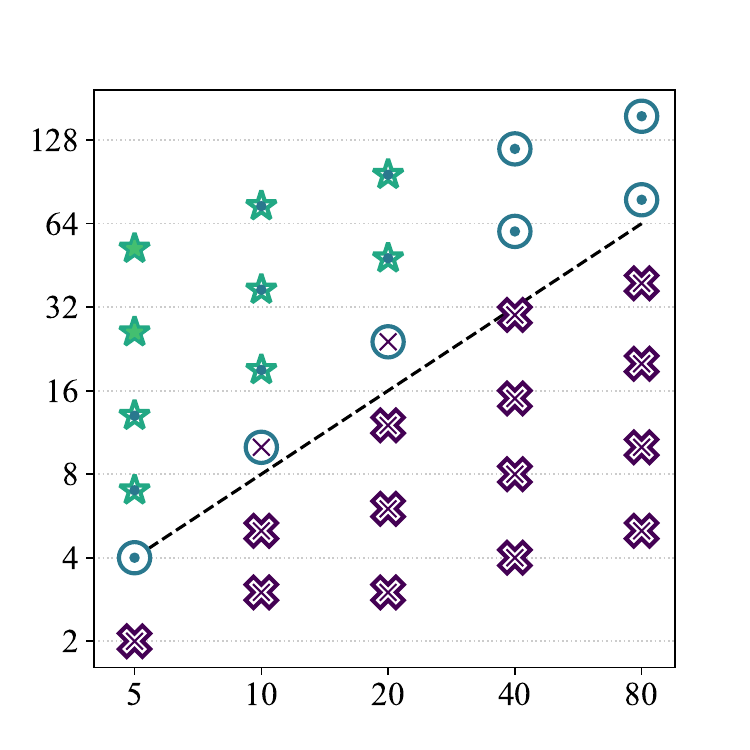}%
    \end{subfigure}
    \hfill
    \begin{subfigure}[b]{0.3\textwidth} 
        {\tiny $N=2|V|$}
        \\ \centering
        \includegraphics[trim={1mm 7mm 12mm 15mm},clip,width=35mm]{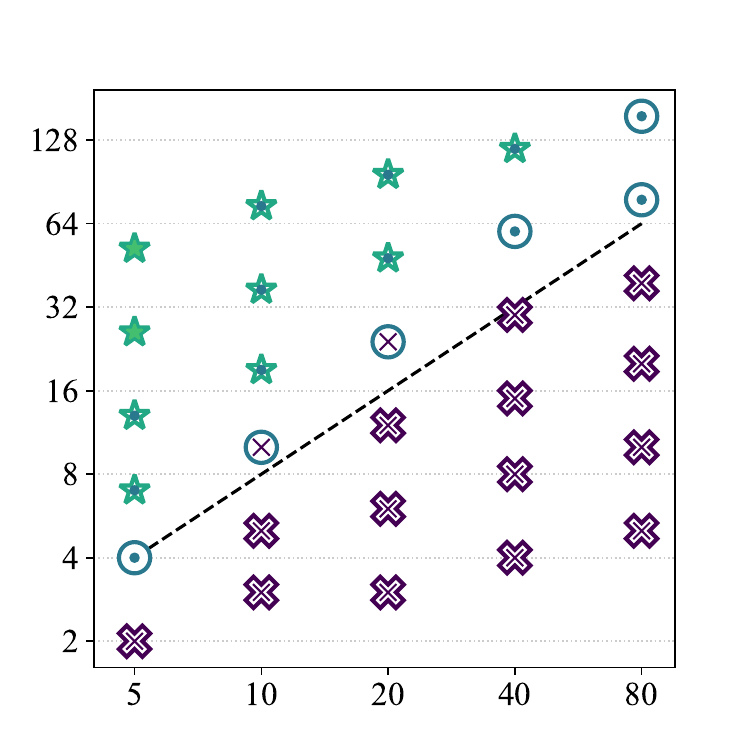}%
    \end{subfigure}
    \hfill
    \begin{subfigure}[b]{0.3\textwidth} 
        {\tiny $N=4|V|$}
        \\ \centering
        \includegraphics[trim={1mm 7mm 12mm 15mm},clip,width=35mm]{figures/one_hop/one_hop_n4_t100.pdf}%
    \end{subfigure}
    \\[-1mm]
    {\tiny Vocabulary size ($|V|$)}
    \caption{Trained models accuracy on \indhead task (best of 5 seeds), varying $|V|$ and $d$, with $N\in \{|V|, 2|V|, 4|V|\}$ (left, middle, right). \Mdt's performance (outlined) is equal or better than Mamba's (filled) and only hits $100\%$ above the theoretical bound from \cref{lem:ind_head} (black).}
    \label{fig:ind_head_ablation}
\end{figure}

\textbf{Experiment Set-up} We generate the data described in \cref{app:indhead_proof} as follows. The data generation requires four scalar parameters: vocabulary size $|V|$, sequence length $T > 2|V| + 1$, hard case probability $p \in [0,1]$, and special range ratio $\gamma \in (0,0.1]$. We first draw $X \sim \operatorname{Bernoulli}(p)$: if $X = 0$, we sample from the standard setting, otherwise the hard setting. The standard setting generates the input sequence $\vx = (x_1, \ldots, x_T)$ by randomly drawing $x_i \overset{i.i.d,}{\sim} V = \{1, \ldots, |V|\}$ for $i=1, \ldots, T$. The hard setting is intended to evaluate the long-range memorization capability (i.e., placing repeated tokens at the beginning and the end of the sequence), which consists of the following steps.
\begin{enumerate}
    \item Randomly pick a special token $v^* \in V$
    \item Generate the input sequence by randomly drawing $\vx = (x_1, \ldots, x_T)$ where $x_i \overset{i.i.d,}{\sim} V\setminus v^*$ for $i=1, \ldots, T$
    \item Randomly draw a position from $r \in \operatorname{Unif}(\{ 1, \ldots, \gamma T\})$
    \item Place the special token $v^*$ at positions $r, T-r$.
\end{enumerate}
In the experiments for \cref{fig:one_hop}, we use $T = 100, |V| \in \{5, 10, 20, 40\}, p=0.75, \gamma = 0.1$. In \cref{fig:ind_head_ablation}, we further ablate the choice of state size $N \in \{|V|, 2|V|, 4|V| \}$.

\textbf{Discussion} We design the hard setting to better differentiate the capabilities from Mamba and our proposed \Mdt. Specifically, solving the standard setting of the \indhead task requires memorizing the latest previous occurrence, or forgetting the earlier previous occurrences. Note that the input sequence generated from the standard setting consists of many repeated tokens (by the requirement $T > 2|V| + 1$), and the expected time for reappearance of any token is $|V|$. Thus, for small and medium-size $|V|$, Mamba can solve for these cases by using the state matrix with negative eigenvalues to discount the remote past pairs, and thereby correctly output the latest previous occurrence. However, solving the hard setting additionally requires the model to memorize long-range information due to the special token (occurring at the beginning part and the end part of the sequence). We see that the Mamba solution with negative eigenvalues is \emph{at odds with} the long-range memorization, as shown in \cref{lemma:exp_decay}. On the other hand, \Mdt can satisfy both selective forgetting and long-range memorization via the input-dependence state matrix that erases outdated information specific to the input key, while retaining other information in the hidden state, as illustrated in the proof of \cref{lem:ind_head} (see details in \cref{app:indhead_proof}).

Below we expand on the results in \cref{sec:induction_head_theory} by reporting a sweep on the hidden state size $N$ for the models used in the \indhead task experiments, complementing the findings shown in \cref{fig:one_hop}.

\end{document}